\documentclass[12pt]{book}
\usepackage{float,amssymb,amsfonts,epsfig,makeidx}
\usepackage{amsmath,amsthm}
\usepackage{color}
\theoremstyle{plain}
\newtheorem{thm}{Theorem}[chapter]
\newtheorem{theorem}[thm]{Theorem}

\newtheorem{proposition}[thm]{Proposition}

\theoremstyle{definition}

\newtheorem{definition}{Definition}[chapter]

\input xy
\xyoption{matrix}
\xyoption{arrow}
\xyoption{rotate}
\xyoption{cmtip}
\xyoption{graph}
\xyoption{frame}

\def\reals{\mathbb{R}}

\def\co{\colon}
\def\mapdef#1#2#3{#1\co #2\rightarrow #3}
\def\transpos#1{#1^{\top}}
\def\s#1{{\cal #1}}
\def\norme#1{\left\|#1\right\|}
\def\remark{\bigskip\noindent{\bf Remark:}\enspace}
\def\smnorme#1{\|#1\|}
\def\lag{\left\langle}
\def\rag{\right\rangle}
\def\mfrac#1{{\mathfrak{#1}}}
\title{\huge 
Notes on Elementary Spectral Graph Theory\\
Applications to Graph Clustering \\
Using Normalized Cuts\\
}
\author{Jean Gallier\\
Department of Computer and Information Science\\
University of Pennsylvania\\
Philadelphia, PA 19104, USA\\
e-mail: {\tt jean@cis.upenn.edu}\\
\ \\
\copyright\ Jean Gallier}
\begin{document}
\maketitle
\tableofcontents
\vfill\eject
\chapter{Introduction}
\label{chap-intro}
\def\Degsym{D}%
In the Fall of 2012, my friend Kurt Reillag suggested that I 
should be ashamed about knowing so little about 
graph Laplacians and normalized graph cuts.
These notes are the result of my efforts to rectify this 
situation.

\medskip
I begin with a review of basic notions of graph theory.
Even though the graph Laplacian is fundamentally associated with an
undirected graph, I review the definition of both directed and
undirected graphs. For both directed and undirected graphs, I define
the degree matrix $D$, the incidence matrix $\widetilde{D}$, and the
adjacency matrix $A$. I also define weighted graphs (with nonnegative
weights), and the notions of {\it volume\/}, $\mathrm{vol}(A)$  of a
set of nodes $A$, of
{\it links\/}, $\mathrm{links}(A, B)$ between two sets of nodes $A, B$,
and of {\it cut\/}, $\mathrm{cut}(A) = \mathrm{links}(A, \overline{A})$ of a set
of nodes $A$. These concepts play a crucial role in the theory of
normalized cuts. Then, I introduce the (unnormalized) {\it graph Laplacian\/} $L$ of
a directed graph $G$ in an  ``old-fashion,'' 
by showing that for any orientation of a graph
$G$, 
\[
\widetilde{D}\transpos{\widetilde{D}} = D - A = L
\]
is an invariant. I also define the (unnormalized) {\it graph
  Laplacian\/} $L$ of a weighted graph $(V, W)$ as $L = D - W$, and
prove that
\[
\transpos{x} L x =
\frac{1}{2}\sum_{i, j = 1}^m w_{i\, j} (x_i - x_j)^2
\quad\mathrm{for\ all}\> x\in \reals^m.
\]
Consequently, $\transpos{x} L x$ does not depend on the
diagonal entries in $W$, and if $w_{i\, j} \geq 0$ for all $i, j\in
\{1, \ldots,m\}$, then $L$ is positive semidefinite.
Then, if $W$ consists of nonnegative entries, 
the eigenvalues  $0 = \lambda_1 \leq \lambda_2 \leq  \ldots \leq 
\lambda_m$ of $L$ are real and nonnegative, and there is an
orthonormal basis of eigenvectors of $L$.
I show that the
number of connected components of the graph $G = (V,W)$
is equal to the dimension of the kernel of $L$.

\medskip
I also define the normalized graph Laplacians $L_{\mathrm{sym}}$ and $L_{\mathrm{rw}}$, given by
\begin{align*}
L_{\mathrm{sym}}& = \Degsym^{-1/2} L \Degsym^{-1/2} = I - \Degsym^{-1/2} W \Degsym^{-1/2}  \\
L_{\mathrm{rw}}& = \Degsym^{-1} L = I - \Degsym^{-1} W,
\end{align*}
and prove some simple properties relating the eigenvalues and the
eigenvectors of $L$, $L_{\mathrm{sym}}$ and $L_{\mathrm{rw}}$.
These normalized graph Laplacians show up when dealing with normalized cuts.

\medskip
Next, I turn to {\it graph drawings\/} (Chapter \ref{chap2}).
Graph drawing is a very
attractive application of so-called spectral techniques, which is a
fancy way of saying that that
eigenvalues and eigenvectors of the graph Laplacian are used.
Furthermore, it turns out that graph clustering using normalized cuts
can be cast as a certain type of graph drawing.

\medskip
Given an undirected graph $G = (V, E)$, with $|V| = m$, 
we would like to draw $G$ in $\reals^n$ for $n$ (much) smaller than
$m$. 
The idea is to assign a point $\rho(v_i)$ in $\reals^n$ to the vertex $v_i\in
V$, for every $v_i \in V$, 
and to draw a line segment between the points $\rho(v_i)$ and
$\rho(v_j)$.  Thus, a {\it graph drawing\/} is a function
$\mapdef{\rho}{V}{\reals^n}$.

\medskip
We define the  {\it matrix of a graph drawing $\rho$ 
(in  $\reals^n$)\/}  as a $m \times n$ matrix $R$ whose $i$th row consists
of the row vector $\rho(v_i)$ corresponding to the point representing $v_i$ in
$\reals^n$.
Typically, we want $n < m$; in fact $n$ should be much smaller than $m$.

\medskip
Since there are infinitely many graph drawings, it is desirable to
have some criterion to decide which graph is better than another.
Inspired by a physical model in which the edges are springs, 
it is natural to consider a representation to be better if it
requires the springs to be less extended. 
We can formalize this by
defining the {\it energy\/} of a drawing $R$ by
\[
\s{E}(R) = \sum_{\{v_i, v_j\}\in E} \norme{\rho(v_i) - \rho(v_j)}^2,
\]
where $\rho(v_i)$ is the $i$th row of $R$ and 
$\norme{\rho(v_i) - \rho(v_j)}^2$
is the square of the Euclidean length of the line segment
joining $\rho(v_i)$  and  $\rho(v_j)$.

\medskip
Then, ``good drawings''  are drawings that minimize the energy function
$\s{E}$.
Of course, the trivial representation corresponding to the zero matrix
is optimum, so we need to impose extra constraints to rule out the
trivial solution.

\medskip
We can consider the more general situation where the springs are not
necessarily identical. This can be modeled by a symmetric weight (or
stiffness)  matrix $W = (w_{i j})$, with $w_{i j} \geq
0$.
In this case,  our energy function becomes
\[
\s{E}(R) = \sum_{\{v_i, v_j\}\in E} w_{i j} \norme{\rho(v_i) - \rho(v_j)}^2.
\]

Following Godsil and Royle \cite{Godsil}, 
we prove that 
\[
\s{E}(R) = \mathrm{tr}(\transpos{R} L R),
\]
where 
\[
L = \Degsym - W,
\]
is the familiar unnormalized Laplacian matrix associated with $W$,
and where $\Degsym$ is the degree matrix associated with $W$.

\medskip
It can be shown that there is no loss in generality in 
assuming that the columns of $R$ are pairwise orthogonal
and that they have unit length. Such a matrix satisfies the equation
$\transpos{R} R = I$ and the corresponding drawing is called an
{\it orthogonal drawing\/}. This condition also rules out trivial
drawings.

\medskip
Then, I prove the main theorem about graph drawings
(Theorem \ref{graphdraw}), which essentially says that
the matrix $R$ of the desired graph drawing is constituted by the
$n$ eigenvectors of $L$ associated with the
smallest nonzero $n$ eigenvalues of $L$. 
We give a number examples of graph drawings, many of which are borrowed
or adapted from Spielman \cite{Spielman}.

\medskip
The next chapter (Chapter \ref{chap3}) 
contains  the ``meat'' of this document. This chapter is devoted to 
the method of normalized graph cuts for graph clustering.
This beautiful and deeply original method first published in
Shi and Malik \cite{ShiMalik},
has now come to be a ``textbook chapter''
of computer vision and machine learning. It was invented  by Jianbo Shi 
and Jitendra Malik, and was the main topic of Shi's dissertation.
This method was extended to $K \geq 3$ clusters by Stella
 Yu in her dissertation \cite{Yu}, and is also the subject of Yu and Shi
\cite{YuShi2003}.

\medskip
Given a set of data, the goal of clustering is to partition the data
into different groups according to their similarities. When the data
is given in terms of a similarity graph $G$, where the weight $w_{i\, j}$
between two nodes $v_i$ and $v_j$ is a measure of similarity of
$v_i$ and $v_j$, the problem can be stated as follows:
Find a partition $(A_1, \ldots, A_K)$ of the set of nodes $V$ into
different groups such   that the edges between different groups have
very low weight (which indicates that the points in different clusters
 are dissimilar),  and the edges within a group have high weight
 (which indicates that points within the same cluster are similar).

\medskip
The above graph clustering problem can be formalized as an
optimization problem, using the notion of cut mentioned earlier.
If we want to partition $V$ into $K$ clusters, we can do so by finding
a partition ($A_1, \ldots, A_K$) that  minimizes the quantity
\[
\mathrm{cut}(A_1, \ldots, A_K) = \frac{1}{2} \sum_{1 = 1}^K \mathrm{cut}(A_i). 
\]

For $K = 2$,  the mincut problem is a classical
problem that can be solved efficiently, but in practice, it does not
yield satisfactory partitions. Indeed, in many cases, the mincut
solution separates one vertex from the rest of the graph.  What we
need is to design our cost function in such a way that it keeps the
subsets $A_i$ ``reasonably large'' (reasonably balanced).

\begin{figure}[http]
  \begin{center}
 \includegraphics[height=2.5truein,width=2.8truein]{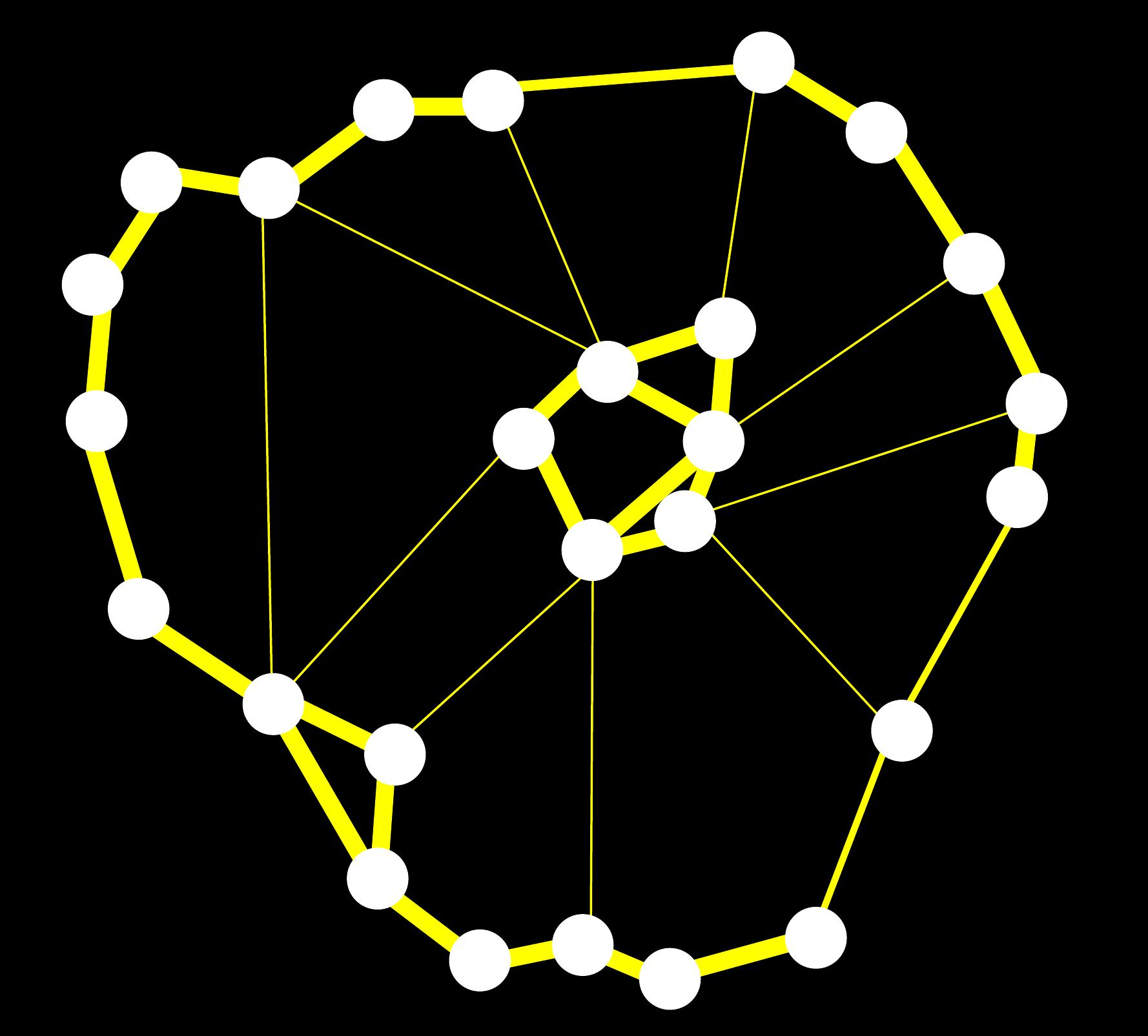}
\hspace{0.5cm}
 \includegraphics[height=2.5truein,width=2.8truein]{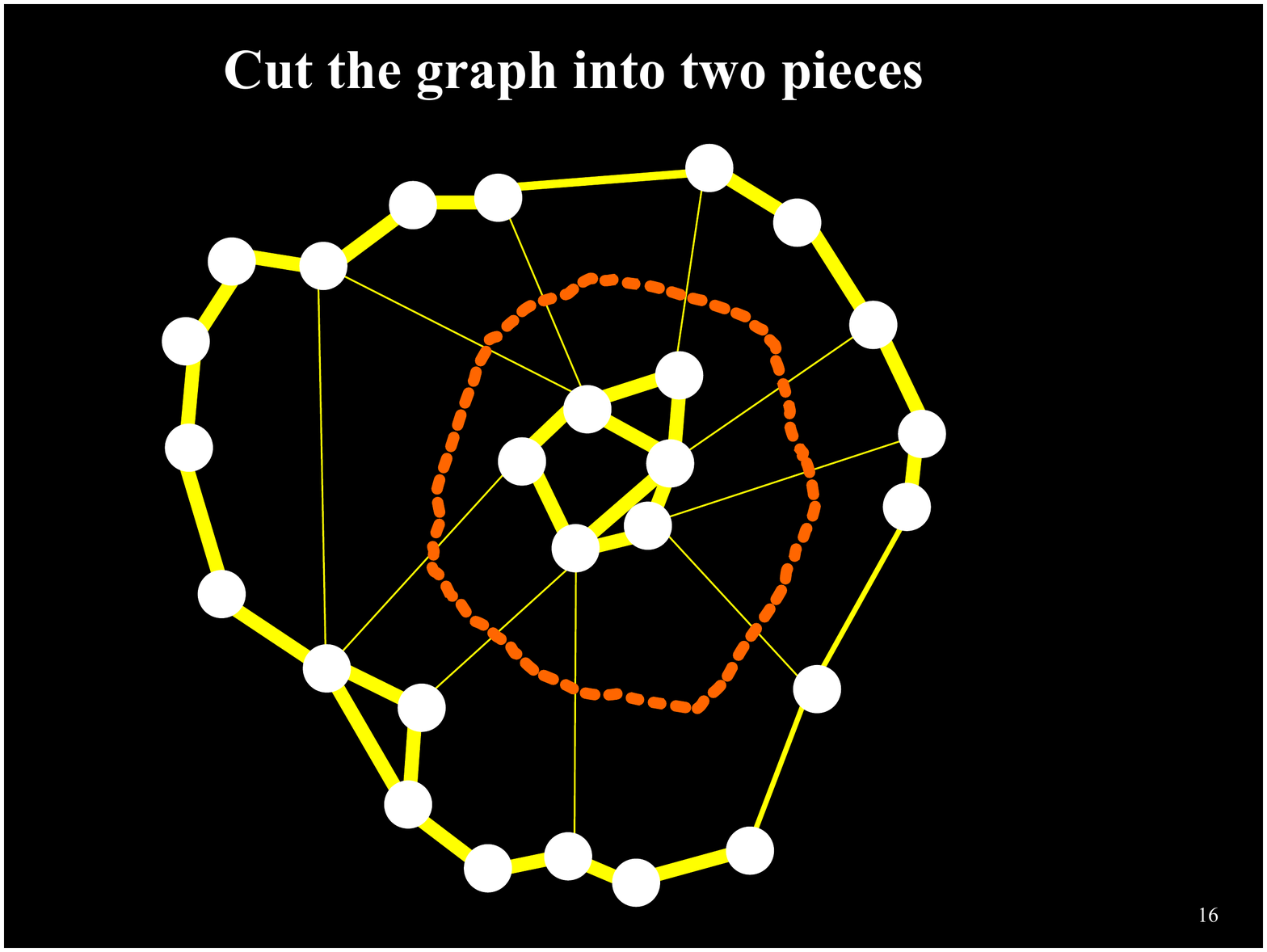}
  \end{center}
  \caption{A weighted graph and its partition into two clusters.}
\label{ncg-fig4a}
\end{figure}

\medskip
A example of a weighted graph and a partition of
its nodes into two clusters is shown in Figure \ref{ncg-fig4a}.

\medskip
A way to get around this problem is to normalize the cuts by dividing
by some measure of each subset $A_i$. 
A solution using the volume $\mathrm{vol}(A_i)$ of $A_i$
(for $K = 2$) was proposed and
investigated in a seminal paper of  Shi and Malik \cite{ShiMalik}.
Subsequently, Yu (in her dissertation \cite{Yu}) and Yu and Shi
\cite{YuShi2003} extended the method to $K > 2$ clusters.
The idea is  to minimize the cost function
\[
\mathrm{Ncut}(A_1, \ldots, A_K) = 
 \sum_{i = 1}^K \frac{\mathrm{links}(A_i, \overline{A_i})}{\mathrm{vol}(A_i)}
= \sum_{i = 1}^K 
\frac{\mathrm{cut}(A_i, \overline{A_i})}{\mathrm{vol}(A_i)}.
\]

\medskip
The first step is to express our  optimization problem in matrix form.
In the case of two clusters, a single vector $X$ can be used to describe the
partition $(A_1, A_2)  = (A, \overline{A})$. We need to choose the structure of this vector
in such a way that 
\[
\mathrm{Ncut}(A, \overline{A}) = \frac{\transpos{X} L X}{\transpos{X} \Degsym X},
\]
where the term on the right-hand side is a Rayleigh ratio.

\medskip
After careful study of the orginal papers, I discovered various
facts that were implicit in these works, but I feel are important to
be pointed out explicitly.

\medskip
First, I realized 
that it is  important to pick a vector representation 
which is invariant under multiplication by a nonzero scalar,
because the Rayleigh ratio is scale-invariant, and it is crucial to take
advantage of this fact to make the denominator go away.
This implies that {\it the solutions $X$ are points in the projective
space $\mathbb{RP}^{N - 1}$\/}. This was my first revelation.

\medskip
Let $N = |V|$ be the number of nodes in the graph $G$.
In view of the desire for a scale-invariant representation,
it is natural to assume that the  vector $X$ is of the form
\[
X = (x_1, \ldots, x_N),
\]
where $x_i \in \{a, b\}$ for $i = 1, \ldots, N$, 
for any two distinct  real numbers $a, b$. 
This is an indicator vector in
the sense that, for $i = 1, \ldots, N$,
\[
x_i =
\begin{cases}
a & \text{if $v_i \in A$} \\
b & \text{if $v_i \notin A$} .
\end{cases}
\]

The  choice $a = +1, b = -1$ is natural, but premature.
The correct interpretation is really to 
view $X$ as a representative of a point 
in the real projective space $\mathbb{RP}^{N-1}$,  
namely the point $\mathbb{P}(X)$ of homogeneous
coordinates $(x_1\co \cdots \co x_N)$.

\medskip
Let  $d = \transpos{\mathbf{1}} \Degsym \mathbf{1}$ and $\alpha =
\mathrm{vol}(A)$.  I prove that
\[
\mathrm{Ncut}(A, \overline{A}) = \frac{\transpos{X} L X}{\transpos{X} \Degsym X}
\]
holds iff the following condition holds:
\begin{equation}
a \alpha + b(d - \alpha) = 0.
\tag{$\dagger$}
\end{equation}
Note that condition $(\dagger)$ applied to a vector $X$ whose
components are $a$ or $b$ is equivalent to the fact that $X$
is orthogonal to $\Degsym \mathbf{1}$, since
\[
\transpos{X} \Degsym \mathbf{1} =  \alpha  a+ (d - \alpha) b,
\]
where $\alpha = \mathrm{vol}(\{v_i\in V \mid x_i = a\})$.

\medskip
If we let
\[
\s{X} = \big\{
(x_1, \ldots, x_N) \mid x_i \in \{a, b\}, \> a, b\in \reals,\> a,
b\not = 0
\big\},
\]
our solution set is
\[
\s{K}  = \big\{
X  \in\s{X}  \mid \transpos{X}  \Degsym\mathbf{1} = 0
\big\}.
\]
Actually, to be perfectly rigorous,  we are looking for solutions in
$\mathbb{RP}^{N-1}$, so our solution set is really
\[
\mathbb{P}(\s{K})  = \big\{
(x_1\co \cdots\co x_N) \in \mathbb{RP}^{N-1}\mid
(x_1, \ldots, x_N) \in \s{K}
\big\}.
\]
Consequently, our minimization problem can be stated as follows:

\medskip\noindent
{\bf Problem PNC1}
\begin{align*}
& \mathrm{minimize}     &  & 
\frac{\transpos{X} L X}{\transpos{X} \Degsym X} & &  &  &\\
& \mathrm{subject\ to} &  &
\transpos{X} \Degsym\mathbf{1} = 0,  & &  X\in \s{X}.      
\end{align*}

It is understood that the solutions are  points $\mathbb{P}(X)$
in $\mathbb{RP}^{N-1}$.

\medskip 
Since the Rayleigh ratio and the constraints 
$\transpos{X}\Degsym\mathbf{1} = 0$ and $X\in \s{X}$ 
are scale-invariant,
we are led to the following formulation of our problem:

\medskip\noindent
{\bf Problem PNC2}
\begin{align*}
& \mathrm{minimize}     &  & 
\transpos{X} L X & &  &  &\\
& \mathrm{subject\ to} &  & \transpos{X} \Degsym X = 1, &&
 \transpos{X} \Degsym\mathbf{1} = 0, && X\in \s{X}.   
\end{align*}

\medskip
Problem PNC2 is equivalent to problem PNC1 in the sense that 
they 
have the same set of minimal solutions as points $\mathbb{P}(X)
\in\mathbb{RP}^{N-1}$ given by their homogenous coordinates $X$.
More precisely, if $X$ is any minimal solution of PNC1, then $X/(\transpos{X} \Degsym
X)^{1/2}$ is a minimal solution of PNC2 (with the same minimal value
for the objective functions), and if $X$ is a minimal solution of
PNC2, then $\lambda X$ is a minimal solution for PNC1 for all
$\lambda\not= 0$ (with the same minimal value
for the objective functions).

\medskip
Now, as in the classical papers, we consider the relaxation of  the
above problem obtained by dropping the condition that $X\in \s{X}$,
and proceed as usual. However,  having found a solution $Z$
to the relaxed problem, we need to find a discrete solution $X$ such
that $d(X, Z)$ is minimum in $\mathbb{RP}^{N-1}$.
All this presented in Section \ref{ch3-sec2}.

\medskip
If the number of clusters $K$ is at least $3$, then we need
to choose  a matrix representation for partitions on the set of vertices.
It is important that such a representation be scale-invariant, and 
it is also necessary to state necessary and sufficient conditions
for such matrices to represent a partition
(to the best of our knowledge, these points are not clearly articulated 
in the literature).

\medskip
We describe a  partition $(A_1, \ldots, A_K)$ of the set of nodes $V$ by
an $N\times K$ matrix  $X = [X^1 \cdots X^K]$
whose columns $X^1, \ldots,
X^K$ are indicator vectors of the partition $(A_1, \ldots, A_K)$.
Inspired by what we did when  $K = 2$, 
we assume that the  vector $X^j$ is of the form
\[
X^j = (x_1^j, \ldots, x_N^j),
\]
where $x_i^j \in \{a_j, b_j\}$ for $j = 1, \ldots, K$ and
$i = 1, \ldots, N$, and
where $a_j, b_j$ are 
any  two distinct  real numbers.
The vector $X^j$  is an indicator vector for $A_j$ in
the sense that, for $i = 1, \ldots, N$,
\[
x_i^j =
\begin{cases}
a_j & \text{if $v_i \in A_j$} \\
b_j & \text{if $v_i \notin A_j$} .
\end{cases}
\]

The choice $\{a_j, b_j\} = \{0, 1\}$ for $j = 1, \ldots, K$ is
natural, but premature. I show that if we pick $b_i = 0$, then we have
\[
\frac{\mathrm{cut}(A_j, \overline{A_j})}{\mathrm{vol}(A_j)} 
= \frac{\transpos{(X^j)} L X^j}{\transpos{(X^j)}\Degsym X^j} \quad j =
1, \ldots, K, 
\]
which  implies that
\[
\mathrm{Ncut}(A_1, \ldots, A_K) 
= \sum_{j = 1}^K 
\frac{\mathrm{cut}(A_j, \overline{A_j})}{\mathrm{vol}(A_j)}
= \sum_{j = 1}^K 
\frac{\transpos{(X^j)} L X^j}{\transpos{(X^j)}\Degsym X^j}.
\]
Then, I give necessary and sufficient conditions for a matrix $X$ to
represent a partition. 

\medskip
If we let
\[
\s{X}  = \Big\{[X^1\> \ldots \> X^K] \mid
X^j = a_j(x_1^j, \ldots, x_N^j) , \>
x_i^j \in \{1, 0\},
 a_j\in \reals, \> X^j \not= 0
\Big\}
\]
(note that the condition $X^j \not= 0$ implies that $a_j \not= 0$),
then the set of matrices representing partitions of $V$ into $K$
blocks is
\begin{align*}
& & &\s{K}  = \Big\{ X = [X^1 \> \cdots \> X^K] \quad \mid & &  X\in\s{X},  &&\\
         & & &  & &  \transpos{(X^i)} \Degsym X^j = 0, \quad 1\leq i, j \leq K,\> 
i\not= j, && \quad\quad\quad\quad\quad\\
&  & & & & 
 X (\transpos{X} X)^{-1} \transpos{X} \mathbf{1} = \mathbf{1}\Big\}. && 
\end{align*}

As in the case $K = 2$, to be rigorous, the {\it solution are really
$K$-tuples of points in $\mathbb{RP}^{N-1}$\/}, so our solution set is
really
\[
\mathbb{P}(\s{K})  = \Big\{(\mathbb{P}(X^1), \ldots, \mathbb{P}(X^K)) \mid
[X^1 \> \cdots \> X^K] \in \s{K} 
\Big\}.
\]

\medskip
In view of the above, we have our first formulation of $K$-way clustering
of a graph using normalized cuts, called problem PNC1 
(the notation PNCX  is used in  Yu \cite{Yu}, Section 2.1):

\medskip\noindent
{\bf $K$-way Clustering of a graph using Normalized Cut, Version 1: \\
Problem PNC1}

\begin{align*}
& \mathrm{minimize}     &  &  \sum_{j = 1}^K 
\frac{\transpos{(X^j)} L X^j}{\transpos{(X^j)}\Degsym X^j}& &  &  &\\
& \mathrm{subject\ to} &  & 
 \transpos{(X^i)} \Degsym X^j = 0, \quad 1\leq i, j \leq K,\> 
i\not= j,  & &  & & \\
& & & 
 X (\transpos{X} X)^{-1} \transpos{X} \mathbf{1} = \mathbf{1},  & & X\in \s{X}. & & 
\end{align*}

As in the case $K = 2$, the solutions that we are seeking are $K$-tuples 
$(\mathbb{P}(X^1), \ldots, \mathbb{P}(X^K))$ of points  in
$\mathbb{RP}^{N-1}$ determined by
their  homogeneous coordinates $X^1, \ldots, X^K$.

\medskip
Then, step by step, we transform problem PNC1 into an equivalent
problem PNC2, which we eventually relax by dropping the condition that
$X\in \s{X}$.

\medskip
Our second revelation is that the relaxation $(*_1)$ of 
version 2 of  our minimization problem (PNC2), which is equivalent to
version 1, reveals that that the solutions of the relaxed problem
$(*_1)$ are members of the {\it Grassmannian\/} $G(K, N)$.

\medskip
This leads us to our third revelation:
{\it we have  two choices of metrics to compare solutions\/}:
(1) a metric on
$(\mathbb{RP}^{N - 1})^K$; (2) a metric on $G(K, N)$.
We  discuss the first choice, which is the choice implicitly
adopted by Shi and Yu.

\medskip
Some of the most technical material on the Rayleigh ratio, which is needed
for some proofs in Chapter \ref{chap2}, is the object of Appendix
\ref{Rayleigh-Ritz}. Appendix \ref{ch3-sec6} may seem a bit out of
place.
Its purpose is to explain how 
to define a metric on the projective space $\mathbb{RP}^n$. For this, we
need to review a few notions of differential geometry. 

\medskip
I hope that these notes will be make it easier for people to
become familiar with the wonderful theory of normalized graph cuts.
As far as I know, except for a short section in one of Gilbert
Strang's book, and  von Luxburg \cite{Luxburg} excellent survey
on spectral clustering, 
there is no comprehensive  writing on the topic
of normalized cuts.

\chapter{Graphs  and Graph Laplacians; Basic Facts}
\label{chap1}
\section[Directed Graphs, Undirected Graphs, Weighted Graphs]
{Directed Graphs, Undirected Graphs, Incidence Matrices,
Adjacency Matrices, Weighted Graphs}
\label{ch1-sec1}%
\begin{definition}
\label{dirgraph}
A {\it directed graph\/} is a pair $G = (V, E)$, where 
$V = \{v_1,  \ldots, v_m\}$ is a set of
{\it nodes\/} or {\it vertices\/}, and $E\subseteq V \times V$ is a
set of ordered pairs of distinct nodes (that is, pairs  
$(u, v)\in V\times V$ with $u\not= v$), called {\it edges\/}.
Given any edge $e = (u, v)$, we let $s(e) = u$ be the {\it source\/}
of $e$ and $t(e) = v$ be the {\it target\/} of $e$.
\end{definition}

\medskip
\remark
Since an edge is a pair $(u, v)$ with $u\not= v$, self-loops are not
allowed.
Also, there is at most one edge from a node $u$ to a node $v$.
Such graphs are sometimes called {\it simple graphs\/}.

\begin{figure}[http]
  \begin{center}
   \includegraphics[height=1.7truein,width=2.3truein]{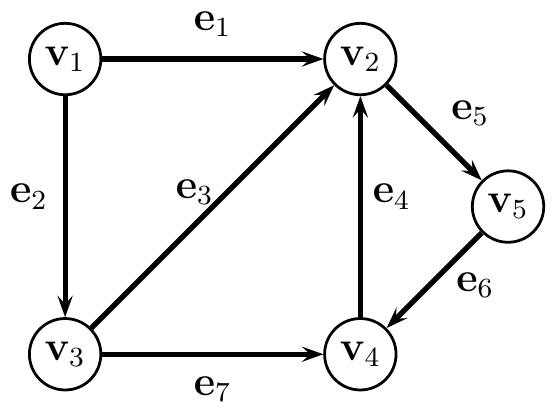}
  \end{center}
  \caption{Graph $G_1$.}
\label{graphfig17}
\end{figure}

\medskip
An example of a directed graph is shown in Figure \ref{graphfig17}.

\medskip
For every node $v\in V$, the {\it degree\/} $d(v)$ of $v$ is the number of
edges leaving or entering $v$:
\[
d(v) = |\{u\in V \mid (v, u)\in E\> \mathrm{or}\>  (u, v)\in E\}|.
\]
The {\it degree matrix\/} $\Degsym(G)$, is the diagonal matrix
\[
\Degsym(G) = \mathrm{diag}(d_1, \ldots, d_m).
\]
For example, for graph $G_1$, we have
\[
\Degsym(G_1) = 
\begin{pmatrix}
2 & 0 & 0 & 0 & 0 \\
0 & 4 & 0 & 0 & 0 \\
0 & 0 & 3 & 0 & 0 \\
0 & 0 & 0 & 3 & 0 \\
0 & 0 & 0 & 0 & 2
\end{pmatrix}.
\]
Unless confusion arises, we write $\Degsym$ instead of $\Degsym(G)$.

\begin{definition}
\label{incidence-matrix1}
Given a directed graph $G = (V, E)$, with $V = \{v_1, \ldots, v_m\}$, if 
$E = \{e_1, \ldots, e_n\}$,  then the {\it incidence matrix\/} $\widetilde{D}(G)$ of $G$ is the $m\times n$
matrix whose entries $\widetilde{d}_{i\, j}$ are given by
\[
\widetilde{d}_{i\, j} = 
\begin{cases}
+1 & \text{if $e_j = (v_i, v_k)$ for some $k$} \\
-1 & \text{if $e_j = (v_k, v_i)$ for some $k$} \\
0 & \text{otherwise}.
\end{cases}
\]
\end{definition}

Here is the incidence matrix of the graph $G_1$:
\[
\widetilde{D} = 
\begin{pmatrix}
1  & 1  & 0  & 0  & 0  & 0  & 0  \\
-1 & 0  & -1 & -1 & 1  & 0  & 0  \\ 
0  & -1 & 1  & 0  & 0  & 0  & 1  \\
0  & 0  & 0  & 1  & 0  & -1 & -1 \\
0  & 0  & 0  & 0  & -1 & 1  & 0
\end{pmatrix}.
\]

\medskip
Again, unless confusion arises, we write $\widetilde{D}$ instead of $\widetilde{D}(G)$.

\medskip
Undirected graphs are obtained from directed graphs by forgetting the
orientation of the edges.

\begin{definition}
\label{graph}
A {\it  graph\/} (or {\it undirected graph\/})
is a pair $G = (V, E)$, where 
$V = \{v_1,  \ldots, v_m\}$ is a set of
{\it nodes\/} or {\it vertices\/}, and $E$ is a
set of two-element subsets of $V$  (that is, subsets  
$\{u, v\}$,  with $u, v\in V$ and $u\not= v$), called {\it edges\/}.
\end{definition}

\medskip
\remark
Since an edge is a set $\{u, v\}$,  we have  $u\not= v$, so self-loops are not
allowed.
Also, for every set of nodes $\{u, v\}$, there
there is at most one edge between $u$ and $v$.
As in the case of directed graphs, 
such graphs are sometimes called {\it simple graphs\/}.

\begin{figure}
  \begin{center}
 \includegraphics[height=1.7truein,width=2.3truein]{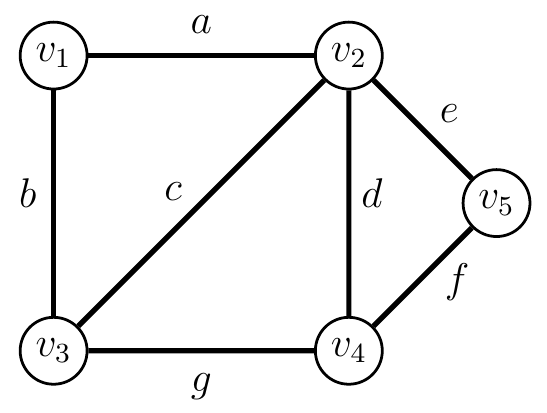}
  \end{center}
  \caption{The undirected graph $G_2$.}
\label{graphfig5bis}
\end{figure}

\medskip
An example of a  graph is shown in Figure \ref{graphfig5bis}.

\medskip
For every node $v\in V$, the {\it degree\/} $d(v)$ of $v$ is the number of
edges adjacent to $v$:
\[
d(v) = |\{u\in V \mid  \{u, v\}\in E\}|.
\]
The degree matrix $\Degsym$ is defined as before.
The notion of incidence matrix for an undirected graph is not as 
useful as the in the case of directed graphs

\begin{definition}
\label{incidence-matrix2}
Given a graph $G = (V, E)$, with $V = \{v_1, \ldots, v_m\}$, if 
$E = \{e_1, \ldots, e_n\}$,  then the {\it incidence matrix\/} $\widetilde{D}(G)$ 
of $G$ is the $m\times n$
matrix whose entries $\widetilde{d}_{i\, j}$ are given by
\[
\widetilde{d}_{i\, j} = 
\begin{cases}
+1 & \text{if $e_j = \{v_i, v_k\}$ for some $k$} \\
0 & \text{otherwise}.
\end{cases}
\]
\end{definition}

Unlike the case of directed graphs, the entries in the incidence
matrix of a graph (undirected) are nonnegative. 
We usally write $\widetilde{D}$ instead of $\widetilde{D}(G)$.

\medskip
The notion of adjacency matrix is basically the same for directed or undirected
graphs.

\begin{definition}
\label{adjacency}
Given a directed or undirected graph $G = (V, E)$, 
with $V = \{v_1, \ldots, v_m\}$,
the {\it adjacency matrix\/}  $A(G)$ of $G$ is the symmetric 
$m\times m$ matrix $(a_{i\, j})$ such that
\begin{enumerate}
\item[(1)] 
If $G$ is directed, then
\[
a_{i\, j} = 
\begin{cases}
1 & \text{if there is some edge $(v_i, v_j)\in E$ or some edge 
$(v_j,  v_i)\in E$} \\
0 & \text{otherwise}. 
\end{cases}
\]
\item[(2)]
Else if $G$ is undirected, then
\[
a_{i\, j} = 
\begin{cases}
1 & \text{if there is some edge $\{v_i, v_j\}\in E $} \\
0 & \text{otherwise}. 
\end{cases}
\]
\end{enumerate}
\end{definition}

\medskip
As usual, unless confusion arises, we write $A$ instead of $A(G)$.
Here is the adjacency matrix of both graphs $G_1$ and $G_2$:

\[
A = 
\begin{pmatrix}
0 & 1 & 1 & 0 & 0 \\
1 & 0 & 1 & 1 & 1 \\
1 & 1 & 0 & 1 & 0 \\
0 & 1 & 1 & 0 & 1 \\
0 & 1 & 0 & 1 & 0
\end{pmatrix}.
\]

\medskip
In many applications, the notion of graph needs to be generalized to
capture the intuitive idea that two nodes $u$ and $v$ are linked with
a degree of certainty (or strength). Thus, we assign a nonnegative weights $w_{i\, j}$ 
to an edge $\{v_i,  v_j\}$; the smaller $w_{i\, j}$ is, the weaker is
the link (or similarity) between $v_i$ and $v_j$, and the greater 
$w_{i\, j}$ is, the stronger is the link (or similarity) between $v_i$
and $v_j$.

\begin{definition}
\label{graph-weighted}
A {\it  weighted graph\/} 
is a pair $G = (V, W)$, where 
$V = \{v_1,  \ldots, v_m\}$ is a set of
{\it nodes\/} or {\it vertices\/}, and $W$ is a symmetric matrix
called the {\it weight matrix\/}, such that $w_{i\, j} \geq 0$
for all $i, j \in \{1, \ldots, m\}$, 
and $w_{i\, i} = 0$ for $i = 1, \ldots, m$.
We say that a set $\{v_i, v_j\}$  is an edge iff
$w_{i\, j} > 0$. The corresponding (undirected) graph $(V, E)$
with $E = \{\{e_i, e_j\} \mid w_{i\, j} > 0\}$, 
is called the {\it underlying graph\/} of $G$.
\end{definition}

\medskip
\remark
Since $w_{i\, i} = 0$, these graphs have no self-loops.
We can think of the matrix $W$ as a generalized adjacency matrix.
The case where $w_{i\, j} \in \{0, 1\}$ is equivalent to the notion
of a graph as in Definition \ref{graph}.

\begin{figure}[http]
  \begin{center}
 \includegraphics[height=2.5truein,width=2.8truein]{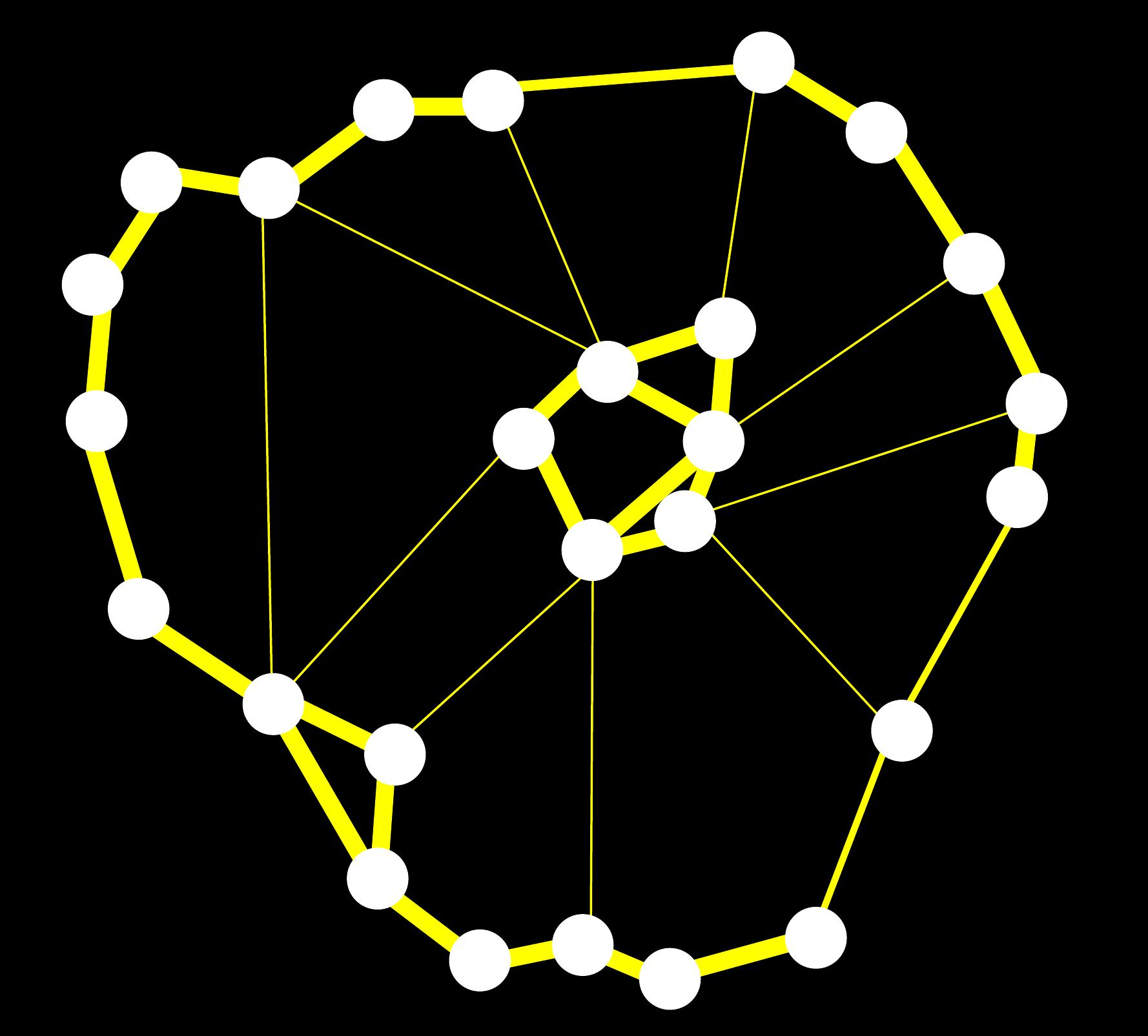}
  \end{center}
  \caption{A weighted graph.}
\label{ncg-fig1}
\end{figure}

\medskip
We can think of the weight $w_{i\, j}$ of an edge $\{v_i, v_j\}$ as a degree of
similarity (or affinity)  in an image, or a cost in a network.
An example of a weighted graph is shown in Figure \ref{ncg-fig1}.
The thickness of the edges corresponds to the magnitude of its weight.

\medskip
For every node $v_i\in V$, the {\it degree\/} $d(v_i)$ of $v_i$ is the sum
of  the weights of the 
edges adjacent to $v_i$:
\[
d(v_i) = \sum_{j = 1}^m w_{i\, j}.
\]
Note that in the above sum, only nodes $v_j$ such that there is an
edge $\{v_i, v_j\}$ have a nonzero contribution. Such nodes are said
to be  {\it adjacent\/} to $v_i$.
The degree matrix $\Degsym$ is defined as before, namely by
$\Degsym = \mathrm{diag}(d(v_1), \ldots, d(v_m))$.

\medskip
Following common practice, we denote by $\mathbf{1}$ the (column)
vector whose components are all equal to $1$.
Then, 
observe that  $W \mathbf{1}$
is the (column) vector $(d(v_1), \ldots, d(v_m))$ consisting of the
degrees of the nodes
of the graph.

\medskip
Given any subset of nodes $A \subseteq V$, we define the 
{\it  volume\/}
$\mathrm{vol}(A)$ of $A$ as the sum of the weights of all edges 
adjacent to nodes in $A$:
\[
\mathrm{vol}(A) = \sum_{v_i\in A} d(v_i) = 
\sum_{v_i \in A} \sum_{j =  1}^m w_{i\, j}.
\]
\remark
Yu and Shi \cite{YuShi2003}
use the notation $\mathrm{degree}(A)$ instead of $\mathrm{vol}(A)$. 

\begin{figure}[http]
  \begin{center}
 \includegraphics[height=2truein,width=2truein]{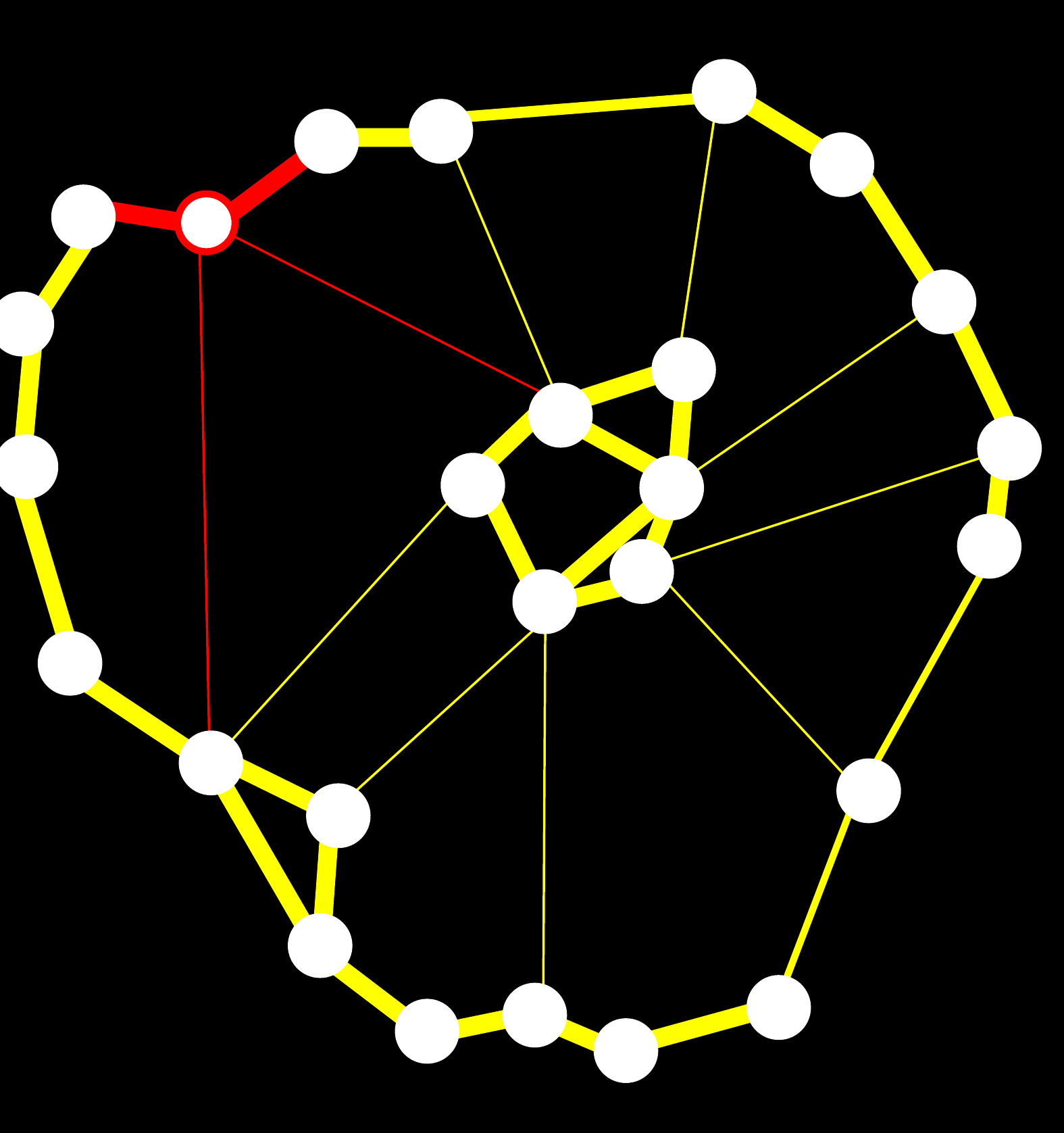}
\hspace{1cm}
 \includegraphics[height=2truein,width=2truein]{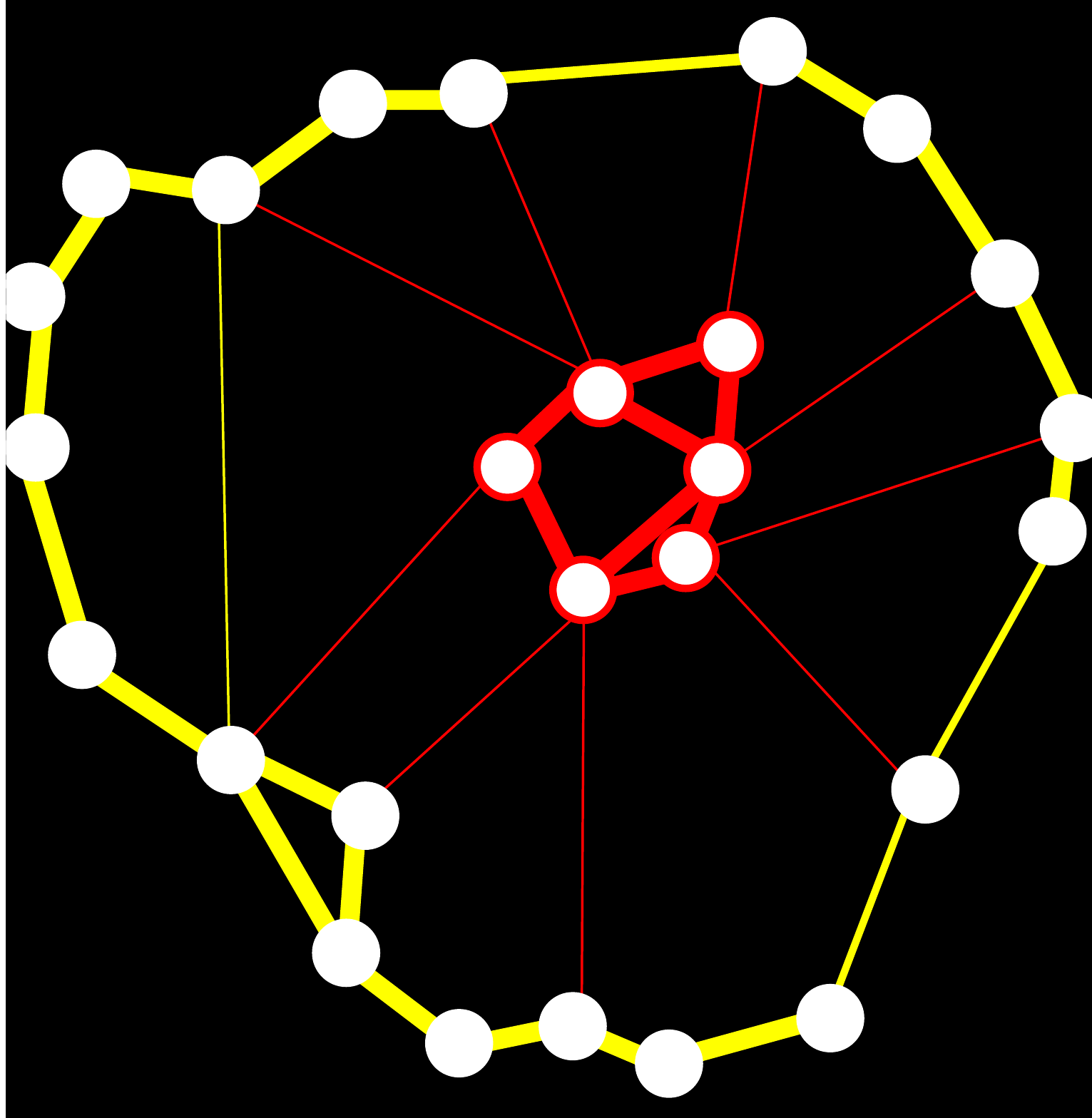}
  \end{center}
  \caption{Degree and volume.}
\label{ncg-fig2}
\end{figure}

The notions of degree and volume are illustrated in Figure \ref{ncg-fig2}.

\medskip
Observe that $\mathrm{vol}(A) = 0$ if $A$ consists of isolated vertices,
that is, if $w_{i\, j} = 0$ for all $v_i\in A$.  Thus, it is best to
assume that $G$ does not have isolated vertices.

\medskip
Given any two subset $A, B\subseteq V$ (not necessarily distinct), we
define
$\mathrm{links}(A, B)$ by
\[
\mathrm{links}(A, B) = \sum_{v_i\in A, v_j\in B} w_{i\, j}.
\]
Since the matrix $W$ is symmetric, we have
\[
\mathrm{links}(A, B) = \mathrm{links}(B, A),  
\]
and observe that
$\mathrm{vol}(A) = \mathrm{links}(A, V)$.

\begin{figure}[http]
  \begin{center}
 \includegraphics[height=2.2truein,width=2.2truein]{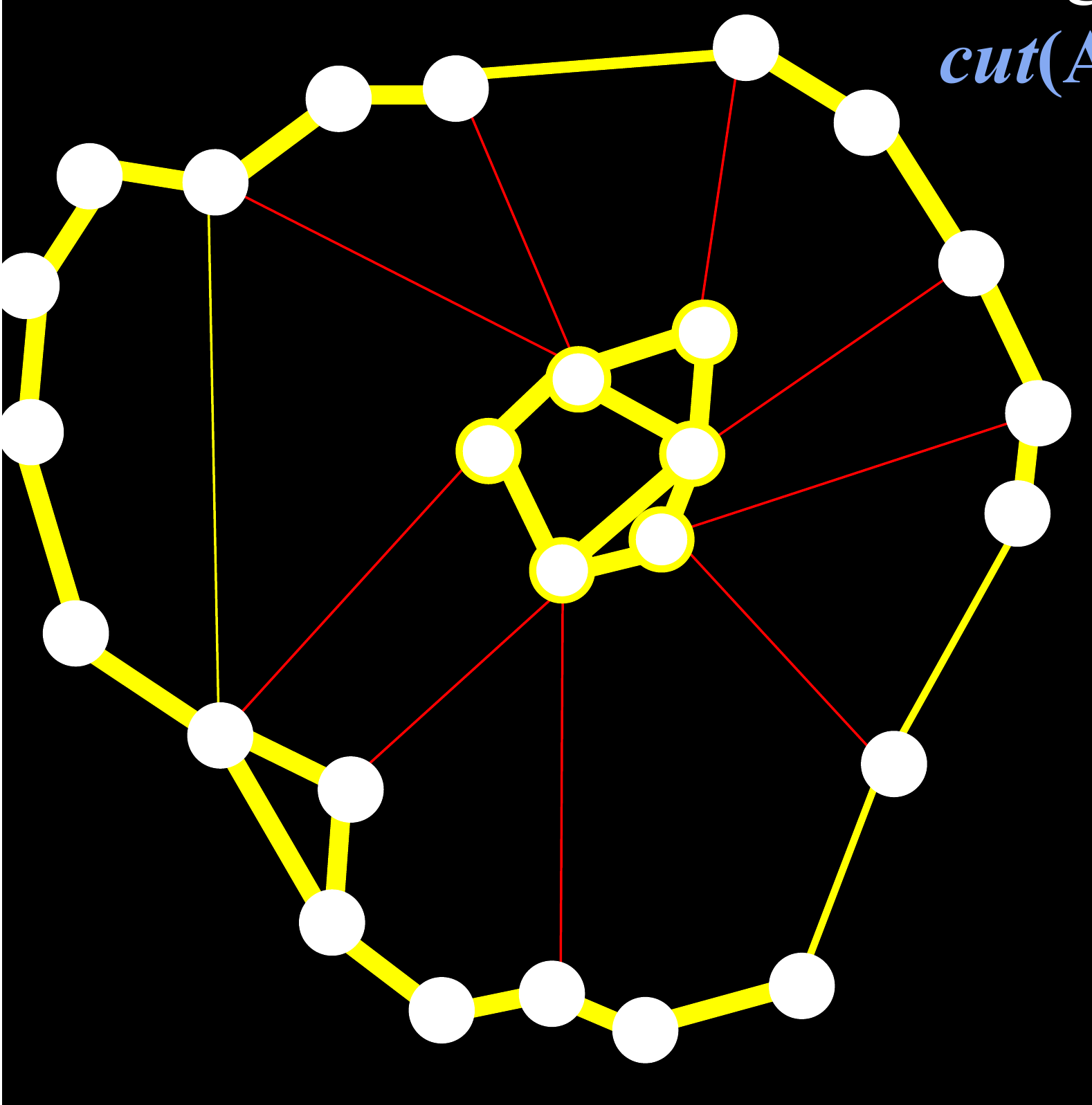}
  \end{center}
  \caption{A Cut involving the set of nodes in the center and the
    nodes on the perimeter.}
\label{ncg-fig3}
\end{figure}

\medskip
The quantity
$ \mathrm{links}(A, \overline{A}) =
\mathrm{links}(\overline{A}, A)$,
where $\overline{A} = V - A$ denotes the complement of $A$ in $V$,
measures
how many links escape from $A$ (and $\overline{A}$), and the quantity
$\mathrm{links}(A,A)$ 
measures how many links stay within $A$ itself.
The quantity  
\[
\mathrm{cut}(A) =
\mathrm{links}(A, \overline{A})
\]
is often called the {\it cut\/} of $A$, and the quantity 
\[
\mathrm{assoc}(A) =
\mathrm{links}(A,A)
\]
is often called the {\it association\/} of $A$.
Clearly,
\[
\mathrm{cut}(A) + \mathrm{assoc}(A) = \mathrm{vol}(A).
\]
The notions of cut is illustrated in Figure \ref{ncg-fig3}.

\medskip
We now define the most important concept of these notes: The Laplacian
matrix of a graph. Actually, as we will see, it comes in several
flavors.

\section{Laplacian Matrices of Graphs}
\label{ch1-sec2}
Let us begin with directed graphs, although as we will see, graph
Laplacians are fundamentally associated with undirected graph.
The key proposition whose proof can be found in Gallier
\cite{GallDiscmath}
and Godsil and Royle \cite{Godsil} is this:

\begin{proposition}
\label{adjp2}
Given any directed graph $G$ if $\widetilde{D}$ is the incidence
matrix of $G$, $A$ is the adjacency matrix of $G$,
and $\Degsym$ is the degree matrix such
that $\Degsym_{i\, i} = d(v_i)$,
then
\[
\widetilde{D}\transpos{\widetilde{D}} = \Degsym - A.
\]
Consequently, $\widetilde{D}\transpos{\widetilde{D}}$ is 
independent of the orientation of $G$ and
$\Degsym - A$ is symmetric,
positive, semidefinite; that is, the eigenvalues of
$\Degsym - A$ are real and nonnegative.
\end{proposition}

\medskip
The matrix $L = \widetilde{D}\transpos{\widetilde{D}} = \Degsym - A$ is called the 
{\it  (unnormalized) graph Laplacian\/} of the graph $G$. For example, the
graph Laplacian of graph $G_1$ is
\[
L = 
\begin{pmatrix}
2 & -1 & -1 & 0 & 0 \\
-1 & 4 & -1 & -1 & -1 \\
-1 & -1 & 3 & -1 & 0 \\
0 & -1 & -1 & 3 & -1 \\
0 & -1 & 0 & -1 & 2
\end{pmatrix}.
\]

\medskip
The {\it  (unnormalized) graph Laplacian\/} of an undirected graph 
$G = (V, E)$ is defined by
\[
L = \Degsym - A.
\]
Since $L$ is equal to $\widetilde{D}\transpos{\widetilde{D}}$ for any orientation of $G$,
it is also positive semidefinite.
Observe that each row of $L$ sums to zero. Consequently, the vector
$\mathbf{1}$ is in the nullspace of $L$.

\remark
With the unoriented version of the incidence matrix
(see Definition \ref{incidence-matrix2}), it can be shown that
\[
\widetilde{D}\transpos{\widetilde{D}} = \Degsym + A.
\]

\medskip
The natural generalization of the notion of graph Laplacian to
weighted graphs is this:

\begin{definition}
\label{graphLaplacian}
Given any weighted directed graph $G = (V, W)$ with
$V = \{v_1, \ldots,v_m\}$, the {\it
  (unnormalized) graph Laplacian $L(G)$ of $G$\/} is defined by
\[
L(G) = \Degsym(G) - W,
\]
where $\Degsym(G) = \mathrm{diag}(d_1, \ldots,d_m)$ is the degree matrix
of $G$ (a diagonal matrix), with
\[
d_i = \sum_{j = 1}^m w_{i \, j}.
\]
As usual, unless confusion arises, we write $L$ instead of $L(G)$. 
\end{definition}

\medskip
It is clear that each row of $L$ sums to $0$, so the vector
$\mathbf{1}$ is the nullspace of $L$, but it is less obvious that $L$ is
positive semidefinite. An easy way to prove this is to evaluate the
quadratic form $\transpos{x} L x$.

\begin{proposition}
\label{Laplace1}
For any  $m\times m$ symmetric matrix $W$, if we let $L = \Degsym - W$
where $D$ is the degree matrix of $W = (w_{i j})$,
then  we have
\[
\transpos{x} L x =
\frac{1}{2}\sum_{i, j = 1}^m w_{i\, j} (x_i - x_j)^2
\quad\mathrm{for\ all}\> x\in \reals^m.
\]
Consequently, $\transpos{x} L x$ does not depend on the
diagonal entries in $W$, and if $w_{i\, j} \geq 0$ for all $i, j\in
\{1, \ldots,m\}$, then $L$ is positive semidefinite.
\end{proposition}
\begin{proof}
We have
\begin{align*}
\transpos{x} L x & = \transpos{x} \Degsym x - \transpos{x} W x \\
& = \sum_{i = 1}^m d_i x_i^2 - \sum_{i, j = 1}^m w_{i\, j} x_i x_j \\
& = \frac{1}{2}\left( \sum_{i = 1}^m d_i x_i^2  
- 2 \sum_{i, j = 1}^m  w_{i\, j} x_i x_j +  \sum_{i = 1}^m d_i x_i^2  
\right) \\
& = 
\frac{1}{2}\sum_{i, j = 1}^m w_{i\, j} (x_i - x_j)^2.
\end{align*}
Obviously, the quantity on the right-hand side does not depend on the
diagonal entries in $W$, and if if $w_{i\, j} \geq 0$ for all $i, j$,
then this quantity is nonnegative.
\end{proof}

\medskip
Proposition \ref{Laplace1} immediately implies the following facts:
For any weighted graph $G = (V, W)$, 
\begin{enumerate}
\item
The eigenvalues  $0 = \lambda_1 \leq \lambda_2 \leq  \ldots \leq 
\lambda_m$ of $L$ are real and nonnegative, and there is an
orthonormal basis of eigenvectors of $L$.
\item
The smallest eigenvalue $\lambda_1$ of $L$ is equal to $0$, and
$\mathbf{1}$ is a corresponding eigenvector.
\end{enumerate}

\medskip
It turns out that the dimension of the nullspace of $L$ (the
eigenspace of $0$)  is equal to the number of  connected components
of the underlying graph of $G$.
  
\begin{proposition}
\label{Laplace2}
Let $G = (V, W)$ be a ,weighted graph. The number $c$ of connected components
$K_1, \ldots, K_c$ of the underlying graph of $G$ is equal to the
dimension of the nullspace of $L$, which is equal to the multiplicity
of the eigenvalue $0$. Furthermore, the nullspace of $L$ has a basis
consisting of indicator vectors of the connected components of $G$,
that is, vectors $(f_1, \ldots, f_m)$
such that $f_j = 1$ iff $v_j\in K_i$ and $f_j  = 0$ otherwise.
\end{proposition}

\begin{proof}
A complete proof can be found in von Luxburg \cite{Luxburg},
and we only  give a sketch of the proof.

\medskip
First, assume that $G$ is connected, so $c = 1$. A nonzero vector $x$
is in the kernel of $L$ iff $L x = 0$, which implies that
\[
\transpos{x} L x = \frac{1}{2}\sum_{i, j = 1}^m w_{i\, j} (x_i - x_j)^2
= 0.
\]
This implies that $x_i = x_j$ whenever $w_{i\, j} > 0$, and thus, 
$x_i = x_j$
whenever nodes $v_i$ and $v_j$ are linked by an edge. By induction,
$x_i = x_j$
whenever there is a path from $v_i$ to $v_j$. Since $G$ is assumed to
be connected, any two nodes are linked by a path, which implies
that $x_i = x_j$ for all $i\not= j$. Therefore, the nullspace of $L$
is spanned by $\mathbf{1}$, which is indeed the indicator vector of
$K_1 = V$, and this nullspace has dimension $1$.

\medskip
Let us now assume that $G$ has $c \geq 2$ connected components.
If so, by renumbering the rows and columns of $W$, we may assume that
$W$ is a block matrix consisting of $c$ blocks, and similarly $L$ is a
block matrix of the form
\[
L =
\begin{pmatrix}
L_1 &        &             &      \\
      & L_2  &             &     \\
      &        & \ddots &     \\
      &        &             & L_c
\end{pmatrix},
\]
where $L_i$ is the graph Laplacian associated with the connected
component $K_i$. By the induction hypothesis, $0$ is an eigenvalue of
multiplicity $1$ for each $L_i$, and so the nullspace of $L$ has
dimension $c$. The rest is left as an exercise (or see  von Luxburg \cite{Luxburg}).
\end{proof}

\medskip
Proposition \ref{Laplace2} implies that if the underlying graph of $G$
is connected, 
then the second eigenvalue, $\lambda_2$, of $L$  is strictly positive.

\medskip
Remarkably, the eigenvalue $\lambda_2$ contains a lot of information
about the graph $G$ (assuming that $G = (V, E)$ is an undirected graph).
This was first discovered by Fiedler in 1973, and
for this reason, $\lambda_2$ is often referred to as the {\it Fiedler
  number\/}.  For more on the properties of the Fiedler number,
see Godsil and Royle \cite{Godsil} (Chapter 13) and Chung \cite{Chung}.
More generally, the spectrum $(0, \lambda_2, \ldots, \lambda_m)$ of
$L$ contains a lot of information about the combinatorial structure of
the graph $G$. Leverage of this information is the object of 
{\it spectral graph theory\/}.

\medskip
It turns out that  normalized variants of the graph Laplacian are
needed, especially in applications to graph clustering. 
These variants make sense only if $G$ has no isolated vertices, which
means that every row of $W$ contains some strictly positive entry.
In this case, the degree matrix $\Degsym$ contains positive entries,
so it is invertible and
$\Degsym^{-1/2}$ makes sense; namely
\[
\Degsym^{-1/2} = \mathrm{diag}(d_1^{-1/2}, \ldots, d_m^{-1/2}),
\]
and similarly for any real exponent $\alpha$.

\begin{definition}
\label{graphLaplacian2}
Given any weighted directed graph $G = (V, W)$ 
with no isolated vertex and with
$V = \{v_1, \ldots,v_m\}$,
the {\it (normalized) graph Laplacians $L_{\mathrm{sym}}$ and
  $L_{\mathrm{rw}}$  of $G$\/} are defined by
\begin{align*}
L_{\mathrm{sym}} & = \Degsym^{-1/2} L \Degsym^{-1/2} = I - \Degsym^{-1/2} W \Degsym^{-1/2}  \\
L_{\mathrm{rw}} & = \Degsym^{-1} L = I - \Degsym^{-1} W.
\end{align*}
\end{definition}

\medskip
Observe that the Laplacian $L_{\mathrm{sym}}  = \Degsym^{-1/2} L \Degsym^{-1/2}$
is a symmetric matrix (because $L$ and $\Degsym^{-1/2}$ are symmetric)
and that
\[
L_{\mathrm{rw}} = \Degsym^{-1/2} L_{\mathrm{sym}} \Degsym^{1/2}.
\]
The reason for the notation $L_{\mathrm{rw}}$ is that this matrix
is closely related to a random walk on the graph $G$.
There are simple relationships between the eigenvalues and the eigenvectors of 
$L_{\mathrm{sym}}$, and $L_{\mathrm{rw}}$. There is also a simple
relationship with  the generalized eigenvalue problem $Lx = \lambda
\Degsym x$.

\begin{proposition}
\label{Laplace3}
Let $G = (V, W)$ be a weighted graph without isolated vertices. The
graph Laplacians, $L, L_{\mathrm{sym}}$, and $L_{\mathrm{rw}}$ satisfy
the following properties:
\begin{enumerate}
\item[(1)]
The matrix $ L_{\mathrm{sym}}$ is symmetric, positive,
semidefinite. In fact, 
\[
\transpos{x} L_{\mathrm{sym}} x =
\frac{1}{2}\sum_{i, j = 1}^m w_{i\, j} \left(\frac{x_i}{\sqrt{d_i}} - \frac{x_j}{\sqrt{d_j}}\right)^2
\quad\mathrm{for\ all}\> x\in \reals^m.
\]
\item[(2)]
The normalized graph Laplacians $L_{\mathrm{sym}}$ and $L_{\mathrm{rw}}$ 
have the same spectrum \\
$(0 = \nu_1 \leq  \nu_2 \leq  \ldots\leq
\nu_m)$, and a vector $u\not= 0$ is an eigenvector of
$L_{\mathrm{rw}}$ for $\lambda$ iff $\Degsym^{1/2} u$ is an eigenvector of
$L_{\mathrm{sym}}$ for $\lambda$.
\item[(3)]
The graph Laplacians, $L, L_{\mathrm{sym}}$, and $L_{\mathrm{rw}}$ are
symmetric, positive, semidefinite.
\item[(4)]
A vector $u\not = 0$ is a solution of the generalized eigenvalue
problem $L u = \lambda \Degsym u$ iff $\Degsym^{1/2} u$ is an eigenvector of
$L_{\mathrm{sym}}$ for the eigenvalue $\lambda$ iff
$u$ is  an eigenvector of
$L_{\mathrm{rw}}$ for the eigenvalue $\lambda$.
 \item[(5)]
The graph Laplacians, $L$ and  $L_{\mathrm{rw}}$ have the same
nullspace.
 \item[(6)]
The vector $\mathbf{1}$ is in the nullspace of $L_{\mathrm{rw}}$, and 
$\Degsym^{1/2} \mathbf{1}$ is in the nullspace of $L_{\mathrm{sym}}$.
\end{enumerate}
\end{proposition}

\begin{proof}
(1)
We have $L_{\mathrm{sym}} = \Degsym^{-1/2} L \Degsym^{-1/2}$, and
$\Degsym^{-1/2}$ is a symmetric invertible matrix (since it is an
invertible diagonal matrix). It is a well-known fact of linear algebra
that if $B$ is an invertible matrix, then
a matrix $S$ is symmetric, positive semidefinite iff 
$B S\transpos{B}$ is  symmetric, positive semidefinite.
Since $L$ is symmetric,  positive semidefinite, so is
$L_{\mathrm{sym}}  = \Degsym^{-1/2} L \Degsym^{-1/2}$. The formula
\[
\transpos{x} L_{\mathrm{sym}} x =
\frac{1}{2}\sum_{i, j = 1}^m w_{i\, j} \left(\frac{x_i}{\sqrt{d_i}} - \frac{x_j}{\sqrt{d_j}}\right)^2
\quad\mathrm{for\ all}\> x\in \reals^m
\]
follows immediately from Proposition \ref{Laplace1} by replacing $x$
by $\Degsym^{-1/2} x$, and also shows that $L_{\mathrm{sym}}$ is
positive semidefinite.

\medskip
(2)
Since 
\[
L_{\mathrm{rw}}= \Degsym^{-1/2} L_{\mathrm{sym}} \Degsym^{1/2},
\]
the matrices $L_{\mathrm{sym}}$ and $L_{\mathrm{rw}}$ are similar,
which implies that they have the same spectrum. In fact, since 
$\Degsym^{1/2}$ is invertible, 
\[
L_{\mathrm{rw}} u =
\Degsym^{-1} L u = \lambda u
\]
iff
\[
\Degsym^{-1/2} L u = \lambda \Degsym^{1/2} u
\]
iff
\[
\Degsym^{-1/2} L \Degsym^{-1/2} \Degsym^{1/2} u =  L_{\mathrm{sym}}
\Degsym^{1/2} u = \lambda \Degsym^{1/2} u,
\]
which shows that a vector $u\not= 0$ is an eigenvector of
$L_{\mathrm{rw}}$ for $\lambda$ iff $\Degsym^{1/2} u$ is an eigenvector of
$L_{\mathrm{sym}}$ for $\lambda$.

\medskip
(3)
We already know that $L$ and $L_{\mathrm{sym}}$ are positive
semidefinite,
and (2) shows that $L_{\mathrm{rw}}$ is also positive semidefinite. 

\medskip
(4)
Since $\Degsym^{-1/2}$ is invertible, we have
\[
Lu = \lambda \Degsym u
\]
iff
\[
\Degsym^{-1/2} Lu = \lambda \Degsym^{1/2} u
\]
iff
\[
\Degsym^{-1/2} L\Degsym^{-1/2} \Degsym^{1/2} u = L_{\mathrm{sym}}
\Degsym^{1/2} u = \lambda \Degsym^{1/2} u,
\]
which shows that a vector $u\not = 0$ is a solution of the generalized eigenvalue
problem $L u = \lambda \Degsym u$ iff $\Degsym^{1/2} u$ is an eigenvector of
$L_{\mathrm{sym}}$ for the eigenvalue $\lambda$. The second part of
the statement follows from (2).

\medskip
(5)
Since  $\Degsym^{-1}$ is invertible, we have $L u = 0$ iff 
$\Degsym^{-1}u = L_{\mathrm{rw}} u = 0$.

\medskip
(6)
Since $L\mathbf{1} = 0$, we get  $L_{\mathrm{rw}} u = \Degsym^{-1} L
\mathbf{1} = 0$. That $\Degsym^{1/2} \mathbf{1}$ is in the nullspace of
$L_{\mathrm{sym}}$
follows from (2).
\end{proof}

\medskip
A version of Proposition \ref{Laplace4} also holds for the graph Laplacians
$L_{\mathrm{sym}}$ and $L_{\mathrm{rw}}$.  The proof is left as an exercise.

\begin{proposition}
\label{Laplace4}
Let $G = (V, W)$ be a weighted graph. The number $c$ of connected components
$K_1, \ldots, K_c$ of the underlying graph of $G$ is equal to the
dimension of the nullspace of both  $L_{\mathrm{sym}}$ and
$L_{\mathrm{rw}}$,  which is equal to the multiplicity
of the eigenvalue $0$. Furthermore, the nullspace of $L_{\mathrm{rw}}$ has a basis
consisting of indicator vectors of the connected components of $G$,
that is, vectors $(f_1, \ldots, f_m)$
such that $f_j = 1$ iff $v_j\in K_i$ and $f_j  = 0$ otherwise.
For  $L_{\mathrm{sym}}$, a basis of the nullpace is obtained by
multipying the above basis of the nullspace of $L_{\mathrm{rw}}$ by $\Degsym^{1/2}$.
\end{proposition}

\chapter{Spectral Graph  Drawing}
\label{chap2}
\section{Graph  Drawing and Energy Minimization}
\label{ch2-sec1}
Let $G = (V, E)$ be some undirected graph. It is often desirable to
draw a graph, usually in the plane but possibly in 3D, and it turns
out that the  graph Laplacian can be used to design 
surprisingly good methods. Say $|V| = m$.
The idea is to assign a point $\rho(v_i)$ in $\reals^n$ to the vertex $v_i\in
V$, for every $v_i \in V$, 
and to draw a line segment between the points $\rho(v_i)$ and
$\rho(v_j)$.  Thus, a {\it graph drawing\/} is a function
$\mapdef{\rho}{V}{\reals^n}$.

\medskip
We define the  {\it matrix of a graph drawing $\rho$ 
(in  $\reals^n$)\/}  as a $m \times n$ matrix $R$ whose $i$th row consists
of the row vector $\rho(v_i)$ corresponding to the point representing $v_i$ in
$\reals^n$.
Typically, we want $n < m$; in fact $n$ should be much smaller than $m$.
A representation is {\it balanced\/} iff the sum of the entries of
every column is zero, that is,
\[
\transpos{\mathbf{1}} R = 0.
\]
If a representation is not balanced, it can be made balanced 
by a suitable translation.  We may also assume that the columns of $R$
are linearly independent, since any basis of the column space also
determines the drawing.  Thus, from now on,  we may assume that $n \leq m$.

\medskip
\remark
A graph drawing $\mapdef{\rho}{V}{\reals^n}$ is not required to be injective,
which may result in degenerate drawings where distinct vertices are
drawn as the same point. For this reason, we prefer not to use the
terminology
{\it graph embedding\/}, which is often used in the literature.  This
is because in differential geometry, an embedding always refers to  an injective map.
The term {\it graph immersion\/} would be more appropriate.

\medskip
As explained in Godsil and Royle \cite{Godsil}, we can imagine building
a physical model of $G$ by connecting adjacent vertices (in
$\reals^n$) by identical
springs.
Then, it is natural to consider a representation to be better if it
requires the springs to be less extended. We can formalize this by
defining the {\it energy\/} of a drawing $R$ by
\[
\s{E}(R) = \sum_{\{v_i, v_j\}\in E} \norme{\rho(v_i) - \rho(v_j)}^2,
\]
where $\rho(v_i)$ is the $i$th row of $R$ and 
$\norme{\rho(v_i) - \rho(v_j)}^2$
is the square of the Euclidean length of the line segment
joining $\rho(v_i)$  and  $\rho(v_j)$.

\medskip
Then, ``good drawings''  are drawings that minimize the energy function
$\s{E}$.
Of course, the trivial representation corresponding to the zero matrix
is optimum, so we need to impose extra constraints to rule out the
trivial solution.

\medskip
We can consider the more general situation where the springs are not
necessarily identical. This can be modeled by a symmetric weight (or
stiffness)  matrix $W = (w_{i j})$, with $w_{i j} \geq
0$.
Then our energy function becomes
\[
\s{E}(R) = \sum_{\{v_i, v_j\}\in E} w_{i j} \norme{\rho(v_i) - \rho(v_j)}^2.
\]
It turns out that this function can be expressed in terms of the
matrix $R$ and a diagonal matrix $\widehat{W}$ obtained from $W$.
Let $p = |E|$ be the number of edges in $E$ and pick any enumeration
of these edges, so that every edge $\{v_i, v_j\}$ is uniquely
represented by some index $e$.
Then, let $\widehat{W}$ be the $p\times p$ diagonal  matrix such that
\[
\widehat{w}_{e e} = w_{i j},\quad \text{where $e$ correspond to the edge
  $\{v_i, v_j\}$}.
\]
We have the following proposition from 
Godsil and Royle \cite{Godsil}.

\begin{proposition}
\label{energyprop1}
Let $G = (V, E)$ be an undirected graph, with $|V| = m$, $|E| = p$, 
let $W$ be a $m\times m$ weight matrix, and let $R$ be the
matrix of a graph drawing $\rho$ of  $G$ in $\reals^n$ 
(a $m\times n$ matrix).
If $\widetilde{D}$ is the incidence matrix 
associated with any orientation of the
graph $G$, and $\widehat{W}$ is the $p\times p$ diagonal matrix associated with
$W$, then
\[
\s{E}(R) = \mathrm{tr}(\transpos{R}\widetilde{D} \widehat{W} \transpos{\widetilde{D}} R).
\]
\end{proposition}

\begin{proof}
Observe that the rows of $\transpos{\widetilde{D}} R$ are indexed by the edges of
$G$, and if $\{v_i, v_j\}\in E$, then the $e$th row of $\transpos{\widetilde{D}}
  R$ is
\[
\pm(\rho(v_i) - \rho(v_j)),
\]
where $e$ is the index corresponding  to the edge $\{v_i, v_j\}$.  As
a consequence, the diagonal entries of $\transpos{\widetilde{D}} R \transpos{R} \widetilde{D}$
have the form $\norme{\rho(v_i) - \rho(v_j)}^2$, where $\{v_i, v_j\}$
ranges over the edges in $E$.  Hence,
\[
\s{E}(R) = \mathrm{tr}(\widehat{W} \transpos{\widetilde{D}} R \transpos{R} \widetilde{D}) = 
\mathrm{tr}(\transpos{R}\widetilde{D} \widehat{W} \transpos{\widetilde{D}} R),
\]
since $\mathrm{tr}(AB) =   \mathrm{tr}(BA)$ for any two matrices $A$
and $B$.
\end{proof}

\medskip
The matrix
\[
L = \widetilde{D} \widehat{W} \transpos{\widetilde{D}} 
\]
may be viewed as a weighted Laplacian of $G$. Observe that $L$ is a
$m\times m$ matrix, and that 
\[
L_{i j} = 
\begin{cases}
- w_{i j} & \text{if $i \not= j$} \\
\sum_{\{v_i, v_k\}\in E} w_{i k} & \text{if $i = j$}.
\end{cases}
\]
Therefore, 
\[
L = \Degsym - W,
\]
the familiar unnormalized Laplacian matrix associated with $W$,
where $\Degsym$ is the degree matrix associated with $W$, and so
\[
\s{E}(R) = \mathrm{tr}(\transpos{R} L R).
\]
Note that
\[
L \mathbf{1} = 0,
\]
as we already observed. 

\medskip
Since the matrix  $\transpos{R}\widetilde{D} \widehat{W} \transpos{\widetilde{D}} R = \transpos{R} L R $ is
symmetric, it has real eigenvalues. Actually, since $L = \widetilde{D} \widehat{W} \transpos{\widetilde{D}}$ is
positive semidefinite, so is $\transpos{R} L R$.
Then, the trace of $\transpos{R} L  R$ is
equal to the sum of its positive eigenvalues, and this is the
energy $\s{E}(R)$ of the graph drawing.

\medskip
If $R$ is the matrix of a graph drawing in $\reals^n$, then for any
invertible matrix $M$, the map that assigns $v_i$ to $\rho(v_i)M$ is
another graph drawing of $G$, and these two drawings convey
the same amount of information. From this point of view, a graph
drawing is determined by the column space of $R$. Therefore, it is
reasonable to assume that the columns of $R$ are pairwise orthogonal
and that they have unit length. Such a matrix satisfies the equation
$\transpos{R} R = I$, and the corresponding drawing is called an
{\it orthogonal drawing\/}. This condition also rules out trivial
drawings.
The following result tells us how to find minimum energy graph
drawings, provided the graph is connected.

\begin{theorem}
\label{graphdraw}
Let $G = (V, W)$ be a weigted graph with $|V| = m$.
If $L = \Degsym - W$ is the (unnormalized) 
Laplacian of  $G$, and if the eigenvalues of $L$ are
$0 = \lambda_1 < \lambda_2 \leq \lambda_3 \leq \ldots \leq \lambda_m$, 
then the minimal energy of any balanced orthogonal graph drawing of
$G$ in $\reals^n$ is equal to $\lambda_2 + \cdots + \lambda_{n + 1}$
(in particular, this implies that $n < m$). The $m \times n$ matrix 
$R$ consisting of any unit eigenvectors $u_2, \ldots, u_{n+1}$
associated with $\lambda_2 \leq \ldots \leq \lambda_{n + 1}$ 
yields an orthogonal graph drawing of minimal energy; it satisfies the
condition $\transpos{R} R = I$.
\end{theorem}

\begin{proof}
We present the  proof  given in  Godsil and Royle \cite{Godsil} (Section 13.4,
Theorem 13.4.1). The key point is that
the sum of the  $n$ smallest eigenvalues of $L$  is a lower bound for
$\mathrm{tr}(\transpos{R} L R)$. This can be shown using an argument
using the Rayleigh ratio; see Proposition \ref{interlace}.
Then, any $n$ eigenvectors $(u_1, \ldots,
u_n)$ associated with
$\lambda_1, \ldots, \lambda_n$ achieve this bound.
Because the first eigenvalue of $L$ is
$\lambda_1 = 0$ and because we are assuming that $\lambda_2 > 0$, 
we have $u_1 = \mathbf{1}/\sqrt{m}$, and by deleting $u_1$ we obtain a
balanced orthogonal graph drawing in $\reals^{n - 1}$ with the  same
energy.
The converse is true, so the minimum energy of an orthogonal graph
drawing in $\reals^n$ is equal to the minimum energy of an orthogonal
graph drawing in $\reals^{n+1}$, and this minimum is
$\lambda_2 + \cdots + \lambda_{n + 1}$. The rest is clear.
\end{proof}

\medskip
Observe that for any orthogonal $n\times n$ matrix $Q$, since
\[
\mathrm{tr}(\transpos{R} L R) = \mathrm{tr}(\transpos{Q}\transpos{R} L
R Q),
\]
the matrix $RQ$ also yields a minimum orthogonal graph drawing. 

\medskip
In summary, if $\lambda_2 > 0$, an automatic method for drawing a
graph in $\reals^2$ is this:

\begin{enumerate}
\item
Compute the two smallest nonzero eigenvalues $\lambda_2 \leq \lambda_3$ of 
the graph Laplacian $L$ (it is possible that
$\lambda_3 = \lambda_2$ if $\lambda_2$ is a multiple eigenvalue);
\item
Compute two unit eigenvectors $u_2, u_3$ associated
with $\lambda_2$ and $\lambda_3$, and let $R = [u_2\> u_3]$
be the $m\times 2$ matrix having $u_2$ and $u_3$ as columns.
\item
Place vertex $v_i$ at the point whose coordinates is the $i$th row of
$R$, that is, $(R_{i 1}, R_{i 2})$.
\end{enumerate}

\medskip
This method generally gives pleasing results, but beware that
there is no guarantee that distinct nodes are assigned distinct images,
because $R$ can have identical rows. This does not seem to happen
often in practice.

\section{Examples of Graph  Drawings}
\label{ch2-sec2}
We now give a number of examples using {\tt Matlab}. Some of these are
borrowed or adapted from Spielman \cite{Spielman}.

\begin{figure}[http]
  \begin{center}
   \includegraphics[height=2.2truein,width=2.4truein]{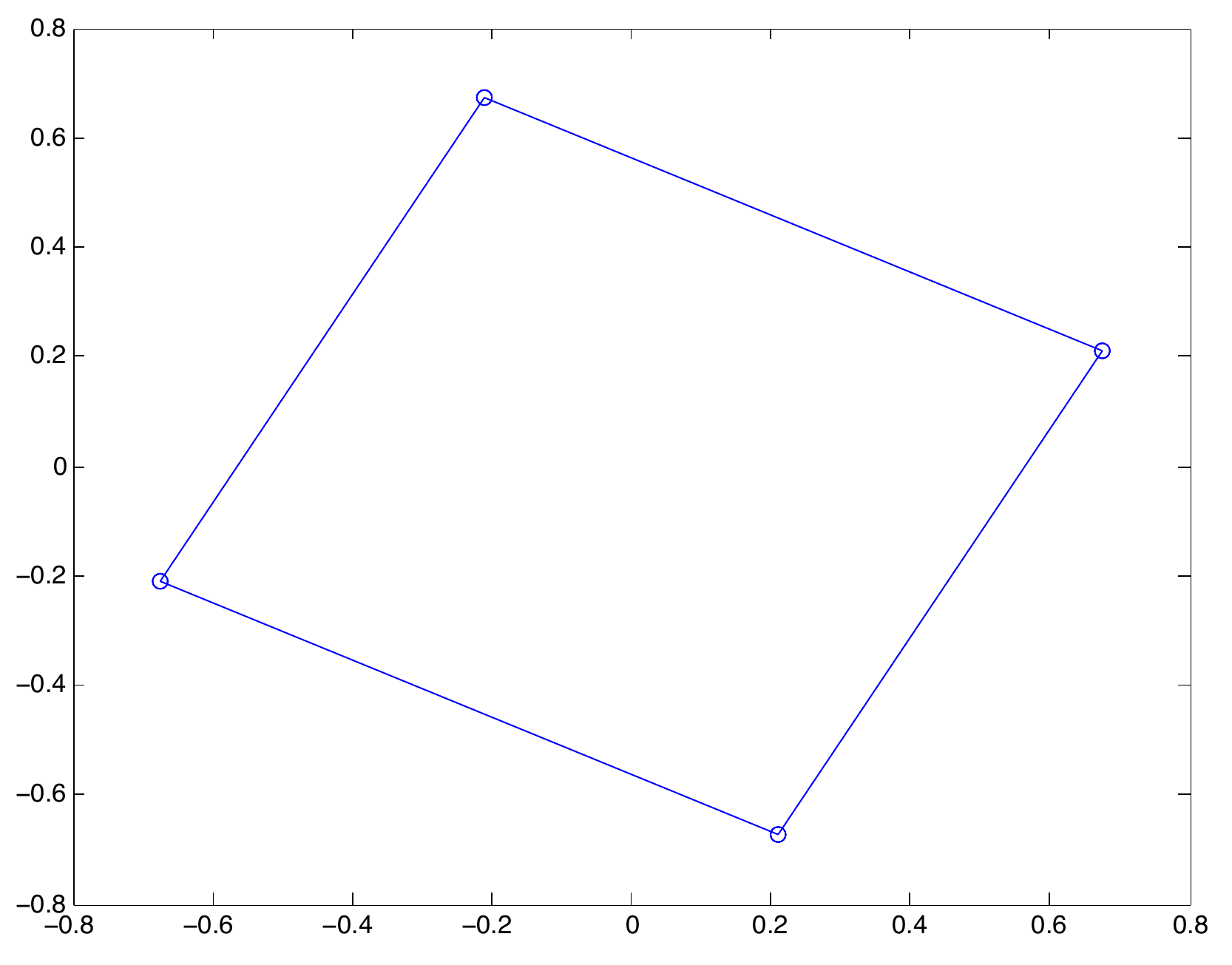}
  \end{center}
  \caption{Drawing of the graph from Example 1.}
\label{graph1}
\end{figure}

\medskip\noindent
{\it Example\/} 1.
Consider the graph with four nodes whose  adjacency matrix is
\[
A = 
\begin{pmatrix}
 0 & 1 & 1 & 0 \\
1 & 0 & 0 & 1 \\
1 & 0 & 0 & 1 \\
0 & 1 & 1 & 0
\end{pmatrix}.
\]
We use the following program to compute $u_2$ and $u_3$:
\begin{verbatim}
A = [0 1 1 0; 1 0 0 1; 1 0 0 1; 0 1 1 0];
D = diag(sum(A));
L = D - A;
[v, e] = eigs(L);
gplot(A, v(:,[3 2]))
hold on;
gplot(A, v(:,[3 2]),'o')
\end{verbatim}

The graph of Example 1 is shown in  Figure \ref{graph1}.
The function  {\tt eigs(L)}  computes the six largest eigenvalues of $L$ in decreasing
order, and corresponding eigenvectors. It turns out that $\lambda_2 =
\lambda_3 = 2$ is a double eigenvalue.

\begin{figure}[http]
  \begin{center}
   \includegraphics[height=2.2truein,width=2.4truein]{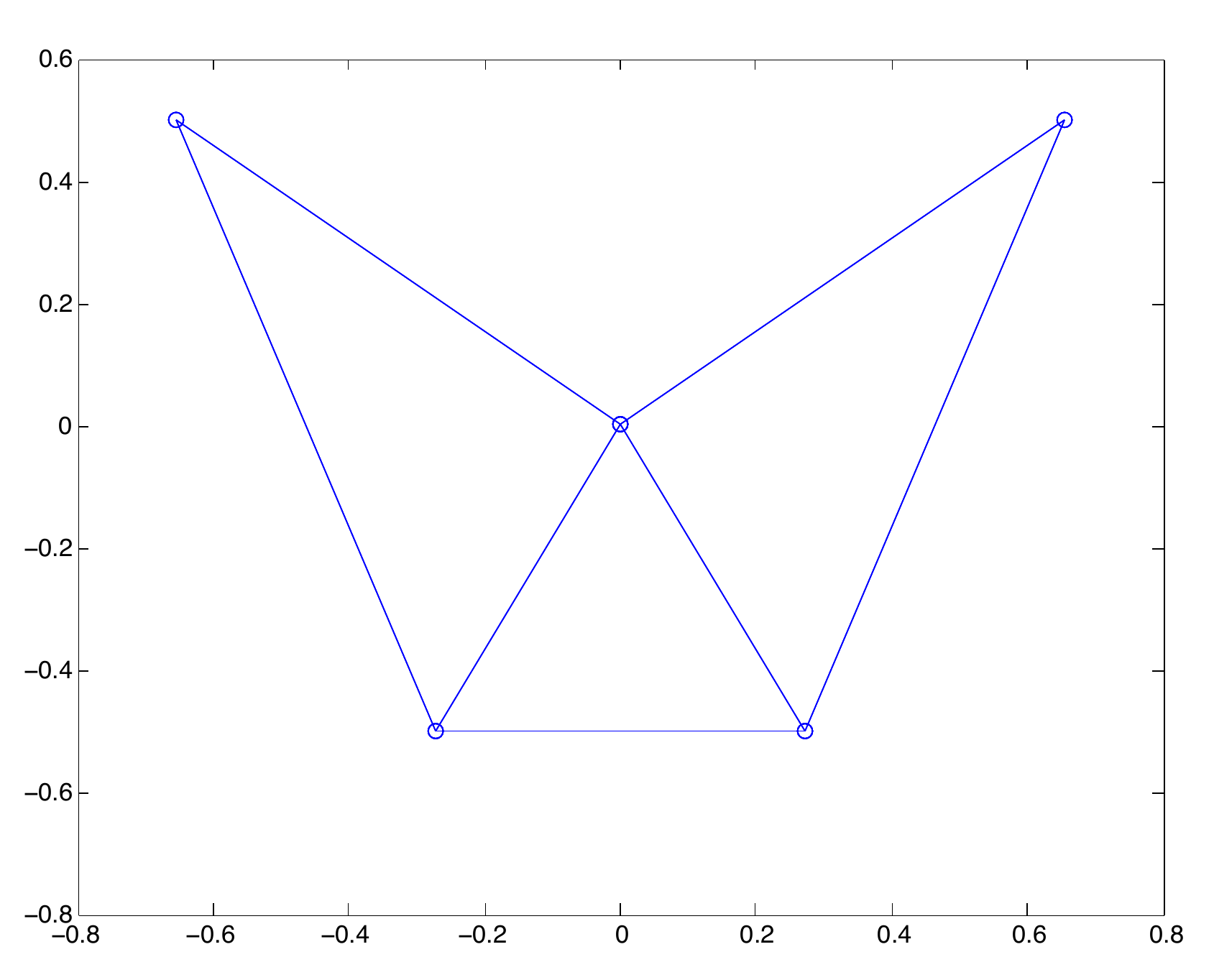}
  \end{center}
  \caption{Drawing of the graph from Example 2.}
\label{graph1b}
\end{figure}

\medskip\noindent
{\it Example\/} 2.
Consider the  graph $G_2$ shown in Figure \ref{graphfig5bis} given 
by the adjacency matrix 
\[
A = 
\begin{pmatrix}
0 & 1 & 1 & 0 & 0 \\
1 & 0 & 1 & 1 & 1 \\
1 & 1 & 0 & 1 & 0 \\
0 & 1 & 1 & 0 & 1 \\
0 & 1 & 0 & 1 & 0
\end{pmatrix}.
\]
We use the following program to compute $u_2$ and $u_3$:

\begin{verbatim}
A = [0 1 1 0 0; 1 0 1 1 1; 1 1 0 1 0; 0 1 1 0 1; 0 1 0 1 0];
D = diag(sum(A));
L = D - A;
[v, e] = eig(L);
gplot(A, v(:, [2 3]))
hold on
gplot(A, v(:, [2 3]),'o')
\end{verbatim}

The function  {\tt eig(L)}  (with no {\tt s} at the end) computes the  eigenvalues of $L$ in increasing
order. The result of drawing the graph is shown in Figure \ref{graph1b}.
Note that node $v_2$ is
assigned to the point $(0, 0)$, so the difference between this drawing
and the drawing  in Figure \ref{graphfig5bis}  is that the
drawing of Figure \ref{graph1b} is not convex.

\medskip\noindent
{\it Example\/} 3.
Consider the ring graph defined by the adjacency matrix $A$ given in
the {\tt Matlab} program shown below:

\begin{verbatim}
A = diag(ones(1, 11),1);
A = A + A';
A(1, 12) = 1; A(12, 1) = 1;
D = diag(sum(A));
L = D - A;
[v, e] = eig(L);
gplot(A, v(:, [2 3]))
hold on
gplot(A, v(:, [2 3]),'o')
\end{verbatim}

\begin{figure}[http]
  \begin{center}
   \includegraphics[height=2.2truein,width=2.4truein]{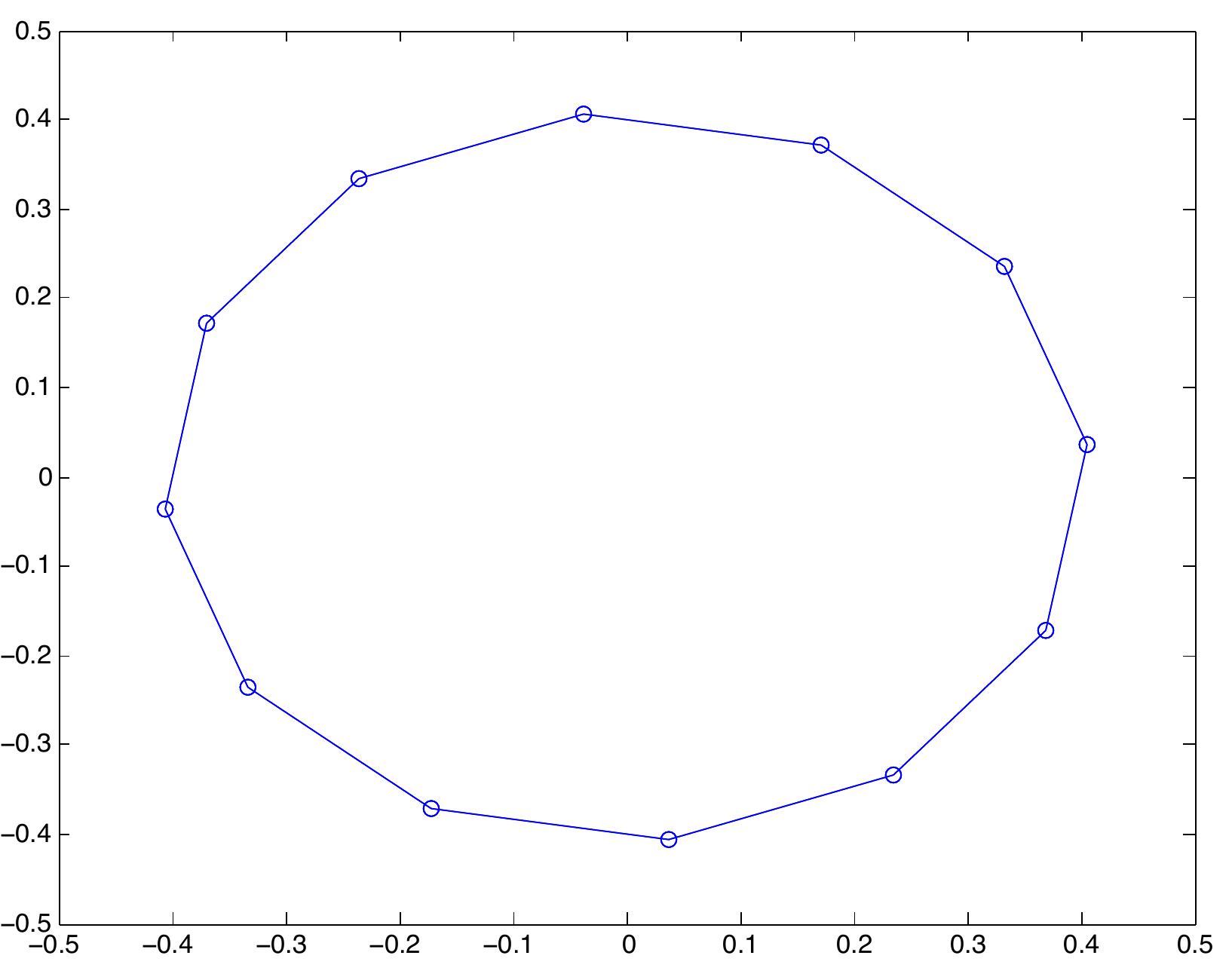}
  \end{center}
  \caption{Drawing of the graph from Example 3.}
\label{graph2}
\end{figure}

Observe that we get a very nice ring; see Figure \ref{graph2}.
Again $\lambda_2 =
0.2679$ is a double eigenvalue (and so are the next pairs of
eigenvalues, except the last, $\lambda_{12} = 4$).

\medskip\noindent
{\it Example\/} 4.
In this example adpated from Spielman, we generate $20$ randomly
chosen points in the unit square, compute their Delaunay
triangulation, then the adjacency matrix of the corresponding graph,
and finally draw the graph using the second and third eigenvalues of
the Laplacian.

\begin{verbatim}
A = zeros(20,20);
xy = rand(20, 2); 
trigs = delaunay(xy(:,1), xy(:,2));
elemtrig = ones(3) - eye(3);
for i = 1:length(trigs),
 A(trigs(i,:),trigs(i,:)) = elemtrig;
end
A = double(A >0);
gplot(A,xy)
D = diag(sum(A));
L = D - A;
[v, e] = eigs(L, 3, 'sm');
figure(2)
gplot(A, v(:, [2 1]))
hold on
gplot(A, v(:, [2 1]),'o')
\end{verbatim}

\begin{figure}[http]
  \begin{center}
   \includegraphics[height=2.2truein,width=2.4truein]{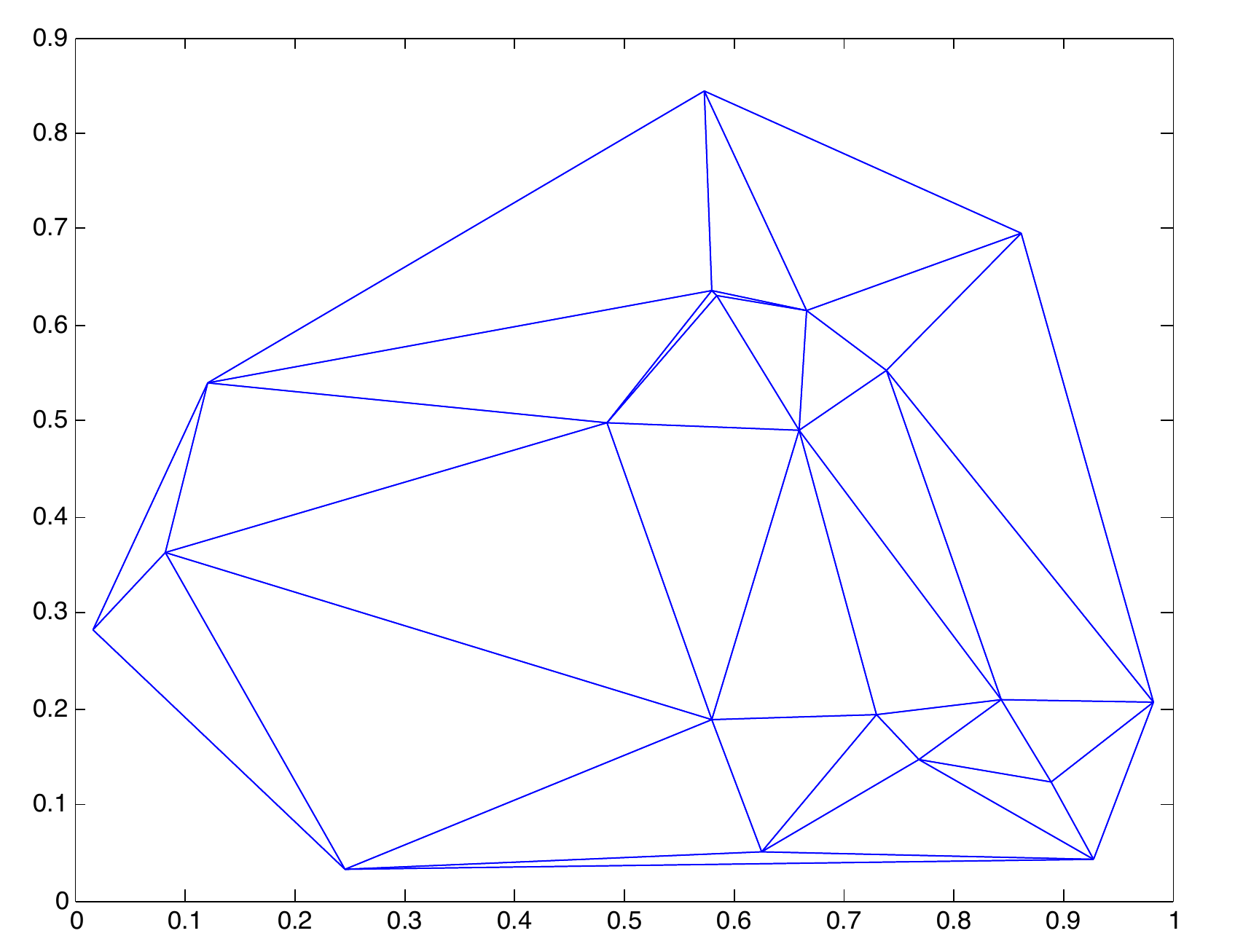}
\hspace{1cm}
   \includegraphics[height=2.2truein,width=2.4truein]{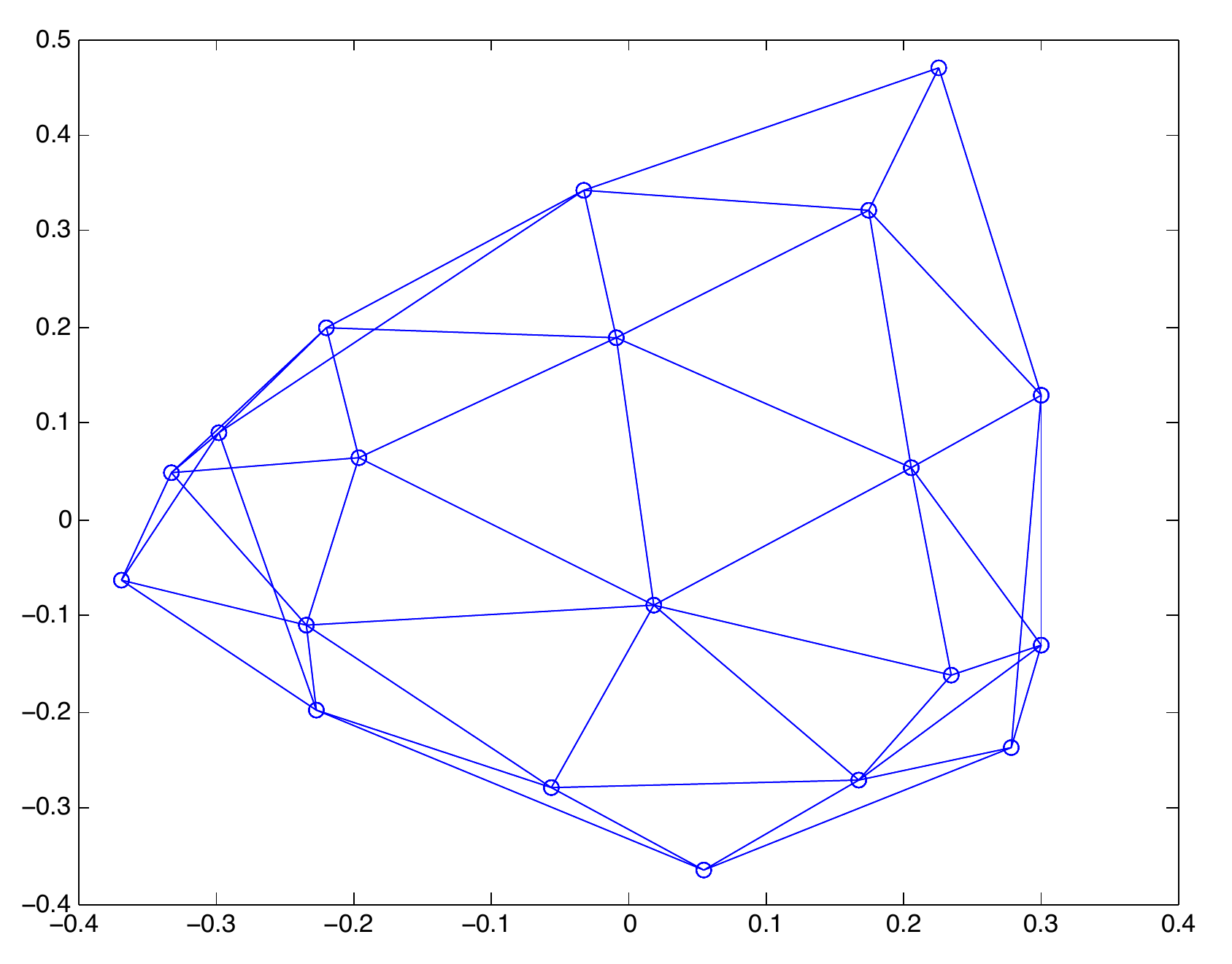}
  \end{center}
  \caption{Delaunay triangulation (left) and
drawing of the graph from Example 4 (right).}
\label{graph3}
\end{figure}

The Delaunay triangulation of the set of $20$ points and the drawing
of the corresponding graph are shown in Figure \ref{graph3}.
The graph drawing on the right looks nicer than the graph on the left
but is is no longer planar.

\medskip\noindent
{\it Example\/} 5.
Our last example, also borrowed from Spielman \cite{Spielman}, corresponds to  the 
skeleton of the ``Buckyball,'',  a geodesic dome 
invented by the architect Richard Buckminster Fuller (1895--1983).
The Montr\'eal Biosph\`ere is an example of a geodesic dome designed by
 Buckminster Fuller.

\begin{verbatim}
A = full(bucky);
D = diag(sum(A));
L = D - A;
[v, e] = eig(L);
gplot(A, v(:, [2 3]))
hold on;
gplot(A,v(:, [2 3]), 'o')
\end{verbatim}

\begin{figure}[http]
  \begin{center}
\hspace{1cm}
   \includegraphics[height=2truein,width=2.2truein]{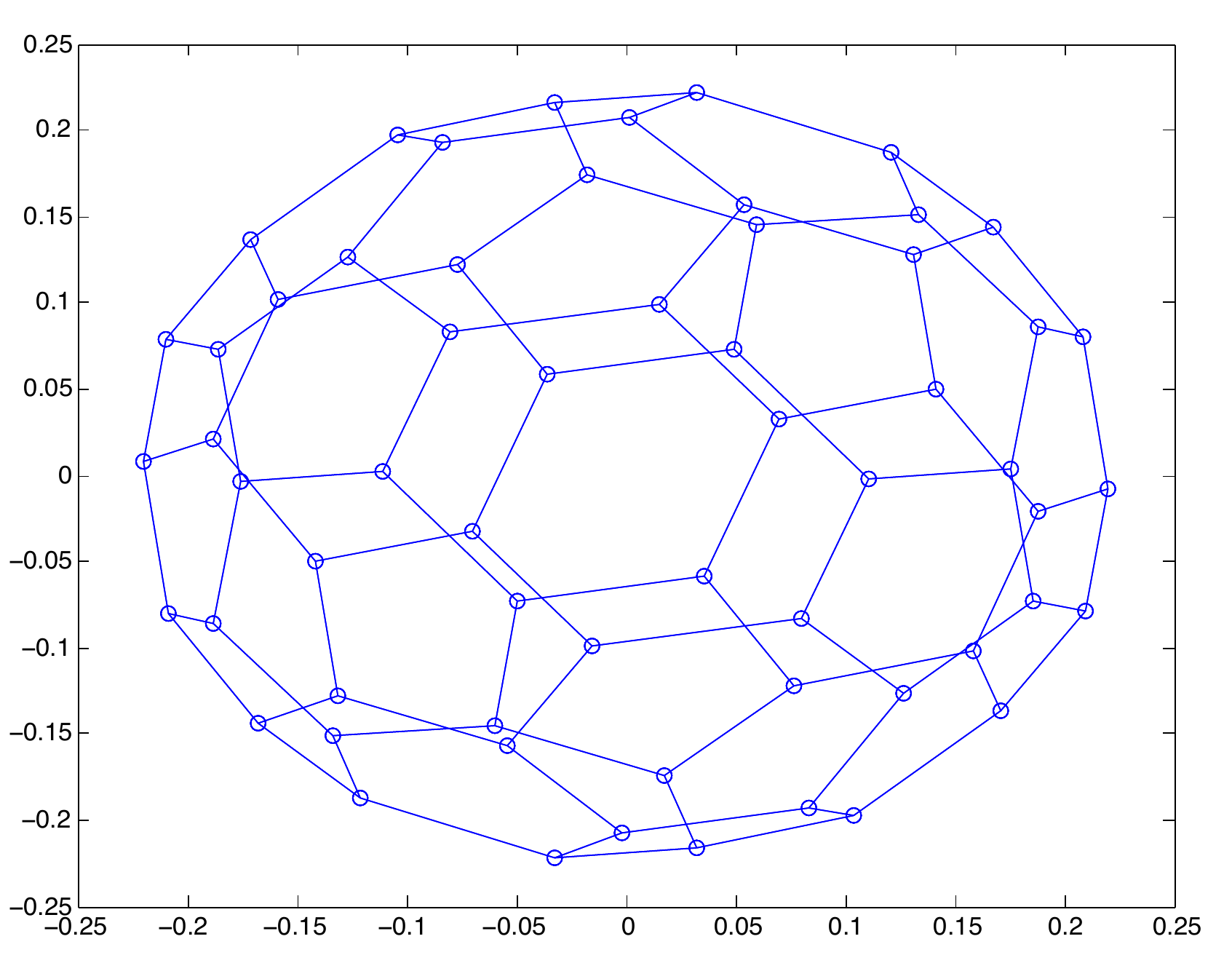}
  \end{center}
  \caption{Drawing of the graph of the Buckyball.}
\label{graph4}
\end{figure}

Figure \ref{graph4} shows a  graph drawing of the Buckyball. This
picture seems a bit squashed for two reasons. First, it is really a
$3$-dimensional graph; second, $\lambda_2 = 0.2434$ is a triple
eigenvalue.
(Actually, the Laplacian of $L$  has many multiple eigenvalues.)
What we should really do is to plot this graph in $\reals^3$ using
three orthonormal eigenvectors associated with $\lambda_2$.

\begin{figure}[http]
  \begin{center}
\hspace{1cm}
   \includegraphics[height=2.7truein,width=4truein]{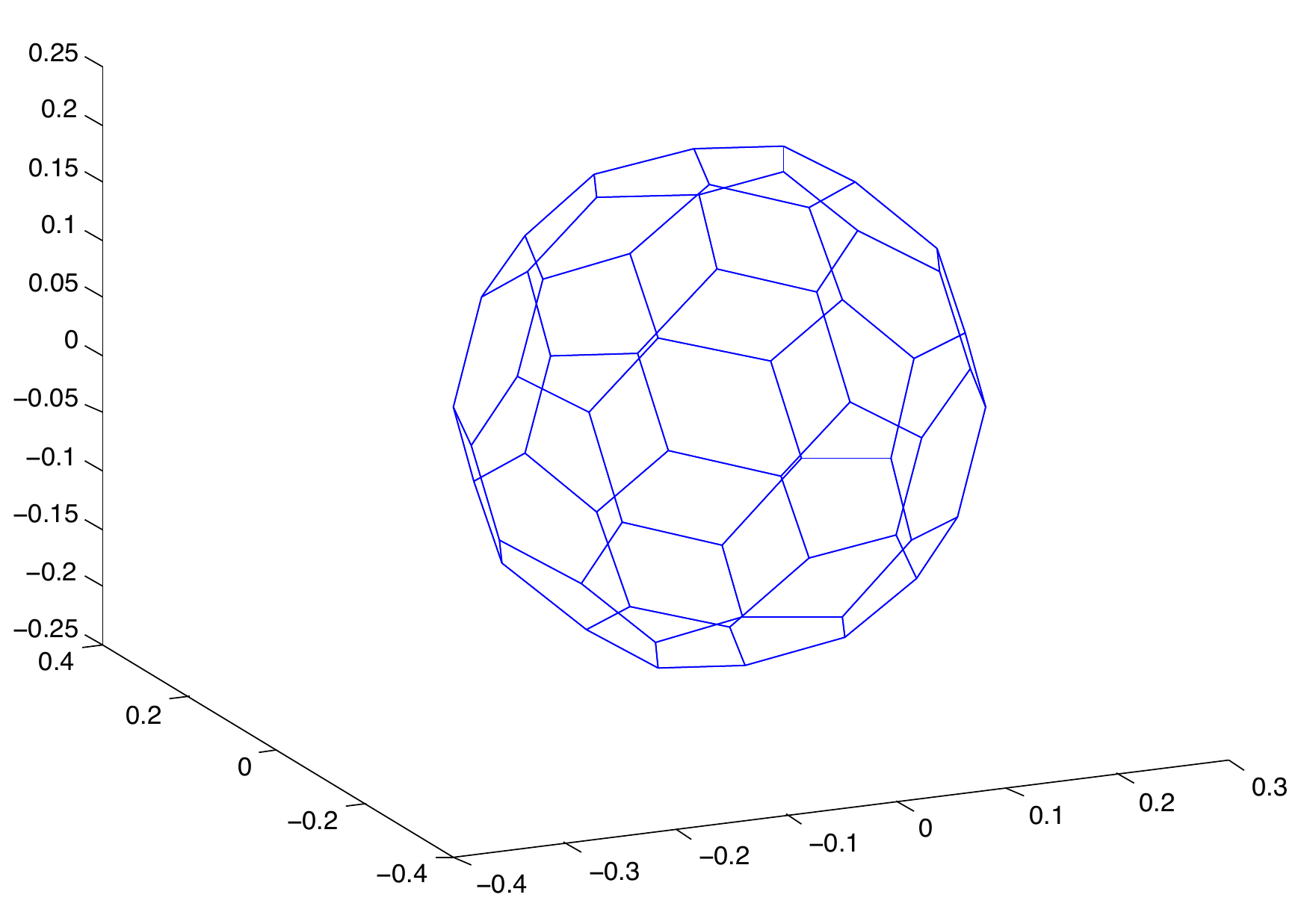}
  \end{center}
  \caption{Drawing of the graph of the Buckyball in $\reals^3$.}
\label{graph5}
\end{figure}

\medskip
A $3$D picture of the  graph of the Buckyball is produced by the
following {\tt Matlab} program, and its image is shown in Figure  \ref{graph5}.
It looks better!

\begin{verbatim}
[x, y] = gplot(A, v(:, [2 3]));
[x, z] = gplot(A, v(:, [2 4]));
plot3(x,y,z)
\end{verbatim}

\chapter{Graph Clustering}
\label{chap3}
\section{Graph Clustering Using Normalized  Cuts}
\label{ch3-sec1}
Given a set of data, the goal of clustering is to partition the data
into different groups according to their similarities. When the data
is given in terms of a similarity graph $G$, where the weight $w_{i\, j}$
between two nodes $v_i$ and $v_j$ is a measure of similarity of
$v_i$ and $v_j$, the problem can be stated as follows:
Find a partition $(A_1, \ldots, A_K)$ of the set of nodes $V$ into
different groups such   that the edges between different groups have
very low weight (which indicates that the points in different clusters
 are dissimilar),  and the edges within a group have high weight
 (which indicates that points within the same cluster are similar).

\medskip
The above graph clustering problem can be formalized as an
optimization problem, using the notion of cut mentioned at the 
end of Section \ref{ch1-sec1}.

\medskip
Given a subset $A$ of the set of vertices $V$, recall that we define $\mathrm{cut}(A)$ by
\[
\mathrm{cut}(A) = \mathrm{links}(A, \overline{A})= 
\sum_{v_i\in A, v_j\in \overline{A}} w_{i\, j},
\]
and that
\[
\mathrm{cut}(A)= \mathrm{links}(A, \overline{A}) = 
\mathrm{links}(\overline{A}, A) = \mathrm{cut}(\overline{A}).
\]
If we want to partition $V$ into $K$ clusters, we can do so by finding
a partition ($A_1, \ldots, A_K$) that  minimizes the quantity
\[
\mathrm{cut}(A_1, \ldots, A_K) = \frac{1}{2} \sum_{1 = 1}^K \mathrm{cut}(A_i). 
\]
The reason for introducing the factor $1/2$ is to avoiding counting
each edge twice.  In particular,
\[
\mathrm{cut}(A,\overline{A}) = \mathrm{links} (A, \overline{A}).
\]
For $K = 2$,  the mincut problem is a classical
problem that can be solved efficiently, but in practice, it does not
yield satisfactory partitions. Indeed, in many cases, the mincut
solution separates one vertex from the rest of the graph.  What we
need is to design our cost function in such a way that it keeps the
subsets $A_i$ ``reasonably large'' (reasonably balanced).

\begin{figure}[http]
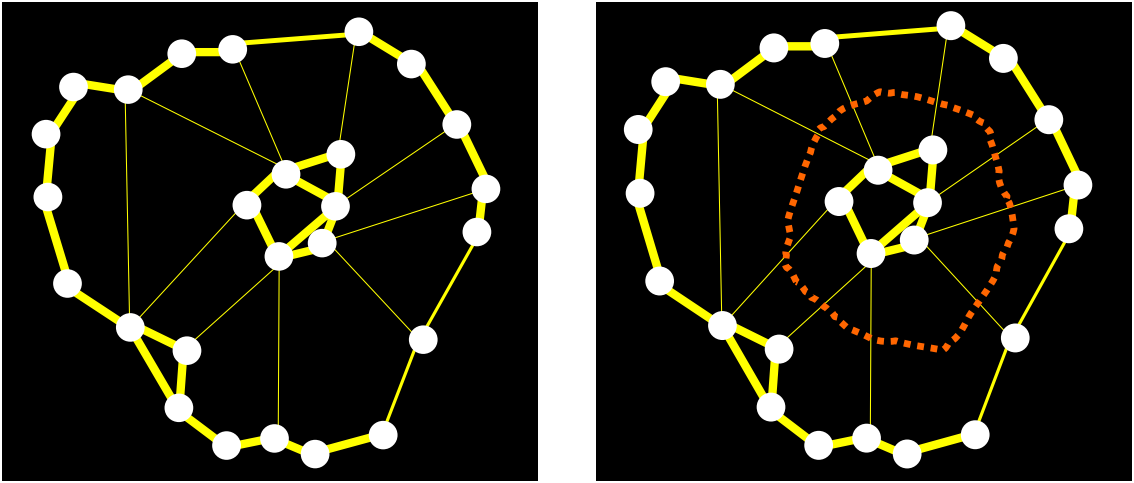

  \begin{center}
 \includegraphics[height=2.5truein,width=2.8truein]{ncuts-figs/ncuts-g-fig1.pdf}
\hspace{0.5cm}
 \includegraphics[height=2.5truein,width=2.8truein]{ncuts-figs/ncuts-g-fig2.pdf}
  \end{center}
  \caption{A weighted graph and its partition into two clusters.}
\label{ncg-fig4}
\end{figure}

\medskip
A example of a weighted graph and a partition of
its nodes into two clusters is shown in Figure \ref{ncg-fig4}.

\medskip
A way to get around this problem is to normalize the cuts by dividing
by some measure of each subset $A_i$. One possibility if to use the
size (the number of elements) of $A_i$. Another is to use the volume
$\mathrm{vol}(A_i)$ of $A_i$. A solution using the second measure (the volume)
(for $K = 2$) was proposed and
investigated in a seminal paper of  Shi and Malik \cite{ShiMalik}.
Subsequently, Yu (in her dissertation \cite{Yu}) and Yu and Shi
\cite{YuShi2003} extended the method to $K > 2$ clusters.
We will describe this  method later.
The idea is  to minimize the cost function
\[
\mathrm{Ncut}(A_1, \ldots, A_K) = 
 \sum_{i = 1}^K \frac{\mathrm{links}(A_i, \overline{A_i})}{\mathrm{vol}(A_i)}
= \sum_{i = 1}^K 
\frac{\mathrm{cut}(A_i, \overline{A_i})}{\mathrm{vol}(A_i)}.
\]

\medskip
We begin with the case $K = 2$, which is easier to handle.

\section{Special Case: $2$-Way Clustering Using Normalized  Cuts}
\label{ch3-sec2}
Our goal is to express our  optimization problem in matrix form.
In the case of two clusters, a single vector $X$ can be used to describe the
partition $(A_1, A_2)  = (A, \overline{A})$. We need to choose the structure of this vector
in such a way that $\mathrm{Ncut}(A, \overline{A})$ 
is equal to the Rayleigh ratio
\[
\frac{\transpos{X} L X}{\transpos{X} \Degsym X}.
\]
It is also important to pick a vector representation 
which is invariant under multiplication by a nonzero scalar,
because the Rayleigh ratio is scale-invariant, and it is crucial to take
advantage of this fact to make the denominator go away.

\medskip
Let $N = |V|$ be the number of nodes in the graph $G$.
In view of the desire for a scale-invariant representation,
it is natural
to assume that the  vector $X$ is of the form
\[
X = (x_1, \ldots, x_N),
\]
where $x_i \in \{a, b\}$ for $i = 1, \ldots, N$, 
for any two distinct  real numbers $a, b$. 
This is an indicator vector in
the sense that, for $i = 1, \ldots, N$,
\[
x_i =
\begin{cases}
a & \text{if $v_i \in A$} \\
b & \text{if $v_i \notin A$} .
\end{cases}
\]

The correct interpretation is really to 
view $X$ as a representative of a point 
in the real projective space $\mathbb{RP}^{N-1}$,  
namely the point $\mathbb{P}(X)$ of homogeneous
coordinates $(x_1\co \cdots \co x_N)$.
Therefore,  from now on, we  view $X$ as a vector of homogeneous
coordinates representing  the point  $\mathbb{P}(X)\in \mathbb{RP}^{N-1}$.

\medskip
Let  $d = \transpos{\mathbf{1}} \Degsym \mathbf{1}$ and $\alpha =
\mathrm{vol}(A)$.  
Then, $\mathrm{vol}(\overline{A}) = d - \alpha$.
By Proposition \ref{Laplace1}, we have
\[
\transpos{X} L X = (a - b)^2\, \mathrm{cut}(A, \overline{A}),
\]
and we easily check that
\[
\transpos{X} \Degsym  X = \alpha a^2 + (d - \alpha)b^2.
\]
Since $\mathrm{cut}(A, \overline{A}) = \mathrm{cut}(\overline{A}, A)$,
we have
\[
\mathrm{Ncut}(A, \overline{A}) = 
\frac{\mathrm{cut}(A,  \overline{A})}{\mathrm{vol}(A)} +
\frac{\mathrm{cut}(\overline{A}, A)}{\mathrm{vol}(\overline{A})} 
= 
\left(\frac{1}{\mathrm{vol}(A)}  +
  \frac{1}{\mathrm{vol}(\overline{A})} \right) \mathrm{cut}(A, \overline{A}),
\]
so we obtain
\[
\mathrm{Ncut}(A, \overline{A}) 
= \left(\frac{1}{\alpha}  +
  \frac{1}{d - \alpha} \right) \mathrm{cut}(A, \overline{A})
=
\frac{d}{\alpha(d - \alpha)}\,\mathrm{cut}(A, \overline{A}).
\]
Since
\[
\frac{\transpos{X} L X}{\transpos{X} \Degsym X} =
\frac{(a - b)^2}{\alpha a^2 + (d - \alpha)b^2}\, \mathrm{cut}(A, \overline{A}),
\]
in order to have
\[
\mathrm{Ncut}(A, \overline{A}) = \frac{\transpos{X} L X}{\transpos{X} \Degsym X},
\]
we need to find $a$ and $b$ so that
\[
\frac{(a - b)^2}{\alpha a^2 + (d - \alpha)b^2} = 
\frac{d}{\alpha(d - \alpha)}.
\]
The above is equivalent to
\[
(a - b)^2 \alpha(d - \alpha) =  \alpha d a^2 + (d - \alpha)d b^2,
\]
which can be rewritten as
\[
a^2(\alpha d - \alpha(d - \alpha)) + b^2(d^2 - \alpha d - \alpha(d -
\alpha)) + 2 \alpha(d - \alpha) a b = 0.
\]
The above yields
\[
a^2 \alpha^2 + b^2(d^2 - 2\alpha d + \alpha^2) + 2\alpha(d - \alpha) a
b = 0,
\]
that is,
\[
a^2\alpha^2 + b^2(d - \alpha)^2 + 2\alpha(d - \alpha) a b = 0,
\]
which reduces to
\[
(a \alpha + b(d - \alpha))^2 = 0.
\]
Therefore, we get the condition
\begin{equation}
a \alpha + b(d - \alpha) = 0.
\tag{$\dagger$}
\end{equation}
Note that condition $(\dagger)$ applied to a vector $X$ whose
components are $a$ or $b$ is equivalent to the fact that $X$
is orthogonal to $\Degsym \mathbf{1}$, since
\[
\transpos{X} \Degsym \mathbf{1} =  \alpha  a+ (d - \alpha) b,
\]
where $\alpha = \mathrm{vol}(\{v_i\in V \mid x_i = a\})$.

\medskip
We claim the following two facts.
For any nonzero vector $X$ whose components are $a$ or
$b$, if  $\transpos{X} \Degsym \mathbf{1} =  \alpha  a+ (d - \alpha) b
= 0$, then
\begin{enumerate}
\item[(1)]
$\alpha\not= 0$ and $\alpha \not= d$ iff 
$a\not= 0$ and  $b\not= 0$.
\item[(2)]
if $a, b\not= 0$, then $a\not = b$. 
\end{enumerate}

\medskip
(1)
First assume that $a \not= 0$ and $b \not= 0$. If $\alpha = 0$, then
$\alpha  a + (d - \alpha) b = 0$
yields $d  b = 0$ with $d\not= 0$, which implies $b = 0$, a contradiction.
If $d - \alpha = 0$, then we get $d a = 0$ with $d\not= 0$, which implies $a = 0$, a contradiction.

\medskip
Conversely, assume that $\alpha\not= 0$ and $\alpha \not= d$. If
$a = 0$, then from  $\alpha  a+ (d - \alpha) b = 0$ we get
$(d - \alpha) b = 0$, which implies $b = 0$, contradicting the fact
that $X\not = 0$. 
Similarly, if $b = 0$,  then we get $\alpha  a = 0$, 
which implies $a = 0$, contradicting the fact
that $X\not = 0$. 

\medskip
(2) 
If $a, b\not= 0$, $a = b$ and $\alpha  a+ (d - \alpha) b = 0$, then 
$\alpha  a+ (d - \alpha) a = 0$,  and since $a\not= 0$, we deduce that
$d = 0$, a contradiction.

\medskip
If  $\transpos{X} \Degsym \mathbf{1} =  \alpha  a+ (d - \alpha) b = 0$
and $a, b\not= 0$, then
\[
b = -\frac{\alpha}{(d - \alpha)}\, a, 
\]
so we get
\begin{align*}
\alpha a^2 + (d - \alpha)b^2& = \alpha \frac{(d -
  \alpha)^2}{\alpha^2} b^2+ (d - \alpha)b^2 \\
& = (d - \alpha)\left( \frac{d - \alpha}{\alpha} + 1 \right)b^2 
 = \frac{(d - \alpha) d b^2}{\alpha},
\end{align*}
and
\begin{align*}
(a - b)^2 & = \left( -\frac{(d - \alpha)}{\alpha}\, b - b \right)^2 \\
& = \left( \frac{d - \alpha}{\alpha} + 1 \right)^2b^2 
 = \frac{d^2 b^2}{\alpha^2}.
\end{align*}
Since 
\begin{align*}
\transpos{X} \Degsym X & = \alpha a^2 + (d - \alpha)b^2 \\
\transpos{X} L X & =  (a - b)^2\, \mathrm{cut}(A, \overline{A}),
\end{align*}
we obtain
\begin{align*}
\transpos{X} \Degsym X & =  \frac{(d - \alpha)d b^2}{\alpha}
=  \frac {\alpha d  a^2}{(d - \alpha)} \\
\transpos{X} L X & =  
\frac{d^2 b^2}{\alpha^2} \, \mathrm{cut}(A, \overline{A}) 
= \frac{d^2 a^2}{(d - \alpha)^2} \, \mathrm{cut}(A, \overline{A}) .
\end{align*} 
If we wish to make $\alpha$ disappear, we   pick
\[
a = \sqrt{\frac{d - \alpha}{\alpha}}, \quad
b = -\sqrt{\frac{\alpha}{d - \alpha}},
\]
and then
\begin{align*}
\transpos{X} \Degsym X & =   d  \\
\transpos{X} L X & =  
\frac{d^2}{\alpha(d - \alpha)} \,
\mathrm{cut}(A, \overline{A})
= d\, \mathrm{Ncut}(A, \overline{A}).
\end{align*} 
In this case, we are  considering indicator vectors of the form
\[
\left\{
(x_1, \ldots, x_N) \mid
x_i \in \left\{
\sqrt{\frac{d - \alpha}{\alpha}},
-\sqrt{\frac{\alpha}{d - \alpha}} \right\},
\alpha = \mathrm{vol}(A)  \right\},
\]
for any nonempty proper subset $A$
of $V$. This is the choice adopted in 
von Luxburg \cite{Luxburg}.
Shi and Malik \cite{ShiMalik} use
\[
a = 1, \quad b = -\frac{\alpha}{d - \alpha} = - \frac{k}{1 - k},
\]
with
\[
k = \frac{\alpha}{d}.
\]
Another choice found in the literature (for example, in Belkin and Niyogi
\cite{Belkin-Niyogi}) is
\[
a  = \frac{1}{\alpha}, \quad
b  = -\frac{1}{d - \alpha} .
\]
However, there is no need to restrict solutions to be of either of these forms. 
So, let
\[
\s{X} = \big\{
(x_1, \ldots, x_N) \mid x_i \in \{a, b\}, \> a, b\in \reals,\> a,
b\not = 0
\big\},
\]
so that our solution set is
\[
\s{K}  = \big\{
X  \in\s{X}  \mid \transpos{X}  \Degsym\mathbf{1} = 0
\big\},
\]
because by previous observations, since  vectors
$X\in \s{X}$ have nonzero components,  $\transpos{X}\Degsym\mathbf{1} = 0$ 
implies that $\alpha \not= 0$,  $\alpha \not= d$, and $a\not= b$, where
$\alpha = \mathrm{vol}(\{v_i\in V \mid x_i = a\})$.
Actually, to be perfectly rigorous, we are looking for solutions in
$\mathbb{RP}^{N-1}$, so our solution set is really
\[
\mathbb{P}(\s{K})  = \big\{
(x_1\co \cdots\co x_N) \in \mathbb{RP}^{N-1}\mid
(x_1, \ldots, x_N) \in \s{K}
\big\}.
\]
Consequently, our minimization problem can be stated as follows:

\medskip\noindent
{\bf Problem PNC1}
\begin{align*}
& \mathrm{minimize}     &  & 
\frac{\transpos{X} L X}{\transpos{X} \Degsym X} & &  &  &\\
& \mathrm{subject\ to} &  &
\transpos{X} \Degsym\mathbf{1} = 0,  & &  X\in \s{X}.      
\end{align*}

It is understood that the solutions are  points $\mathbb{P}(X)$
in $\mathbb{RP}^{N-1}$.

\medskip 
Since the Rayleigh ratio and the constraints 
$\transpos{X}\Degsym\mathbf{1} = 0$ and $X\in \s{X}$ 
are scale-invariant 
(for any $\lambda \not= 0$, 
the Rayleigh ratio does not change if $X$ is
replaced by $\lambda X$, $X\in \s{X}$ iff $\lambda X \in \s{X}$,
and $\transpos{ (\lambda X)}\Degsym\mathbf{1} = \lambda \transpos{
  X}\Degsym\mathbf{1} = 0$),
we are led to the following formulation of our problem:

\medskip\noindent
{\bf Problem PNC2}
\begin{align*}
& \mathrm{minimize}     &  & 
\transpos{X} L X & &  &  &\\
& \mathrm{subject\ to} &  & \transpos{X} \Degsym X = 1, &&
 \transpos{X} \Degsym\mathbf{1} = 0, && X\in \s{X}.   
\end{align*}

\medskip
Problem PNC2 is equivalent to problem PNC1 in the sense that 
if $X$ is any minimal solution of PNC1, then $X/(\transpos{X} \Degsym
X)^{1/2}$ is a minimal solution of PNC2 (with the same minimal value
for the objective functions), and if $X$ is a minimal solution of
PNC2, then $\lambda X$ is a minimal solution for PNC1 for all
$\lambda\not= 0$ (with the same minimal value
for the objective functions). Equivalently, problems PNC1 and PNC2
have the same set of minimal solutions as points $\mathbb{P}(X)
\in\mathbb{RP}^{N-1}$ given by their homogenous coordinates $X$.

\medskip
Unfortunately, this is an NP-complete problem, as
shown by Shi and Malik \cite{ShiMalik}.
As often with hard combinatorial problems, we can look for a {\it
  relaxation\/} of our problem, which means looking for an optimum in
a larger continuous domain. After doing this, the problem is to find
a discrete solution which is close to a continuous optimum of the
relaxed problem.

\medskip
The natural relaxation of this problem is to allow $X$ to be any nonzero
vector in $\reals^N$, and we get the problem:

\[
\mathrm{minimize} \quad \transpos{X} L X
\quad\mathrm{subject\ to} \quad \transpos{X} \Degsym X = 1, \quad
\transpos{X} \Degsym\mathbf{1} = 0.
\]

\medskip
As usual, let
$Y = \Degsym^{1/2} X$, so that $X = \Degsym^{-1/2} Y$. Then, 
the condition   $\transpos{X}\Degsym X = 1$ becomes
\[
\transpos{Y} Y = 1,
\]
the condition 
\[
\transpos{X} \Degsym \mathbf{1} = 0
\]
becomes
\[
\transpos{Y} \Degsym^{1/2} \mathbf{1} = 0,
\]
and 
\[
\transpos{X} L X = \transpos{Y} \Degsym^{-1/2} L \Degsym^{-1/2} Y. 
\]

We obtain the problem:

\[
\mathrm{minimize} \quad \transpos{Y} \Degsym^{-1/2} L \Degsym^{-1/2}  Y
\quad\mathrm{subject\ to} \quad \transpos{Y}  Y = 1, \quad
\transpos{Y} \Degsym^{1/2}\mathbf{1} = 0.
\]

\medskip
Because $L\mathbf{1} = 0$, the vector $ \Degsym^{1/2} \mathbf{1}$
belongs to
the nullspace of the symmetric Laplacian $L_{\mathrm{sym}} =
\Degsym^{-1/2} L \Degsym^{-1/2}$.
By Proposition   \ref{PCAlem1}, 
minima are achieved by any unit eigenvector $Y$ of the
second eigenvalue $\nu_2$ of $L_{\mathrm{sym}}$.
Then, $Z = \Degsym^{-1/2} Y$ is a solution of our original relaxed
problem.
Note that because $Z$ is nonzero and  orthogonal to $\Degsym \mathbf{1}$, a vector
with positive entries, it must have negative and positive entries.

\medskip
The next question is to figure how close is $Z$  to an exact  solution in
$\s{X}$.  
Actually, because solutions are points in
$\mathbb{RP}^{N-1}$, the correct statement of the question is:
Find an exact solution $\mathbb{P}(X) \in\mathbb{P}(\s{X})$
which is the closest (in a suitable sense) to the approximate
solution $\mathbb{P}(Z)\in \mathbb{RP}^{N-1}$. 
However, because $\s{X}$ is closed under the antipodal map, as
explained in Appendix \ref{ch3-sec6}, minimizing the distance
$d(\mathbb{P}(X), \mathbb{P}(Z))$ on $\mathbb{RP}^{N-1}$ is equivalent
to minimizing the Euclidean distance $\norme{X - Z}_2$
(if we use the Riemannian metric on $\mathbb{RP}^{N-1}$ induced by the Euclidean
metric on $\reals^N$).

\medskip
We may assume $b < 0$, in which case $a > 0$.
If all entries in $Z$ are nonzero, due to the projective nature of the
solution set, it seems reasonable to
say that
the partition of $V$ is defined by the signs of the entries in $Z$. 
Thus, $A$ will consist of nodes those $v_i$ for which $x_i > 0$.
Elements corresponding to zero entries can be assigned to either $A$
or $\overline{A}$, unless additional information is available.

\medskip
Now, using the fact that
\[
b = - \frac{\alpha a}{d - \alpha},
\]
a better solution is to look for a vector  $X\in \reals^N$ with $X_i
\in \{a, b\}$ which is closest to a  minimum $Z$ of the relaxed problem.
Here is a proposal for an algorithm. 

\medskip
For any  solution $Z$ of the relaxed problem,
let $I_{Z}^{+} = \{i \mid Z_i > 0\}$ be the set of indices of positive
entries in $Z$,  $I_{Z}^{-} = \{i \mid Z_i < 0\}$ the set of
indices of negative
entries in $Z$, $I_{Z}^{0} = \{i \mid Z_i = 0\}$  the set of indices
of zero entries in $Z$, and let
$Z^+$ and $Z^-$ be the vectors given
by
\begin{align*}
Z^+_i & = 
\begin{cases}
Z_i & \text{if $i \in I_Z^+$} \\
0 & \text{if $i \notin I_Z^+$} 
\end{cases}
&
Z^-_i & = 
\begin{cases}
Z_i & \text{if $i \in I_Z^-$} \\
0 & \text{if $i \notin I_Z^-$} 
\end{cases}.
\end{align*}
Also let 
$n_a  = |I_Z^{+} |$, $n_b  = |I_Z^{-} |$, let $\overline{a}$ and $\overline{b}$ be the
average of the positive and negative entries in $Z$ respectively, that is,
\begin{align*}
\overline{a} &  = \frac{\sum_{i\in I_Z^+} Z_i}{n_a}  & 
\overline{b} & = \frac{\sum_{i\in I_Z^-} Z_i}{n_b}, 
\end{align*}
and let $\overline{Z^+}$ and $\overline{Z^-}$ be the vectors given by
\begin{align*}
(\overline{Z^+})_i & = 
\begin{cases}
\overline{a} & \text{if $i \in I_Z^+$} \\
0 & \text{if $i \notin I_Z^+$} 
\end{cases}
&
(\overline{Z^-})_i & = 
\begin{cases}
\overline{b} & \text{if $i \in I_Z^-$} \\
0 & \text{if $i \notin I_Z^-$} 
\end{cases}.
\end{align*}
If $\norme{\overline{Z^+} - Z^+} > \norme{\overline{Z^-} - Z^-}$, then 
replace $Z$ by $-Z$.
Then, perform the following steps:
\begin{enumerate}
\item[(1)]
Let 
 \begin{align*}
n_a & = |I_Z^{+} |, &
\alpha & = \mathrm{vol}(\{v_i \mid i \in I_Z^{+}\}), &
\beta & = \frac{\alpha}{d - \alpha} ,
\end{align*}
and form the vector $\overline{X}$ with
\[
\overline{X}_i = 
\begin{cases}
a & \text{if $i \in I_Z^{+}$} \\
-\beta a & \text{otherwise}, 
\end{cases}
\]
such that $\norme{\overline{X} - Z}$ is minimal; the scalar $a$ is
determined by finding the  solution of the equation 
\[
\s{Z} a  =Z,
\]
in the least squares sense, 
where 
\[
\s{Z} = 
\begin{cases}
1 & \text{if $i \in I_Z^{+}$} \\
-\beta  & \text{otherwise}, 
\end{cases}
\]
and is given by
\[
a = (\transpos{\s{Z}} \s{Z})^{-1}\transpos{\s{Z}}Z;
\]
that is,
\[
a = \frac{\sum_{i \in I_Z^{+}} Z_i}{n_a + \beta^2(N - n_a)} - 
\frac{\sum_{i \in I_{Z}^{-}} \beta Z_i}{n_a + \beta^2(N - n_a)} .
\]

\item[(2)]
While  $I_Z^0 \not= \emptyset$,  
pick the smallest index $i\in I_Z^0$, 
 compute
\begin{align*}
\widetilde{I}_{Z}^{+} & =  I_{Z}^{+} \cup \{i\} \\  
\widetilde{n}_a  & = n_a + 1 \\
\widetilde{\alpha} & =  \alpha + d(v_i) \\
\widetilde{\beta} & = \frac{\widetilde{\alpha}}{d - \widetilde{\alpha}},
\end{align*}
and then $\widetilde{X}$ with
\[
\widetilde{X}_j = 
\begin{cases}
\widetilde{a} & \text{if $j \in \widetilde{I}_Z^{+}$} \\
-\widetilde{\beta} \widetilde{a}  & \text{otherwise} ,
\end{cases}
\]
and
\[
\widetilde{a} = \frac{\sum_{j \in \widetilde{I}_Z^{+}} Z_j}{\widetilde{n}_a + \widetilde{\beta}^2(N - \widetilde{n}_a)} - 
\frac{\sum_{j \in I_{Z}^{-}} \widetilde{\beta}
  Z_j}{\widetilde{n}_a + 
\widetilde{\beta}^2(N - \widetilde{n}_a)} .
\]
If $\smnorme{\widetilde{X} - Z} < \smnorme{\overline{X} - Z}$, 
then
let $\overline{X} = \widetilde{X}$, $I_Z^+ = \widetilde{I}_Z^+$,
$n_a = \widetilde{n}_a$,  $\alpha = \widetilde{\alpha}$, and $I_Z^0 =
I_Z^0 - \{i\}$; go back to (2).
\item[(3)]
The final answer if $\overline{X}$.
\end{enumerate}

\section{$K$-Way Clustering Using Normalized  Cuts}
\label{ch3-sec3}
We now consider the general case in which $K \geq 3$.
Two crucial  issues need  to be addressed
(to the best of our knowledge, these points are not clearly articulated 
in the literature).
\begin{enumerate}
\item
The choice of a matrix representation for partitions on the set of vertices.
It is important that such a representation be scale-invariant.
It is also necessary to state necessary and sufficient conditions
for such matrices to represent a partition.
\item
The choice of a metric to compare solutions.
It turns out that the space of discrete solutions can be
viewed as a subset of the $K$-fold product 
$(\mathbb{RP}^{N - 1})^K$ of the projective space $\mathbb{RP}^{N - 1}$.
Version 1 of the formulation of our minimization problem (PNC1)
makes this point clear. However, the relaxation $(*_1)$ of 
version 2 of  our minimization problem (PNC2), which is equivalent to
version 1, reveals that that the solutions of the relaxed problem
$(*_1)$ are members of the {\it Grassmannian\/} $G(K, N)$.
Thus, we have two choices of metrics: (1) a metric on
$(\mathbb{RP}^{N - 1})^K$; (2) a metric on $G(K, N)$.
We  discuss the first choice, which is the choice implicitly
adopted by Shi and Yu.
\end{enumerate}

\medskip
We describe a  partition $(A_1, \ldots, A_K)$ of the set of nodes $V$ by
an $N\times K$ matrix  $X = [X^1 \cdots X^K]$
whose columns $X^1, \ldots,
X^K$ are indicator vectors of the partition $(A_1, \ldots, A_K)$.
Inspired by what we did in Section \ref{ch3-sec2}, 
we assume that the  vector $X^j$ is of the form
\[
X^j = (x_1^j, \ldots, x_N^j),
\]
where $x_i^j \in \{a_j, b_j\}$ for $j = 1, \ldots, K$ and
$i = 1, \ldots, N$, and
where $a_j, b_j$ are 
any  two distinct  real numbers.
The vector $X^j$  is an indicator vector for $A_j$ in
the sense that, for $i = 1, \ldots, N$,
\[
x_i^j =
\begin{cases}
a_j & \text{if $v_i \in A_j$} \\
b_j & \text{if $v_i \notin A_j$} .
\end{cases}
\]

When $\{a_j, b_j\} = \{0, 1\}$ for $j = 1, \ldots, K$,  such a matrix is called
a {\it partition matrix\/} by Yu and Shi. However, such a choice is
premature, since it is better to have a scale-invariant representation
to make the denominators of the Rayleigh ratios go away.

\medskip
Since the partition $(A_1, \ldots, A_K)$ consists of nonempty pairwise
disjoint blocks whose union is $V$,
some  conditions on $X$ are required to  reflect
these properties, but  we will worry about this later. 

\medskip
Let  $d = \transpos{\mathbf{1}} \Degsym \mathbf{1}$ and $\alpha_j =
\mathrm{vol}(A_j)$, so that $\alpha_1 + \cdots + \alpha_K = d$.
Then, $\mathrm{vol}(\overline{A_j}) = d - \alpha_j$,
and as in Section \ref{ch3-sec2},  we have
\begin{align*}
\transpos{(X^j)} L X^j & = 
(a_j - b_j)^2\, \mathrm{cut}(A_j, \overline{A_j}), \\
\transpos{(X^j)} \Degsym  X^j & = \alpha_j a_j^2 + (d - \alpha_j)b_j^2.
\end{align*}
When $K \geq 3$,  unlike the case $K = 2$,   in general 
we have $\mathrm{cut}(A_j, \overline{A_j})  \not=  \mathrm{cut}(A_k,
\overline{A_k}) $
if $j \not= k$, and  since 
\[
 \mathrm{Ncut}(A_1, \ldots, A_K) 
= \sum_{j = 1}^K 
\frac{\mathrm{cut}(A_j, \overline{A_j})}{\mathrm{vol}(A_j)},
\]
we would like to choose  $a_j , b_j$ so that 
\[
\frac{\mathrm{cut}(A_j, \overline{A_j})}{\mathrm{vol}(A_j)} 
= \frac{\transpos{(X^j)} L X^j}{\transpos{(X^j)}\Degsym X^j} \quad j =
1, \ldots, K, 
\]
because this implies that
\[
\mu(X) = \mathrm{Ncut}(A_1, \ldots, A_K) 
= \sum_{j = 1}^K 
\frac{\mathrm{cut}(A_j, \overline{A_j})}{\mathrm{vol}(A_j)}
= \sum_{j = 1}^K 
\frac{\transpos{(X^j)} L X^j}{\transpos{(X^j)}\Degsym X^j}.
\]

Since
\[
\frac{\transpos{(X^j)} L X^j}{\transpos{(X^j)}\Degsym X^j} = 
\frac{(a_j - b_j)^2\, \mathrm{cut}(A_j, \overline{A_j})}
{\alpha_j a_j^2 + (d - \alpha_j)b_j^2}
\]
and $\mathrm{vol}(A_j) = \alpha_j$, in order to have
\[
\frac{\mathrm{cut}(A_j, \overline{A_j})}{\mathrm{vol}(A_j)} 
= \frac{\transpos{(X^j)} L X^j}{\transpos{(X^j)}\Degsym X^j} \quad j =
1, \ldots, K, 
\]
we need to have
\[
\frac{(a_j - b_j)^2} 
{\alpha_j a_j^2 + (d - \alpha_j)b_j^2} = \frac{1}{\alpha_j}
\quad j = 1, \ldots, K.
\]
Thus, we must have
\[
(a_j^2 - 2a_jb_j + b_j^2)\alpha_j = \alpha_j a_j^2 + (d - \alpha_j)b_j^2,
\]
which yields
\[
2 \alpha_jb_j(b_j - a_j) = db_j^2.
\]
The above equation is trivially satisfied if $b_j = 0$. If
$b_j\not= 0$, then
\[
2 \alpha_j(b_j - a_j) = db_j,
\]
which yields
\[
a_j = \frac{2\alpha_j - d}{2\alpha_j} b_j.
\]
This choice seems more complicated that the choice $b_j = 0$, so
we will opt for the choice $b_j = 0$, $j = 1, \ldots, K$.
With this choice, we get
\[
\transpos{(X^j)} \Degsym  X^j  = \alpha_j a_j^2.
\]
Thus, it makes sense to pick
\[
a_j = \frac{1}{\sqrt{\alpha_j}} =
\frac{1}{\sqrt{\mathrm{vol}(A_j)}}, \quad j = 1, \ldots, K,
\]
which is the solution presented in von Luxburg \cite{Luxburg}.
This choice also corresponds to the scaled partition matrix used in
Yu \cite{Yu} and Yu and Shi \cite{YuShi2003}.

\medskip
When $N = 10$ and $K = 4$, 
an example  of a matrix $X$ 
representing the partition
of $V = \{v_1, v_2, \ldots, v_{10}\}$
into the four   blocks 
\[
\{A_1, A_2, A_3, A_4\} = 
\{\{v_2, v_4, v_6\}, \{v_1, v_5\}, \{v_3, v_8, v_{10}\}, \{v_7, v_9\}\} , 
\]
is shown below:
\[
X = 
\begin{pmatrix}
0  & a_2  & 0  & 0 \\
a_1 & 0  & 0  & 0  \\
0  & 0  & a_3  & 0  \\
a_1 & 0  & 0  & 0  \\
0   & a_2  & 0  & 0  \\
a_1 & 0  & 0  & 0  \\
0  & 0  & 0  & a_4  \\
0  & 0  & a_3  & 0  \\
0  & 0  & 0  & a_4  \\
0   & 0  & a_3   & 0  
\end{pmatrix}.
\]

\medskip
Let us now consider the problem of finding necessary and sufficient
conditions for a matrix $X$ to represent a partition of $V$.

\medskip
When $b_j = 0$, the pairwise disjointness of the $A_i$ is captured
by the orthogonality of the $X^i$:
\begin{equation}
\transpos{(X^i)}X^j = 0, \quad 1 \leq i, j \leq  K, \> i \not= j.
\tag{$*$}
\end{equation}
This is because,  for any matrix $X$ where the nonzero entries in each
column have the same sign,
for any $i\not= j$, the condition
\[
\transpos{(X^i)} X^j = 0
\]
says that for every $k = 1,\ldots, N$,
if $x^i_k \not= 0$ then  $x^j_k = 0$.

\medskip
When we formulate our minimization problem in terms of Rayleigh
ratios, conditions on the quantities $\transpos{(X^i)} \Degsym X^i$ show up, and
it is more convenient to express the orthogonality conditions
using the quantities $\transpos{(X^i)} \Degsym X^j$ instead of  
the $\transpos{(X^i)}  X^j$, because
these various conditions can  be combined into a single condition 
involving the matrix $\transpos{X} \Degsym X$.
Now, because $\Degsym$ is a diagonal
matrix with positive entries and because  the nonzero entries in each
column of $X$ have the same sign,
for any $i\not= j$, the condition
\[
\transpos{(X^i)} X^j = 0
\]
is equivalent to
\begin{equation}
\transpos{(X^i)} \Degsym X^j = 0,
\tag{$**$}
\end{equation}
since, as above, it means that for    $k = 1, \ldots, N$,
if $x^i_k \not= 0$ then  $x^j_k = 0$.
Observe that the orthogonality conditions $(*)$ (and $(**)$)
are equivalent to the fact
that every row of $X$ has at most one nonzero entry.

\medskip
\remark
The disjointness condition
\[
X \mathbf{1}_K= \mathbf{1}_N
\]
is used in Yu \cite{Yu}.
However, this condition does guarantee the disjointness of
the blocks. For example, it is satisfied by the matrix $X$ whose
first column is $\mathbf{1}_N$,  with $0$ everywhere else.

\medskip
Each $A_j$ is nonempty iff $X^j \not= 0$, and the fact that the union of
the $A_j$ is $V$
is captured by the fact that each row of $X$ must have some
nonzero entry
(every vertex appears in some block).
It is not obvious how to state conveniently this condition in  matrix form.

\medskip
Observe that the diagonal entries of the matrix $X\transpos{X}$ are the
square Euclidean norms of the rows of $X$. Therefore, we can assert
that  these entries are all nonzero.   Let $\mathrm{DIAG}$ be the function which returns the
diagonal matrix (containing the diagonal of $A$),
\[
\mathrm{DIAG}(A) = \mathrm{diag}(a_{1\, 1}, \ldots, a_{n\, n}),
\]
for any square matrix $A = (a_{i \, j})$.
Then, the condition for the rows of $X$ to be nonzero can be stated as
\begin{equation*}
\det(\mathrm{DIAG}(X\transpos{X})) \not= 0. 
\end{equation*}

Observe that the matrix
\[
\mathrm{DIAG}(X\transpos{X})^{-1/2} X 
\]
is the result of normalizing the rows of $X$ so that they have
Euclidean norm $1$. This normalization step  is used by   Yu \cite{Yu}
in the search for  a discrete solution closest to a solution of 
a relaxation of our original problem.
For our special matrices representing  partitions, 
normalizing the rows will have the effect of rescaling the columns
(if row $i$ has $a_j$ in column $j$, then all nonzero entries
in column $j$ are equal to $a_j$), but for a more general matrix,
this is false.  Since  our solution matrices  
are invariant under rescaling the columns, but not the rows,
rescaling the rows does not appear to be a good idea.

\medskip
A better idea which leads to a scale-invariant condition stems from the observation 
that since every row of any matrix $X$ representing a partition
has a single nonzero entry $a_j$, we have
\[
\transpos{X}\mathbf{1} =
\begin{pmatrix}
n_1a_1 \\
\vdots \\
n_K a_K
\end{pmatrix},
\quad
\transpos{X}X =
\mathrm{diag}\left(n_1a_1^2,  \ldots, 
n_K a_K^2\right),
\]
where $n_j$ is the number of elements in $A_j$, the $j$th block of the
partition, which implies that
\[
(\transpos{X} X)^{-1} \transpos{X} \mathbf{1} = 
\begin{pmatrix}
\frac{1}{a_1} \\
\vdots \\
\frac{1}{a_K}
\end{pmatrix},
\]  
and thus,
\begin{equation}
X (\transpos{X} X)^{-1} \transpos{X} \mathbf{1} = \mathbf{1}.
\tag{$\dagger$}
\end{equation}
When $a_j = 1$ for $j = 1, \ldots, K$, we have
$(\transpos{X} X)^{-1} \transpos{X} \mathbf{1} = \mathbf{1}$, and 
condition $(\dagger)$ reduces to
\[
X\mathbf{1}_K = \mathbf{1}_N.
\]
Note that because the columns of $X$ are linearly independent,
$(\transpos{X} X)^{-1} \transpos{X}$ is the pseudo-inverse of
$X$. Consequently, condition $(\dagger)$, can also be written as
\[
XX^+ \mathbf{1} = \mathbf{1},
\]
where $X^+ = (\transpos{X} X)^{-1} \transpos{X}$ is the pseudo-inverse
of $X$. However, it is well known that $XX^+$ is the orthogonal
projection of $\reals^K$ onto the range of $X$
(see Gallier \cite{Gallbook2}, Section 14.1), 
so the condition
$XX^+ \mathbf{1} = \mathbf{1}$ is equivalent to the fact
that $\mathbf{1}$ belongs to the
range of $X$. In retrospect, this should have been obvious since
the columns of a solution $X$ satisfy the equation
\[
a_1^{-1} X^1 + \cdots + a_K^{-1} X^K  = \mathbf{1}.
\]

\medskip
We emphasize that it is important to use conditions that are invariant
under multiplication by a nonzero scalar, 
because the Rayleigh ratio is scale-invariant, and it is crucial to take
advantage of this fact to make the denominators go away.

\medskip
If we let
\[
\s{X}  = \Big\{[X^1\> \ldots \> X^K] \mid
X^j = a_j(x_1^j, \ldots, x_N^j) , \>
x_i^j \in \{1, 0\},
 a_j\in \reals, \> X^j \not= 0
\Big\}
\]
(note that the condition $X^j \not= 0$ implies that $a_j \not= 0$),
then the set of matrices representing partitions of $V$ into $K$
blocks is
\begin{align*}
& & &\s{K}  = \Big\{ X = [X^1 \> \cdots \> X^K] \quad \mid & &  X\in\s{X},  &&\\
         & & &  & &  \transpos{(X^i)} \Degsym X^j = 0, \quad 1\leq i, j \leq K,\> 
i\not= j, && \quad\quad\quad\quad\quad\\
&  & & & & 
 X (\transpos{X} X)^{-1} \transpos{X} \mathbf{1} = \mathbf{1}\Big\}. && 
\end{align*}

As in the case $K = 2$, to be rigorous, the solution are really
$K$-tuples of points in $\mathbb{RP}^{N-1}$, so our solution set is
really
\[
\mathbb{P}(\s{K})  = \Big\{(\mathbb{P}(X^1), \ldots, \mathbb{P}(X^K)) \mid
[X^1 \> \cdots \> X^K] \in \s{K} 
\Big\}.
\]

\medskip
In view of the above, we have our first formulation of $K$-way clustering
of a graph using normalized cuts, called problem PNC1 
(the notation PNCX  is used in  Yu \cite{Yu}, Section 2.1):

\medskip\noindent
{\bf $K$-way Clustering of a graph using Normalized Cut, Version 1: \\
Problem PNC1}

\begin{align*}
& \mathrm{minimize}     &  &  \sum_{j = 1}^K 
\frac{\transpos{(X^j)} L X^j}{\transpos{(X^j)}\Degsym X^j}& &  &  &\\
& \mathrm{subject\ to} &  & 
 \transpos{(X^i)} \Degsym X^j = 0, \quad 1\leq i, j \leq K,\> 
i\not= j,  & &  & & \\
& & & 
 X (\transpos{X} X)^{-1} \transpos{X} \mathbf{1} = \mathbf{1},  & & X\in \s{X}. & & 
\end{align*}

As in the case $K = 2$, the solutions that we are seeking are $K$-tuples 
$(\mathbb{P}(X^1), \ldots, \mathbb{P}(X^K))$ of points  in
$\mathbb{RP}^{N-1}$ determined by
their  homogeneous coordinates $X^1, \ldots, X^K$.

\medskip
\remark
Because
\[
\transpos{(X^j)} L X^j = 
\transpos{(X^j)} \Degsym X^j - \transpos{(X^j)} W X^j = 
\mathrm{vol}(A_j)   - \transpos{(X^j)} W X^j,
\]
Instead of minimizing
\[
\mu(X^1, \ldots, X^K) 
= \sum_{j = 1}^K 
\frac{\transpos{(X^j)} L X^j}{\transpos{(X^j)}\Degsym X^j},
\]
we can maximize
\[
\epsilon(X^1, \ldots, X^K) =  \sum_{j = 1}^K 
\frac{\transpos{(X^j)} W X^j}{\transpos{(X^j)}\Degsym X^j},
\]
since
\[
\epsilon(X^1, \ldots, X^K) = K - \mu(X^1, \ldots, X^K).
\]
This second option is the one chosen by
Yu \cite{Yu} and Yu and Shi \cite{YuShi2003} (actually, they work
with $\frac{1}{K}(K - \mu(X^1, \ldots, X^K))$, but this doesn't make any difference).

\medskip
Let us now show how our original formulation (PNC1) can be converted
to a more convenient form, by chasing the denominators in the Rayleigh ratios,
and by expressing the objective function in
terms of the {\it trace\/} of a certain matrix.

\medskip
For any $N\times N$ matrix $A$, 
because
\begin{align*}
\transpos{X} A X & = 
\begin{bmatrix}
 \transpos{(X^1)}  \\
 \vdots \\
\transpos{(X^K)}  
\end{bmatrix}
A [X^1 \cdots X^K]   \\
& = 
\begin{pmatrix}
\transpos{(X^1)} A X^1 & \transpos{(X^1)} A X^2 &  \cdots &
\transpos{(X^1)} A X^K \\
\transpos{(X^2)} A X^1 & \transpos{(X^2)} A X^2 &  \cdots &
\transpos{(X^2)} A X^K \\
\vdots & \vdots & \ddots & \vdots \\
\transpos{(X^K)} A X^1 & \transpos{(X^K)} A X^2 &  \cdots &
\transpos{(X^K)} A X^K 
\end{pmatrix},
\end{align*}
we have
\[
\mathrm{tr}(\transpos{X} A X) = \sum_{j = 1}^K \transpos{(X^j)} A X^j ,
\]
and the conditions
\[
 \transpos{(X^i)} A X^j = 0, \quad 1\leq i, j \leq K,\>  i\not= j, 
\]
are equivalent to
\[
\transpos{X} A X= \mathrm{diag}(\transpos{(X^1)} A X^1, \ldots, \transpos{(X^K)} A X^K).
\]
As a consequence, if we assume that
\[
\transpos{(X^1)} A X^1 = \cdots = \transpos{(X^K)} A X^K = \alpha^2,
\]
then we have
\[
\transpos{X} A X = \alpha^2 I,
\]
and if $R$ is any orthogonal $K\times K$ matrix, then
by multiplying on the left by $\transpos{R}$ and on the right by $R$,
we get
\[
\transpos{R}\transpos{X} A XR = \transpos{R} \alpha^2 I R = \alpha^2
\transpos{R} R = \alpha^2 I.
\]
Therefore,  if 
\[
\transpos{X} A X = \alpha^2 I,
\]
then 
\[
\transpos{(XR)} A (XR) = \alpha^2 I,
\]
for any orthogonal $K\times K$ matrix $R$.
Furthermore, because $\mathrm{tr}(AB)  =
\mathrm{tr}(BA)$ for all matrices $A, B$, we have 
\[
 \mathrm{tr}(\transpos{R}\transpos{X} A X R)  =  \mathrm{tr}(\transpos{X} A X).
\]
Since the Rayleigh ratios
\[
\frac{\transpos{(X^j)} L X^j}{\transpos{(X^j)}\Degsym X^j}
\]
are invariant under rescaling by a nonzero number, we have
\begin{align*}
\mu(X) & = 
\mu(X^1, \ldots, X^K) 
 = \sum_{j = 1}^K 
\frac{\transpos{(X^j)} L X^j}{\transpos{(X^j)}\Degsym X^j} \\
& = \mu((\transpos{(X^1)} \Degsym X^1)^{-1/2} X^1, \ldots,
(\transpos{(X^K)} \Degsym X^K)^{-1/2} X^K) \\
& = \sum_{j = 1}^K (\transpos{(X^j)} \Degsym X^j)^{-1/2}
\transpos{(X^j)} L \, (\transpos{(X^j)} \Degsym X^j)^{-1/2} X^j \\
& = \mathrm{tr}(\Lambda^{-1/2}\transpos{X} L X \Lambda^{-1/2}),
\end{align*}
where 
\[
\Lambda = \mathrm{diag}(
\transpos{(X^1)} \Degsym X^1, \ldots, \transpos{(X^K)} \Degsym X^K).
\]
If $\transpos{(X^1)} \Degsym X^1 =  \cdots =  
\transpos{(X^K)} \Degsym X^K = \alpha^2$, 
then $\Lambda = \alpha^2 I_K$,  so $\Lambda$
commutes with any $K\times K$ matrix which implies that
\[
\mathrm{tr}(\Lambda^{-1/2}\transpos{R}\transpos{X} L X R
\Lambda^{-1/2})
= \mathrm{tr}(\transpos{R}\Lambda^{-1/2}\transpos{R}\transpos{X} L X 
\Lambda^{-1/2} R) 
= \mathrm{tr}(\Lambda^{-1/2}\transpos{X} L X \Lambda^{-1/2}),
\]
and thus, 
\[
\mu(X) = \mu(XR),
\]
for any orthogonal $K\times K$ matrix $R$.

\medskip
The condition
\[
 X (\transpos{X} X)^{-1} \transpos{X} \mathbf{1} = \mathbf{1}
\]
is also invariant if  we replace $X$ by $XR$, where $R$ is any
invertible matrix, because
\begin{align*}
 XR (\transpos{(XR)} (XR))^{-1} \transpos{(XR)} \mathbf{1} & = 
 XR (\transpos{R}\transpos{X} X R)^{-1} \transpos{R}\transpos{X}
 \mathbf{1}\\
& =  XR R^{-1}(\transpos{X} X)^{-1}(\transpos{R})^{-1}  \transpos{R}\transpos{X}
 \mathbf{1} \\
& = X (\transpos{X} X)^{-1} \transpos{X} \mathbf{1} = \mathbf{1}.
\end{align*}
In summary we proved the following proposition:

\begin{proposition}
\label{Kway1}
For any orthogonal $K\times K$ matrix $R$, any symmetric $N\times N$
matrix $A$, and any $N\times K$ matrix $X = [X^1 \> \cdots \> X^K]$, the following properties hold:
\begin{enumerate}
\item[(1)]
$\mu(X) =  \mathrm{tr}(\Lambda^{-1/2}\transpos{X} L X \Lambda^{-1/2})$,
where 
\[
\Lambda = \mathrm{diag}(
\transpos{(X^1)} \Degsym X^1, \ldots, \transpos{(X^K)} \Degsym X^K).
\]
\item[(2)]
If
$\transpos{(X^1)} \Degsym X^1 =  \cdots =  
\transpos{(X^K)} \Degsym X^K = \alpha^2$, then
$\mu(X) = \mu(XR)$.
\item[(3)]
The condition $\transpos{X} A X = \alpha^2 I$ is  
preserved if $X$ is replaced by $XR$.
\item[(4)]
 The condition $X (\transpos{X} X)^{-1} \transpos{X} \mathbf{1} =
 \mathbf{1}$ is preserved if $X$ is replaced by $XR$.
\end{enumerate}
\end{proposition}

\medskip
Now, by Proposition \ref{Kway1}(1)  and the
fact that the conditions in PNC1  are scale-invariant, 
we are led to the following  formulation of our problem:

\begin{align*}
& \mathrm{minimize}     &  &   
\mathrm{tr}(\transpos{X} L X) & &  &  &\\
& \mathrm{subject\ to} &  & 
\transpos{(X^i)} \Degsym X^j = 0, \quad 1\leq i, j \leq K,\> 
i\not= j,  & &  & & \\
& & & \transpos{(X^j)}\Degsym X^j = 1, \quad 1\leq j \leq K, & & & &\\
& & & 
 X (\transpos{X} X)^{-1} \transpos{X} \mathbf{1} = \mathbf{1},  & & X\in \s{X}. & & 
\end{align*}

Conditions on lines 2 and 3 can be combined in the equation
\[
\transpos{X} \Degsym X = I,
\]
and, we obtain the following  formulation of our  minimization
problem:

\medskip\noindent
{\bf $K$-way Clustering of a graph using Normalized Cut, Version 2: \\
Problem PNC2}

\begin{align*}
& \mathrm{minimize}     &  &  
\mathrm{tr}(\transpos{X} L X)& &  &  &\\
& \mathrm{subject\ to} &  & 
\transpos{X} \Degsym X = I, 
 & &  & & \\
& & & 
 X (\transpos{X} X)^{-1} \transpos{X} \mathbf{1} = \mathbf{1},  & & X\in \s{X}. & & 
\end{align*}

\medskip
Problem PNC2 is equivalent to problem PNC1 is the sense that for every
minimal solution $(X^1, \ldots, X^K)$ of PNC1, 
$((\transpos{(X^1)} D X^1)^{-1/2} X^1, \ldots,
(\transpos{(X^K)} D X^K)^{-1/2} X^K)$
is a minimal solution of PNC2 (with the same minimum for the objective functions),
and that for every minimal solution $(Z^1, \ldots, Z^k)$ of PNC2,  
$(\lambda_1 Z^1, \ldots, \lambda_K Z^K)$ 
is a minimal solution of PNC1, for all  $\lambda_i \not= 0$,  $i = 1,
\ldots, K$ (with the same minimum for the objective functions).
In other words, problems PNC1   and PNC2 have the same
set of minimal solutions as $K$-tuples of points 
$(\mathbb{P}(X^1), \ldots, \mathbb{P}(X^K))$ in
$\mathbb{RP}^{N-1}$ determined by
their  homogeneous coordinates $X^1, \ldots, X^K$.

\medskip
Formulation PNC2 reveals that finding a minimum normalized cut 
has a geometric interpretation in terms of the graph drawings 
discussed in Section  \ref{ch2-sec1}. Indeed, PNC2 has the following
equivalent formulation: Find a minimal energy graph drawing $X$ in $\reals^K$ of the
weighted graph $G = (V, W)$ such that:
\begin{enumerate}
\item
The matrix $X$ is orthogonal with respect to the inner product
$\lag -, - \rag_{\Degsym}$ in $\reals^N$ induced by $\Degsym$, with
\[
\lag x, y \rag_{\Degsym} = \transpos{x} \Degsym y, \quad x, y \in \reals^N. 
\]
\item
The rows of $X$ are nonzero; this means that no vertex $v_i\in V$ is
assigned to the origin of $\reals^K$ (the zero vector $0_K$).
\item
Every vertex $v_i$ is assigned a point of the form 
$(0, \ldots, 0,a_j,0,\ldots,0)$  on some axis (in $\reals^K$).
\item
Every axis in  $\reals^K$ is assigned at least some vertex.
\end{enumerate}

\medskip
Condition 1 can be reduced to the standard condition for graph
drawings ($\transpos{R}R = I$)  by making the change of variable $Y = \Degsym^{1/2} X$ or equivalently
$X = \Degsym^{-1/2} Y$. Indeed, 
\[
\mathrm{tr}(\transpos{X} LX) = 
\mathrm{tr}
(\transpos{Y}\Degsym^{-1/2} L \Degsym^{-1/2} Y),
\]
so we use the normalized Laplacian $L_{\mathrm{sym}} = \Degsym^{-1/2} L \Degsym^{-1/2}$
instead of $L$, 
\[
\transpos{X} \Degsym X   = \transpos{Y} Y   = I,
\]
and conditions (2), (3), (4) are preserved under the change of
variable  $Y = \Degsym^{1/2} X$, since  $\Degsym^{1/2} $ is invertible.
However, conditions (2), (3), (4) are ``hard'' constraints, especially
condition (3). 
In fact, condition (3) implies that the columns of $X$ are  orthogonal
with respect to  both the Euclidean inner product and the inner
product $\lag -, - \rag_{\Degsym}$, so condition (1) is redundant,
except for the fact  that it prescribes the norm of the columns, but 
this is not essential due to the projective nature of the solutions.

\medskip
The main problem in finding a good relaxation of
problem PNC2 is that it is very difficult to enforce the condition
$X\in \s{X}$.
Also, the solutions $X$ are not preserved under arbitrary
rotations,
but only by very special rotations which leave $\s{X}$ invariant
(they exchange the axes).
The first natural relaxation of problem PNC2 is to drop the condition
that $X\in \s{X}$, and we obtain the 

\medskip\noindent
{\bf Problem $(*_1)$}

\begin{align*}
& \mathrm{minimize}     &  &  
\mathrm{tr}(\transpos{X} L X)& &  &  &\\
& \mathrm{subject\ to} &  & 
\transpos{X} \Degsym X = I, 
 & &  & & \\
& & & 
X (\transpos{X} X)^{-1} \transpos{X} \mathbf{1} = \mathbf{1}.  & &  & & 
\end{align*}

\medskip
By Proposition \ref{Kway1}, for every
orthogonal matrix $R\in \mathbf{O}(K)$  and 
for every $X$ minimizing $(*_1)$, the matrix $XR$ also
minimizes $(*_1)$.  As a consequence, as explained below, 
we can view the solutions of
problem $(*_1)$ as elements of the {\it Grassmannian\/} $G(N, K)$.

\medskip
Recall that the {\it Stiefel manifold\/} $St(k, n)$ consists of the
set of orthogonal $k$-frames in $\reals^n$, that is, the 
$k$-tuples of orthonormal vectors $(u_1, \ldots, u_k)$ with $u_i\in \reals^n$.
For $k = n$, the manifold $St(n, n)$ is identical to the orthogonal
group $\mathbf{O}(n)$. For $1 \leq n \leq n - 1$,  the group
$\mathbf{SO}(n)$ acts transitively on $St(k, n)$, and $St(k, n)$ is
isomorphic to the coset manifold $\mathbf{SO}(n)/\mathbf{SO}(n - k)$.
The {\it Grassmann manifold\/} $G(k, n)$ consists of all (linear) 
$k$-dimensional subspaces of $\reals^n$. Again, 
the group $\mathbf{SO}(n)$ acts transitively on $G(k, n)$, and $G(k, n)$ is
isomorphic to the coset manifold $\mathbf{SO}(n)/S(\mathbf{SO}(k)\times\mathbf{SO}(n - k))$.
The  group $\mathbf{O}(k)$ acts on the right on the Stiefel manifold
$St(k, n)$ (by multiplication), and the orbit manifold
$St(k,n)/\mathbf{O}(k)$ is isomorphic to the Grassmann manifold $G(k,
n)$. Furthermore, both $St(k, n)$ and $G(k, n)$ are {\it naturally
reductive homogeneous manifolds\/} (for the Stiefel manifold, when $n \geq 3$), and
$G(k, n)$ is even a {\it symmetric space\/}
(see O'Neill \cite{Oneill}). The upshot of all this is
that to a large extent, the differential geometry of these manifolds
is completely determined by some subspace $\mfrac{m}$ of the Lie algebra
$\mfrac{so}(n)$, such that we have a direct sum 
\[
\mfrac{so}(n) = \mfrac{m} \oplus \mfrac{h},
\]
where $\mfrac{h} = \mfrac{so}(n - k)$ in the case of the Stiefel
manifold, and $\mfrac{h} = \mfrac{so}(k) \times \mfrac{so}(n - k)$ in
the case of the Grassmannian manifold
(some additional condition on $\mfrac{m}$ is required). 
In particular,  the geodesics in both manifolds can be determined
quite explicitly, and thus we obtain closed form formulae for
distances, {\it etc}.

\medskip
The Stiefel manifold $St(k, n)$ can be viewed as the set of all $n
\times k$ matrices $X$ such that
\[
\transpos{X}  X = I_k.
\]
In our situation, we are considering $N\times K$ matrices $X$ such
that
\[
\transpos{X} D X = I.
\]
This is not quite the Stiefel manifold, but if we write 
$Y = D^{1/2} X$, then we have
\[
\transpos{Y} Y = I,
\]
so the space of matrices $X$ satisfying the condition 
$\transpos{X} D X = I$ is the image $\s{D}(St(K, N))$ of the
Stiefel manifold $St(K, N)$ under the linear map $\s{D}$ given by
\[
\s{D}(X) = D^{1/2} X.
\]
Now, the right action of $\mathbf{O}(K)$ on $\s{D}(St(K, N))$  yields a coset
manifold
$\s{D}(St(K, N))/\mathbf{O}(K)$ which is obviously isomorphic to the
Grassmann manidold $G(K, N)$. 

\medskip
Therefore, 
{\it the solutions of
problem $(*_1)$ can be viewed 
as elements of the {\it Grassmannian\/} $G(N, K)$\/}.
We can take advantage of this fact to find a discrete solution of our
original optimization problem PNC2 
approximated by  a continuous solution of $(*_1)$. 

\medskip
Recall that condition
$X (\transpos{X} X)^{-1} \transpos{X} \mathbf{1} = \mathbf{1}$
is equivalent to $XX^+ \mathbf{1} = \mathbf{1}$, which is also
equivalent to the fact that $\mathbf{1}$ is in the range of $X$.
If we make the change of variable $Y = \Degsym^{1/2} X$ or equivalently
$X = \Degsym^{-1/2} Y$,  the condition that 
$\mathbf{1}$ is in the range of $X$ becomes the condition that
$\Degsym^{1/2} \mathbf{1}$ is in the range of $Y$, which is equivalent
to
\[
Y Y^+ \Degsym^{1/2} \mathbf{1} = \Degsym^{1/2} \mathbf{1}.
\]
However, since $\transpos{Y} Y = I$, we have
\[
Y^+ = \transpos{Y},
\]
so we get the equivalent problem

\medskip\noindent
{\bf Problem $(**_1)$}

\begin{align*}
& \mathrm{minimize}     &  &  
\mathrm{tr}(\transpos{Y}\Degsym^{-1/2} L \Degsym^{-1/2} Y)& &  &  &\\
& \mathrm{subject\ to} &  & 
\transpos{Y} Y = I, 
 & &  & & \\
& & & 
Y  \transpos{Y} \Degsym^{1/2} \mathbf{1}  = \Degsym^{1/2} \mathbf{1}.  & &  & & 
\end{align*}

\medskip
This time, the matrices $Y$ satisfying condition $\transpos{Y} Y = I$
do belong to the Stiefel manifold $St(K,  N)$, and again, {\it we view the
solutions of problem $(**_1)$ as elements of the Grassmannian $G(K, N)$\/}.
We pass from a solution $Y$ of problem $(**_1)$ in $G(K, N)$ to a
solution $Z$ of of problem $(*_1)$ in $G(K, N)$  by the linear map
$\s{D}^{-1}$; namely, $Z = \s{D}(Y) = D^{-1/2} Y$.

\medskip
The Rayleigh--Ritz Theorem (see Proposition \ref{PCAlem1})
tells us that if we temporarily ignore the second constraint, 
minima
of  problem $(**_1)$
are obtained by
picking any $K$ unit eigenvectors $(u_1, \ldots, u_k)$ associated with the smallest
eigenvalues
\[
0 = \nu_1\leq  \nu_2 \leq  \ldots \leq  \nu_K
\]
of $L_{\mathrm{sym}} = \Degsym^{-1/2} L \Degsym^{-1/2}$.
We may assume that $\nu_2  > 0$, namely that the underlying graph is
connected (otherwise, we work with each connected component), in which
case $Y^1 = \Degsym^{1/2}\mathbf{1}/\norme{\Degsym^{1/2}\mathbf{1}}_2$, 
because $\mathbf{1}$ is in the nullspace of $L$.
Since $Y^1 =
\Degsym^{1/2}\mathbf{1}/\norme{\Degsym^{1/2}\mathbf{1}}_2$,
the vector $\Degsym^{1/2}\mathbf{1}$ is in the range of $Y$, so the
condition
\[
Y  \transpos{Y} \Degsym^{1/2} \mathbf{1}  = \Degsym^{1/2} \mathbf{1}
\]
is also satisfied.
Then,
$Z  = \Degsym^{-1/2}Y$ with
$Y = [u_1\> \ldots\>  u_K]$ yields a minimum of our
relaxed problem $(*_1)$ (the second constraint is satisfied because
$\mathbf{1}$ is in the range of $Z$).

\medskip
By Proposition \ref{Laplace3}, the vectors
$Z^j$  are
eigenvectors of $L_{\mathrm{rw}}$ associated with the eigenvalues
$0 = \nu_1 \leq  \nu_2 \leq  \ldots \leq  \nu_K$.
Recall that $\mathbf{1}$ is an eigenvector for the eigenvalue $\nu_1 = 0$,
and  
$Z^1 = \mathbf{1}/\norme{\Degsym^{1/2}\mathbf{1}}_2$.
Because, $\transpos{(Y^i)}Y^j = 0$ whenever $i \not= j$, 
we have
\[
\transpos{(Z^i)}\Degsym Z^j = 0, \quad\text{whenever $i \not= j$}.
\]
This implies that  $Z^2, \ldots, Z^K$ are all orthogonal to
$\Degsym\mathbf{1}$,
and thus, that each $Z^j$ has both some positive and some negative
coordinate, for $j = 2, \ldots, K$. 

\medskip
The conditions
$\transpos{(Z^i)}\Degsym Z^j = 0$ do not necessarily imply that
$Z^i$ and $Z^j$ are orthogonal (w.r.t. the Euclidean inner product),
but we can obtain a solution of Problem $(*_1)$ achieving the same
minimum for which distinct columns $Z^i$ an $Z^j$ are simultaneoulsy
orthogonal and $\Degsym$-orthogonal, 
by multiplying $Z$ by some
$K\times K$ orthogonal matrix $R$ on the right.
Indeed, the $K\times K$ symmetric matrix $\transpos{Z}Z$ can be
diagonalized by some orthogonal $K\times K$ matrix $R$ as  
\[
\transpos{Z}Z = R \Sigma \transpos{R},
\]
where $\Sigma$ is a diagonal matrix, 
and thus,
\[
\transpos{R}\transpos{Z} Z R = \transpos{(ZR)} ZR = \Sigma, 
\]
which shows that the columns of $ZR$ are orthogonal.
By Proposition \ref{Kway1}, $ZR$ also satisfies the 
constraints of $(*_1)$, and
$\mathrm{tr}(\transpos{(ZR)} L (ZR)) = \mathrm{tr}(\transpos{Z} L Z)$.

\medskip

\remark
Since $Y$ has linearly independent columns (in fact,
orthogonal) and since $Z = D^{-1/2} Y$, the matrix $Z$ also has linearly
independent columns, so $\transpos{Z} Z$ is positive definite and 
the entries in $\Sigma$ are all positive.

\medskip
In summary,   we  should look for a solution $X$ that corresponds to 
an element of the Grassmannian $G(K, N)$, and
hope that  for some suitable orthogonal matrix $R$,
the vectors in  $XR$  are close to a true solution of the original problem.

\section[$K$-Way Clustering; Using The Dependencies Among $X^1, \ldots, X^K$]
{$K$-Way Clustering; Using The Dependencies \\
Among $X^1, \ldots, X^K$}
\label{ch3-sec4}
At this stage, it is interesting to reconsider the case $K = 2$ in the
light of what we just did when $K \geq 3$. When $K = 2$, $X^1$ and
$X^2$ are not independent, and it is convenient  to assume that the nonzero
entries in $X^1$ and $X^2$ are both equal to some positive real  $c\in
\reals$, so that 
\[
X^1 + X^2 = c \mathbf{1}.
\]
To avoid subscripts, write  $(A, \overline{A})$
for the partition of $V$ that we are seeking, and as before let
$d = \transpos{\mathbf{1}} \Degsym \mathbf{1}$ and $\alpha = \mathrm{vol}(A)$.
We know from Section 
\ref{ch3-sec2} that
\begin{align*}
\transpos{(X^1)} \Degsym X^1& = \alpha c^2 \\
\transpos{(X^2)} \Degsym X^2 &=  (d - \alpha) c^2,
\end{align*}
so we normalize $X^1$ and $X^2$ so that $\transpos{(X^1)} \Degsym X^1
= \transpos{(X^2)} \Degsym X^2 = c^2$, and we consider 
\[
X = \left[\frac{X^1}{\sqrt{\alpha}} \> \frac{X^2}{\sqrt{d - \alpha}}\right]. 
\]
Now, we claim that there is an orthogonal matrix $R$ so that if $X$ as above
is a solution to our discrete problem, then
$X R$ contains a multiple of $\mathbf{1}$ as a first column.
A similar observation is  made in Yu \cite{Yu} and Yu and Shi
\cite{YuShi2003}
(but beware that in these works $\alpha = \mathrm{vol}(A)/\sqrt{d}$).
In fact,  
\[
R = 
\frac{1}{\sqrt{d}} 
\begin{pmatrix}
 \sqrt{\alpha} & \sqrt{d - \alpha} \\[6pt]
\sqrt{d - \alpha} & - \sqrt{\alpha} 
\end{pmatrix}.
\]
Indeed, we have
\begin{align*}
X R & =  \left[\frac{X^1}{\sqrt{\alpha}} \> \frac{c \mathbf{1} - X^1}{\sqrt{d -
      \alpha}}\right] R \\
& =  \left[\frac{X^1}{\sqrt{\alpha}} \> \frac{c \mathbf{1} - X^1}{\sqrt{d -
      \alpha}}\right]
\frac{1}{\sqrt{d}} 
\begin{pmatrix}
 \sqrt{\alpha} & \sqrt{d - \alpha} \\[6pt]
\sqrt{d - \alpha} & - \sqrt{\alpha} 
\end{pmatrix} \\
& = \frac{1}{\sqrt{d}} 
\left[
c \mathbf{1} \>\> \sqrt{\frac{d - \alpha}{\alpha}}\, X^1 
- \sqrt{\frac {\alpha}{d   - \alpha}}\,(c \mathbf{1} - X^1)
\right].
\end{align*}
If we let
\[
a = c\sqrt{\frac{d - \alpha}{\alpha}}, \quad b = - c\sqrt{\frac {\alpha}{d   - \alpha}},
\]
then we check that
\[
\alpha a + b (d - \alpha) = 0, 
\]
which shows that the vector
\[
Z = \sqrt{\frac{d - \alpha}{\alpha}}\, X^1 
- \sqrt{\frac {\alpha}{d   - \alpha}}\,(c \mathbf{1} - X^1)
\]
is a potential solution of our discrete problem in the sense of Section
\ref{ch3-sec2}.
Furthermore, because $L \mathbf{1} = 0$, 
\[
\mathrm{tr}(\transpos{X} L X) = \mathrm{tr}(\transpos{(XR)} L (XR))  = \transpos{Z} L Z, 
\]
the vector $Z$ is indeed a solution of our discrete problem. 
Thus, we reconfirm the fact that the second eigenvector of
$L _{\mathrm{rw}} = \Degsym^{-1}  L$
is indeed a continuous approximation to the clustering problem when
$K = 2$. This can be generalized for any $K \geq 2$.

\medskip
Again, we may assume that the nonzero entries in $X^1, \ldots, X^K$
are some positive real $c\in \reals$, so that
\[
X^1 + \cdots + X^K = c \mathbf{1},
\]
and if $(A_1, \ldots, A_K)$ is
the partition of $V$ that we are seeking, write
$\alpha_j = \mathrm{vol}(A_j)$. We have
$\alpha_1 + \cdots + \alpha_K = d = \transpos{\mathbf{1}} \Degsym \mathbf{1}$. 
Since
\[
\transpos{(X^j)} \Degsym X^j = \alpha_j c^2,
\]
we normalize the  $X^j$ so that $\transpos{(X^j)} \Degsym X^j
= \cdots = \transpos{(X^K)} \Degsym X^K = c^2$, and we consider 
\[
X = \left[\frac{X^1}{\sqrt{\alpha_1}} \> \frac{X^2}{\sqrt{\alpha_2}}
  \>\cdots \> \> \frac{X^K}{\sqrt{\alpha_K}}\right]. 
\]
Then, we have the following result.

\begin{proposition}
\label{propdep1}
If $X = \left[\frac{X^1}{\sqrt{\alpha_1}} \> \frac{X^2}{\sqrt{\alpha_2}}
  \>\cdots \> \> \frac{X^K}{\sqrt{\alpha_K}}\right]$ is a solution of
our discrete problem, then there is an orthogonal matrix $R$ such that
its first column  $R^1$ is
\[
R^1 =
\frac{1}{\sqrt{d}}
\begin{pmatrix}
\sqrt{\alpha_1} \\ 
\sqrt{\alpha_2} \\
\vdots             \\
\sqrt{\alpha_K}
\end{pmatrix}
\] 
and
\[
XR = \left[\frac{c}{\sqrt{d}}  \mathbf{1} \> Z^2 \>\cdots \> Z^{K}\right]. 
\]
Furthermore, 
\[
\transpos{(XR)}\Degsym (XR) = c^2 I
\]
and
\[
\mathrm{tr} (\transpos{(XR)} L (XR)) = \mathrm{tr}(\transpos{Z} L Z),
\]
with $Z = [Z^2 \> \cdots \> Z^{K}]$.
\end{proposition}
\begin{proof}
Apply  Gram--Schmidt to $(R^1, e_2, \ldots, e_K)$ (where $(e_1,
\ldots, e_K)$ is the canonical basis of $\reals^K$) to
form an orthonormal basis. The rest follows from Proposition \ref{Kway1}.
\end{proof}

Proposition \ref{propdep1} suggests that if $Z = [\mathbf{1}\> Z^2 \>\cdots\> Z^K]$
is a solution of the relaxed problem $(*_1)$, then  there should be
an orthogonal matrix $R$ 
such that $Z\transpos{R}$ is an approximation of a 
solution of the discrete problem PNC1.

\medskip
The next step is to find 
an exact solution  $(\mathbb{P}(X^1), \ldots, \mathbb{P}(X^K)) \in \mathbb{P}(\s{K})$
which is the closest (in a suitable sense) to our
approximate solution $(Z^1, \ldots, Z^K)\in G(K, N)$.
The set $\s{K}$ is not necessarily closed under all orthogonal
transformations
in $\mathbf{O}(K)$, so we can't view $\s{K}$ as a subset of the
Grassmannian $G(K, N)$. However, we can think of   $\s{K}$ as a
subset of  $G(K, N)$ by considering  the subspace 
spanned by $(X^1, \ldots, X^K)$ for every $[X^1\> \cdots X^K\> ] \in \s{K}$.
Then, we have two choices of
distances.
\begin{enumerate}
\item
We view $\s{K}$ as a subset of $(\mathbb{RP}^{N-1})^K$.
Because $\s{K}$ is closed under the antipodal map, as
explained in Appendix \ref{ch3-sec6}, minimizing the distance
$d(\mathbb{P}(X^j), \mathbb{P}(Z^j))$ on $\mathbb{RP}^{N-1}$ is equivalent
to minimizing the Euclidean distance $\norme{X^j - Z^j}_2$, for $j =
1, \ldots, K$ (if we use the Riemannian metric on $\mathbb{RP}^{N-1}$ induced by the Euclidean
metric on $\reals^N$).
Then, minimizing the distance $d(X, Z)$ in  $(\mathbb{RP}^{N-1})^K$
is equivalent to minimizing $\norme{X - Z}_F$, where
\[
\norme{X - Z}_F^2 =  \sum_{j = 1}^K \norme{X^j - Z^j}_2^2
\]
is  the Frobenius norm. This is implicitly the choice made by Yu. 
\item
We view $\s{K}$ as a subset of the Grassmannian $G(K, N)$. 
In this case, we need to pick a metric on the Grassmannian,
and we minimize the corresponding  Riemannian distance $d(X, Z)$.
A natural choice is the metric on $\mfrac{se}(n)$ given by
\[
\lag X, Y\rag = \mathrm{tr}(\transpos{X} Y).
\]
This choice remains to be explored, and will be the subject of 
a forthcoming report.
\end{enumerate}

\section[Discrete Solution Close to a Continuous  Approximation]
{Finding a Discrete Solution Close to a Continuous  Approximation}
\label{ch3-sec5}
Inspired by Yu \cite{Yu} and the previous section,
given a solution $Z_0$ of problem $(*_1)$, 
we  look for pairs
$(X, R) \in \s{K}\times \mathbf{O}(K)$ (where $R$ is a $K\times K$
orthogonal matrix), with $\norme{X^j} = \norme{Z^j_0}$ for $j = 1, \ldots, K$, 
that minimize
\[
\varphi(X, R) = \norme{X - Z_0R}_F.
\]
Here, $\norme{A}_F$ is the Frobenius norm of $A$, with
$\norme{A}_F^2 = \mathrm{tr}(\transpos{A} A)$.

\medskip
It may seem desirable to look for discrete solutions $X\in \s{K}$
whose entries are $0$ or $1$, in which case
\[
X \mathbf{1}_K = \mathbf{1}_N.
\]
Therefore, we begin by finding a diagonal matrix
 $\Lambda =
\mathrm{diag}(\lambda_1, \ldots, \lambda_K)$ 
such that
\[
\norme{Z_0\Lambda\mathbf{1}_K - \mathbf{1}_N}_2
\]
is minimal in the least-square sense. 
As we remarked earlier, since the columns of $Z_0$ are orthogonal
with respect to the inner product $\lag u, v\rag_{\Degsym} = \transpos{x}\Degsym y$, 
they are linearly independent, thus the pseudo-inverse of $Z_0$ is
$(\transpos{Z_0} Z_0)^{-1} \transpos{Z_0}$,  and the  best solution
$(\lambda_1, \ldots, \lambda_K)$ of least Euclidean norm is
given by
\[
 (\transpos{Z_0} Z_0)^{-1} \transpos{Z_0}\mathbf{1}_N.
\]
Therefore, we form the (column-rescaled) matrix
\[
Z = 
Z_0 \,\mathrm{diag}( (\transpos{Z_0} Z_0)^{-1} \transpos{Z_0}\mathbf{1}_N).
\]

\remark
In Yu \cite{Yu} and Yu and Shi \cite{YuShi2003},  the rows of $Z_0$ are
normalized by forming the matrix
\[
\mathrm{DIAG}(Z_0\transpos{Z_0})^{-1/2} Z_0.
\]
However, this does not yield a matrix whose  columns are obtained from
those of  $Z_0$ by rescaling, so the resulting matrix is no longer a
rescale of a correct solution of problem
$(*_1)$, which seems undesirable.

\medskip
Actually, even though the columns of $Z$ are  $\Degsym$-orthogonal,
the matrix $Z$ generally does not satisfy
the condition $\transpos{Z}\Degsym Z = I$, so $ZR$ may not have
$\Degsym$-orthogonal columns (with $R \in \mathbf{O}(K)$), yet
$\mathrm{tr}(\transpos{Z} L Z) = \mathrm{tr}(\transpos{(ZR)} L (ZR))$ holds! 
The problem is
that the conditions $\transpos{Z}\Degsym Z = I$ and
$Z  \mathbf{1} = \mathbf{1}$
are antagonistic.
If we try to force condition $Z   \mathbf{1} = \mathbf{1}$,
we modify the $\Degsym$-norm of the columns of $Z$, and then
$ZR$ may no longer have $\Degsym$-orthogonal columns.
Unless these methods are implemented and tested, it seems almost
impossible
to tell which option yields the best result. We will proceed under
the assumption that $Z_0$ has been rescaled as explained above, but
the method described next also applies if we pick $Z = Z_0$ .
In this latter case,  by Proposition \ref{Kway1}, 
$Z$ satisfies the condition 
$\transpos{Z}\Degsym Z = I$, and so does $ZR$.

\medskip 
The key to minimizing $\norme{X - ZR}_F$  rests on the following computation:
\begin{align*}
\norme{X - ZR}_F^2 & = \mathrm{tr}(\transpos{(X - ZR)}(X - ZR)) \\
& = \mathrm{tr}((\transpos{X} -\transpos{R}\transpos{Z})(X - ZR)) \\
& = \mathrm{tr}(\transpos{X}X- \transpos{X}ZR - \transpos{R}\transpos{Z}X
+ \transpos{R}\transpos{Z}ZR)\\
& = \mathrm{tr}(\transpos{X}X)  -
\mathrm{tr}(\transpos{X}ZR)   - \mathrm{tr}(\transpos{R}\transpos{Z}X)  
+ \mathrm{tr}(\transpos{R}\transpos{Z}ZR)\\
& = \mathrm{tr}(\transpos{X}X)  -
\mathrm{tr}(\transpos{(\transpos{R}\transpos{Z}X)})  
- \mathrm{tr}(\transpos{R}\transpos{Z}X)  
+ \mathrm{tr}(\transpos{Z}ZR \transpos{R})\\
& = \mathrm{tr}(\transpos{X}X)  - 2\mathrm{tr}(\transpos{R}\transpos{Z}X)   
+ \mathrm{tr}(\transpos{Z}Z).
\end{align*}
Therefore, minimizing $\norme{X - ZR}_F^2$ is equivalent to maximizing
 $\mathrm{tr}(\transpos{R}\transpos{Z}X)$. 
This will be done by alternating steps during which we minimize 
$\varphi(X, R) = \norme{X - ZR}_F$ with respect to $X$ holding $R$
fixed, and steps during which we minimize 
$\varphi(X, R) = \norme{X - ZR}_F$ with respect to $R$ holding $X$
fixed. 
For this second step, we  need the following
proposition.

\begin{proposition}
\label{Kway2}
For any $n \times n$ matrix $A$ and  any orthogonal matrix $Q$, we
have
\[
\max\{\mathrm{tr}(QA) \mid Q\in \mathbf{O}(n)\} 
= \sigma_1  + \cdots + \sigma_n,
\]
where $\sigma_1 \geq \cdots \geq \sigma_n$ are the singular values of
$A$.
Furthermore, this maximum is achieved by  $Q =
V\transpos{U}$,
where $A = U \Sigma \transpos{V}$ is any SVD for $A$.
\end{proposition}

\begin{proof}
Let $A = U \Sigma \transpos{V}$ be any SVD for $A$. Then we have
\begin{align*}
\mathrm{tr}(QA) &  = \mathrm{tr}(Q U \Sigma \transpos{V})\\
&  = \mathrm{tr}(\transpos{V}  Q U \Sigma). 
\end{align*}
The matrix $Z = \transpos{V}  Q U$ is an orthogonal matrix 
so $|z_{i j}| \leq 1$ for $1\leq i, j \leq n$, and
$\Sigma$ is a diagonal matrix, so we have
\[
\mathrm{tr}(Z\Sigma) = z_{1 1} \sigma_1 + \cdots + z_{n n } \sigma_n
\leq \sigma_1 + \cdots + \sigma_n, 
\]
which proves the first statement of the proposition.
For $Q = V\transpos{U}$, we get
\begin{align*}
\mathrm{tr}(QA)  &  = \mathrm{tr}(Q U \Sigma \transpos{V}) \\
 &  = \mathrm{tr}(V\transpos{U} U \Sigma \transpos{V}) \\
&  = \mathrm{tr}(V\Sigma \transpos{V}) = \sigma_1  + \cdots + \sigma_n,
\end{align*}
which proves the second part of the proposition.
\end{proof}

As a corollary of Proposition \ref{Kway2} (with $A = \transpos{Z} X$
and $Q = \transpos{R}$), we get the following
result (see  Golub and Van Loan \cite{Golub}, Section 12.4.1):

\begin{proposition}
\label{Kway3}
For any two fixed $N\times K$ matrices $X$ and $Z$,  the minimum of
the set
\[
\{\norme{X - ZR}_F \mid R\in \mathbf{O}(K)\}
\]
is achieved by $R = U \transpos{V}$, for any SVD decomposition 
$U \Sigma \transpos{V} = \transpos{Z} X$ of $\transpos{Z} X$.
\end{proposition}

\medskip
We now deal with step 1.
The solutions $Z$  of the relaxed problem $(*_1)$ have
columns $Z^j$ of  norm  $\rho_j$.  Then, 
for fixed $Z$ and $R$, we would like to find some $X\in \s{K}$
with $\norme{X^j} = \norme{Z^j} = \rho_j$ for $j = 1, \ldots, K$, 
so that $\norme{X - ZR}_F$ is minimal. 
Without loss of generality, we may assume that the entries
$a_1, \ldots, a_K$ occurring in the matrix $X$ are positive.
To find $X\in \s{K}$, first we find the shape $\widehat{X}$ of $X$, which
is the matrix obtained from $X$ by rescaling the columns of $X$ so
that $\widehat{X}$ has entries $+1,  0$. Then, we rescale the columns of $\widehat{X}$
so that $\norme{X^j} = \rho_j$ for $j = 1, \ldots, K$.

\medskip
Since
\[
\norme{X - ZR}_F^2 = \norme{X}_F^2 + \norme{Z}_F^2 -
\mathrm{tr}(\transpos{R}\transpos{Z}X)
= 2\sum_{j = 1}^K\rho_j^2  - \mathrm{tr}(\transpos{R}\transpos{Z}X),
\]
minimizing $\norme{X - ZR}_F$ is equivalent to maximizing 
\[
\mathrm{tr}(\transpos{R}\transpos{Z}X) =
\mathrm{tr}(\transpos{(ZR)}X) =\mathrm{tr}(X\transpos{(ZR)}),
\]
and since the $i$th row of $X$ contains a single nonzero entry, say 
$a_{j_i}$ (in column $j_i$, $1\leq j_i \leq K$), if we write $Y = ZR$, then 
\begin{equation}
\mathrm{tr}(X \transpos{Y}) = \sum_{i = 1}^N a_{j_i} y_{i\, j_i }. 
\tag{$*$}
\end{equation}
By $(*)$, $\mathrm{tr}(X \transpos{Y})$ is maximized iff  $a_{j_i}y_{i\, j_i }$ is 
maximized for $i = 1, \ldots, N$. Since the $a_k$ are positive, 
this is achieved if, for the  $i$th
row of $X$, we pick a column index $\ell$ such that $y_{i\, \ell}$ is
maximum.

\medskip
Observe that if we change the $\rho_j$s, 
minimal solutions for these new values of the $\rho_j$ are obtained
by rescaling the $a_\ell$'s. Thus, 
to find the shape $\widehat{X}$ of $X$,
we may assume that $a_{\ell} = 1$.  

\medskip
Actually, to find the shape
$\widehat{X}$ of $X$, we first find a matrix $\overline{X}$ according
to the following method.
If we let
\begin{align*}
\mu_i & = \max_{1 \leq j \leq K} y_{i j} \\
J_i & = \{j \in \{1, \ldots, K\} \mid y_{i j} = \mu_i\},
\end{align*}
 for $i = 1, \ldots, N$, 
then 
\[
\overline{x}_{i j} =
\begin{cases}
+1 & \text{for some chosen  $j \in J_i$,} \\
0 & \text{otherwise}.
\end{cases}
\]
Of course, a single column index is chosen for each
row. Unfortunately, the matrix
$\overline{X}$ may not be a correct solution, because the above prescription does
not guarantee that every column of $\overline{X}$ is nonzero. Therefore, we may
have to reassign certain nonzero entries in columns having ``many''
nonzero entries to zero columns, so that we get a matrix in $\s{K}$.
When we do so, we set the nonzero entry  in the column from which it
is moved to zero. This new matrix is  $\widehat{X}$, and finally we
normalize each column of $\widehat{X}$ to obtain $X$, so that
$\norme{X^j} = \rho_j$, for $j = 1, \ldots, K$. This last step may not
be necessary since $Z$ was chosen so that $\norme{Z\mathbf{1}_K -
  \mathbf{1}_N}_2$ is miminal. A practical way to deal with zero
columns in $\overline{X}$ is to simply decrease $K$. 
Clearly, further work is needed to justify the soundness of such a method.

\medskip
The above method is essentially the method described in Yu \cite{Yu}
and Yu and Shi \cite{YuShi2003}, except that in these works
(in which  $X, Z$ and $Y$ are denoted by $X^*,
\widetilde{X}^*$, and $\widetilde{X}$, respectively)
the entries in $X$
belong to $\{0, 1\}$;  as described above,
 for row $i$, the index $\ell$ 
corresponding to the entry $+1$ is given by
\[
\arg \max_{1\leq j \leq K} \widetilde{X}(i, j).
\]
The fact that  $\overline{X}$ may have zero columns is  not
addressed by Yu.
Furthermore, it is important to make sure that each column of $X$ has
the same norm as the corresponding column of  $ZR$, but this normalization step
is not performed in the above works. On the hand, the rows of $Z$ are
normalized, but the resulting matrix may no longer
be a correct solution of the relaxed problem.
Only a comparison of tests
obtained by implementating both methods will reveal which method
works best in practice.

\medskip
The method due to Yu and Shi (see Yu \cite{Yu} and Yu and Shi
\cite{YuShi2003})  
to find $X\in \s{K}$ and $R\in \mathbf{O}(K)$ that minimize
$\varphi(X, R) = \norme{X - ZR}_F$ is 
to alternate steps during which either $R$ is held fixed (step PODX) or $X$ is
held fixed (step PODR). 

\begin{enumerate}
\item[(1)]
In step PODX, the next discrete solution $X^*$ is obtained fom the previous
pair $(R^*, Z)$ by computing $\overline{X}$ and then $X^* =
\widehat{X}$ from $Y = ZR^*$,
as just explained above.
\item[(2)]
In step PODR, the next matrix $R^*$ is obtained from the previous pair
$(X^*, Z)$ by
\[
R^* =  U \transpos{V},
\]
 for any SVD decomposition 
$U \Sigma \transpos{V}$ of $\transpos{Z} X^*$.
\end{enumerate}

\medskip
It remains to initialize $R^*$ to start the process, and then steps
(1) and (2) are iterated, starting with step (1). 
The method advocated by Yu \cite{Yu} is to pick $K$ rows of $Z$ that
are as orthogonal to each other as possible. This corresponds to a
$K$-means clustering strategy with $K$ nearly orthogonal data points
as centers. Here is the algorithm given in Yu \cite{Yu}.

\medskip
Given the $N\times K$ matrix $Z$ (whose columns all have the same
norm), we compute a matrix $R$ whose columns are  certain rows of $Z$.
We use a vector $c\in \reals^N$ to keep track of the inner products
of all rows of $Z$ with the columns $R^1, \ldots, R^{k-1}$ that have been
constructed so far, and initially when $k = 1$, we set $c = 0$.

\medskip
The first column $R^1$ of $R$ is any chosen row of $Z$.

\medskip
Next, for $k = 2, \ldots, K$, we compute all the inner products of
$R^{k-1}$ with all rows in $Z$, which are recorded in the vector
$ZR^{k-1}$, and  we update $c$ as follows:
\[
c = c + \mathtt{abs}(ZR^{k - 1}).
\]
We take the  absolute values of the entries in $ZR^{k - 1}$
so that the $i$th entry in $c$ is  a score of how orthogonal
is the $i$th row  of $Z$ to $R^1, \ldots, R^{k-1}$.
Then, we  choose $R^k$ as any row $Z_i$ of $Z$ for which $c_i$ is minimal
(the customary (and ambiguous) $i = \arg \min c$).

\appendix
\chapter{The Rayleigh Ratio and the Courant-Fischer Theorem}
\label{Rayleigh-Ritz} 
The most important property of symmetric matrices is that they
have real eigenvalues and that they can be diagonalized with respect
to an orthogonal matrix. Thus, if $A$ is an $n\times n$ symmetric
matrix, then it has $n$ real eigenvalues $\lambda_1, \ldots,
\lambda_n$  (not necessarily distinct), and there is
an orthonormal basis of eigenvectors $(u_1, \ldots, u_n)$
(for a proof, see Gallier \cite{Gallbook2}).
Another  fact that is used frequently in optimization problem is that
the eigenvallues of a symmetric matrix are characterized in terms of
what is know as the {\it Rayleigh ratio\/}, defined by
\[
R(A)(x) = \frac{\transpos{x} A x}{\transpos{x} x},\quad x\in \reals^n,
x\not= 0.
\]

The following proposition is often used to prove various
optimization or approximation problems
(for example PCA).

\begin{proposition} ({\it Rayleigh--Ritz})
\label{PCAlem1a} 
If $A$ is a symmetric  $n\times n$ matrix with eigenvalues
$\lambda_1 \geq \lambda_2 \geq \cdots \geq \lambda_n$ and if
$(u_1, \ldots, u_n)$ is any orthonormal basis of eigenvectors
of $A$, where $u_i$ is a unit eigenvector associated with $\lambda_i$,
then
\[
\max_{x\not= 0} \frac{\transpos{x} A x}{\transpos{x}{x}} = \lambda_1
\]
(with the maximum attained for $x = u_1$),  and
\[
\max_{x\not= 0, x \in \{u_1, \ldots, u_k\}^{\perp}} 
\frac{\transpos{x} A x}{\transpos{x}{x}} = \lambda_{k+1}
\]
 (with the maximum attained for $x = u_{k+1}$), where 
$1 \leq k \leq n - 1$. 
Equivalently, if $V_k$ is the subspace spanned by 
$(u_k, \ldots, u_n)$, then
\[
\lambda_k = 
\max_{x\not= 0, x \in V_k} 
\frac{\transpos{x} A x}{\transpos{x}{x}},
\quad k = 1, \ldots, n. 
\]
\end{proposition}

\begin{proof}
First, observe that
\[
\max_{x\not= 0} \frac{\transpos{x} A x}{\transpos{x}{x}} =
\max_{x} \{ \transpos{x} A x \mid  \transpos{x}{x} = 1\},
\]
and similarly,
\[
\max_{x\not= 0, x \in \{u_1, \ldots, u_k\}^{\perp}} 
\frac{\transpos{x} A x}{\transpos{x}{x}} =
\max_{x} \left\{\transpos{x} A x \mid 
(x \in \{u_1, \ldots, u_k\}^{\perp}) \land (\transpos{x}{x} = 1)\right\}.
\]
Since $A$ is a symmetric matrix, its eigenvalues are real and
it can be diagonalized with respect
to an orthonormal basis of eigenvectors, so let
$(u_1, \ldots, u_n)$ be such a basis. If we write
\[ 
x = \sum_{i = 1}^n x_i u_i, 
\]
a simple computation shows that
\[
\transpos{x} A x = \sum_{i = 1}^n \lambda_i x_i^2.
\]
If $\transpos{x} x = 1$, then $\sum_{i = 1}^n  x_i^2 = 1$,
and since we assumed that 
$\lambda_1 \geq \lambda_2 \geq \cdots \geq \lambda_n$, we get
\[
\transpos{x} A x  =  \sum_{i = 1}^n \lambda_i x_i^2 
 \leq   \lambda_1 \biggl(\sum_{i = 1}^n  x_i^2\biggr)
 =  \lambda_1.
\]
Thus, 
\[
\max_{x} \left\{ \transpos{x} A x \mid  \transpos{x}{x} = 1\right\} 
\leq \lambda_1,
\]
and since this maximum is achieved for $e_1 =  (1, 0, \ldots, 0)$,
we conclude that
\[
\max_{x} \left\{ \transpos{x} A x \mid  \transpos{x}{x} = 1\right\} 
= \lambda_1.
\]
Next, observe that $x\in \{u_1, \ldots, u_k\}^{\perp}$ 
and $\transpos{x}{x} = 1$ iff
$x_1 = \cdots = x_k = 0$ and $\sum_{i = 1}^n x_i = 1$.
Consequently, for such an $x$, we have
\[
\transpos{x} A x  =  \sum_{i = k + 1}^n \lambda_i x_i^2 
 \leq   \lambda_{k + 1} \biggl(\sum_{i = k + 1}^n  x_i^2\biggr)
 =  \lambda_{k + 1}.
\]
Thus, 
\[
\max_{x} \left\{ \transpos{x} A x \mid  
(x \in \{u_1, \ldots, u_k\}^{\perp}) \land (\transpos{x}{x} = 1)\right\}
\leq \lambda_{k + 1},
\]
and since this maximum is achieved for 
$e_{k + 1} =  (0, \ldots, 0, 1, 0, \ldots, 0)$ with a $1$ in position $k + 1$,
we conclude that
\[
\max_{x} \left\{ \transpos{x} A x \mid  
(x \in \{u_1, \ldots, u_k\}^{\perp}) \land (\transpos{x}{x} = 1)\right\}
= \lambda_{k + 1},
\]
as claimed.
\end{proof} 

\medskip
Proposition  \ref{PCAlem1} is often known as part of the
{\it Rayleigh--Ritz theorem\/}.
For our purposes, we need the version of Proposition \ref{PCAlem1a}
applying to $\min$ instead of $\max$, whose proof is obtain by
a trivial modification of the proof of Proposition \ref{PCAlem1a}.

\begin{proposition} ({\it Rayleigh--Ritz})
\label{PCAlem1}
If $A$ is a symmetric  $n\times n$ matrix with eigenvalues
$\lambda_1 \geq \lambda_2 \geq \cdots \geq \lambda_n$ and if
$(u_1, \ldots, u_n)$ is any orthonormal basis of eigenvectors
of $A$, where $u_i$ is a unit eigenvector associated with $\lambda_i$,
then
\[
\min_{x\not= 0} \frac{\transpos{x} A x}{\transpos{x}{x}} = \lambda_n
\]
(with the minimum attained for $x = u_n$),  and
\[
\min_{x\not= 0, x \in \{u_{i + 1}, \ldots, u_n\}^{\perp}} 
\frac{\transpos{x} A x}{\transpos{x}{x}} = \lambda_{i}
\]
 (with the minimum attained for $x = u_{i}$), where 
$1 \leq i \leq n - 1$. 
Equivalently, if $W_k = V_{k+1}^{\perp}$ denotes the subspace spanned by
$(u_1, \ldots, u_k)$ (with $V_{n+1} = (0)$), then
\[
\lambda _k =
\min_{x\not= 0, x \in W_k} 
\frac{\transpos{x} A x}{\transpos{x}{x}}
= \min_{x\not= 0, x \in V_{k+1}^{\perp}} 
\frac{\transpos{x} A x}{\transpos{x}{x}}, 
\quad k = 1, \ldots, n.
\]
\end{proposition}

\medskip
As an application of Propositions \ref{PCAlem1a} and \ref{PCAlem1}, we give a proof of
a proposition which is the key to the proof of Theorem
\ref{graphdraw}. First, we need a definition. Given an $n\times n$
symmetric matrix $A$ and an $m\times m$ symmetric $B$, with $m \leq
n$, if $\lambda_1 \geq \lambda_2 \geq \cdots \geq \lambda_n$
are the eigenvalues of $A$ and $\mu_1  \geq \mu_2 \geq \cdots \geq
\mu_m$ are the eigenvalues of $B$, then we say that the
 eigenvalues of $B$  {\it interlace\/} the eigenvalues of $A$ if
\[
\lambda_{n - m + i} \leq \mu_i \leq \lambda_i, \quad i = 1, \dots, m.
\]

\begin{proposition}
\label{interlace}
Let $A$ be an $n\times n$ symmetric matrix, $R$ be an $n\times m$
matrix such that $\transpos{R} R = I$ (with $m \leq n$), and let $B =
\transpos{R} A R$ (an $m\times m$ matrix). The following properties hold:
\begin{enumerate}
\item[(a)]
The eigenvalues of $B$ interlace the eigenvalues of $A$. 
\item[(b)]
If $\lambda_1 \geq \lambda_2 \geq \cdots \geq \lambda_n$
are the eigenvalues of $A$ and $\mu_1  \geq \mu_2 \geq \cdots \geq
\mu_m$ are the eigenvalues of $B$, and if $\lambda_i = \mu_i$, then
there is an eigenvector $v$ of $B$ with eigenvalue $\mu_i$  such that
$R v$ is an eigenvector of $A$ with eigenvalue $\lambda_i$.
\end{enumerate}
\end{proposition}

\begin{proof}
(a)
Let $(u_1, \ldots, u_n)$ be an orthonormal basis of eigenvectors for
$A$,
and let $(v_1, \ldots,v_m)$ be an orthonormal basis of eigenvectors
for $B$. Let $U_j$ be the subspace spanned by $(u_1, \ldots, u_j)$ and
let $V_j$  be the subspace spanned by $(v_1, \ldots, v_j)$.
For any $i$, the subpace $V_i$ has dimension $i$ and the subspace
$\transpos{R} U_{i - 1}$ has dimension at most $i - 1$. Therefore,
there is some nonzero vector 
$v\in V_i \cap (\transpos{R} u_{i -  1})^{\perp}$, and since
\[
\transpos{v} \transpos{R} u_j =  \transpos{(Rv)}  u_j  = 0, \quad j = 1, \ldots, i - 1,
\]
we have $Rv \in (U_{i - 1})^{\perp}$.  By Proposition \ref{PCAlem1a} 
and using the fact that $\transpos{R} R = I$,  we have
\[
\lambda_i \geq \frac{\transpos{(R v)} A R v}{\transpos{(R v)}  R v}
=  \frac{\transpos{v}B v}{\transpos{v} v}.
\]
On the other hand, by Proposition \ref{PCAlem1},
\[
\mu_i =
\min_{x\not= 0, x \in \{v_{i + 1}, \ldots, v_n\}^{\perp}} 
\frac{\transpos{x} B x}{\transpos{x}{x}} = 
\min_{x\not= 0, x \in \{v_{1}, \ldots, v_i\}} 
\frac{\transpos{x} B x}{\transpos{x}{x}},  
\]
so 
\[
\mu_i \leq \frac{\transpos{w}B w}{\transpos{w} w}
\quad
\hbox{for all $w\in V_{i}$},
\] 
and since $v\in V_i$,  we have
\[
\mu_i \leq \frac{\transpos{v}B v}{\transpos{v} v}
\leq 
\lambda_i, \quad i = 1, \ldots, m.
\]
We can apply the  same argument to the symmetric matrices $-A$ and
$-B$, to conclude that 
\[
- \mu_i \leq - \lambda_{n - m + i},
\]
that is,
\[
\lambda_{n - m + i} \leq \mu_i, \quad i = 1, \ldots, m.
\]
Therefore, 
\[
\lambda_{n - m + i} \leq \mu_i \leq \lambda_i, \quad i = 1, \ldots, m,
\]
as desired.

\medskip
(b)
If $\lambda_i = \mu_i$, then 
\[
\lambda_i  = \frac{\transpos{(R v)} A R v}{\transpos{(R v)}  R v} =
\frac{\transpos{v}B v}{\transpos{v} v} = \mu_i,
\]
so $v$ must be an eigenvector for $B$ and $Rv$ must be an eigenvector
for $A$, both for the eigenvalue  $\lambda_i = \mu_i$. 
\end{proof}

\medskip
Observe that Proposition \ref{interlace} implies that
\[
\lambda_n + \cdots + \lambda_{n - m + 1} \leq 
\mathrm{tr}(\transpos{R} A R) \leq \lambda_1 + \cdots + \lambda_m.
\]
The left inequality is used to prove Theorem \ref{graphdraw}.

\medskip
For the sake of completeness, we also prove the Courant--Fischer
characterization
of the eigenvalues of a symmetric matrix.

\begin{theorem} ({\it Courant--Fischer\/})
\label{Courant-Fischer}
Let  $A$ be  a symmetric  $n\times n$ matrix  with eigenvalues
$\lambda_1 \geq \lambda_2 \geq \cdots \geq \lambda_n$ and let
$(u_1, \ldots, u_n)$ be any orthonormal basis of eigenvectors
of $A$, where $u_i$ is a unit eigenvector associated with $\lambda_i$.
If $\s{V}_k$ denotes the set of subspaces of $\reals^n$ of dimension
$k$, then
\begin{align*}
\lambda_k & = \max_{W\in \s{V}_k} \min_{x\in W, x\not= 0} 
\frac{\transpos{x} A x}{\transpos{x} x} \\
\lambda_k & =\min_{W\in \s{V}_{n - k + 1}} \max_{x\in W, x\not= 0} 
\frac{\transpos{x} A x}{\transpos{x} x} .
\end{align*}
\end{theorem}

\begin{proof}
Let us consider the second equality, the proof of the first
equality being similar.
Observe that the space $V_k$ spanned by $(u_k, \ldots, u_n)$ has
dimension $n - k + 1$, and by Proposition \ref{PCAlem1a}, we have
\[
\lambda_k = 
\max_{x\not= 0, x \in V_k} 
\frac{\transpos{x} A x}{\transpos{x}{x}}
\geq \min_{W\in \s{V}_{n - k + 1}} \max_{x\in W, x\not= 0} 
\frac{\transpos{x} A x}{\transpos{x} x}. 
\]
Therefore, we need to prove the reverse inequality; that is, we have
to show that
\[
\lambda _k \leq \max _{x\not= 0, x\in W} 
\frac{\transpos{x} A  x}{\transpos{x} x},
\quad \hbox{for all}\quad W\in \s{V}_{n - k + 1}.
\]
Now, for any  $W\in \s{V}_{n - k + 1}$, if we can prove that
$W\cap V_{k+1}^{\perp} \not= (0)$, then for any nonzero $v\in W\cap
V_{k+1}^{\perp}$,
by Proposition \ref{PCAlem1} , we have
\[
\lambda _k 
= \min_{x\not= 0, x \in V_{k+1}^{\perp}} 
\frac{\transpos{x} A x}{\transpos{x}{x}} 
\leq \frac{\transpos{v} A v}{\transpos{v}{v}}
\leq \max_{x\in W, x\not= 0}  \frac{\transpos{x} A x}{\transpos{x}{x}}.
\]
It remains to prove that $ \mathrm{dim}(W\cap   V_{k+1}^{\perp}) \geq 1$. 
However, $\mathrm{dim}(V_{k+1}) = n - k$, so
$\mathrm{dim}(V_{k+1}^{\perp}) = k$, and by hypothesis
$\mathrm{dim}(W) = n - k + 1$. By the Grassmann relation,
\[
\mathrm{dim}(W) + \mathrm{dim}(V_{k+1}^{\perp}) =
\mathrm{dim}(W\cap   V_{k+1}^{\perp}) + \mathrm{dim}(W + V_{k+1}^{\perp}), 
\]
and since $ \mathrm{dim}(W + V_{k+1}^{\perp}) \leq
\mathrm{dim}(\reals^n) = n$, we get
\[
n - k + 1 + k \leq \mathrm{dim}(W\cap   V_{k+1}^{\perp}) + n;
\]
that is, $1 \leq  \mathrm{dim}(W\cap   V_{k+1}^{\perp})$, as claimed.
\end{proof}

\chapter{Riemannian Metrics on  Quotient Manifolds}
\label{ch3-sec6} 
In order to define a metric on the projective space $\mathbb{RP}^n$, we
need to review a few notions of differential geometry. First, we need to
define the quotient $M/G$ of a manifold by a group acting on $M$.
This section relies heavily on Gallot, Hulin, Lafontaine \cite{Gallot}
and Lee \cite{Lee}, which contain thorough expositions and should be 
consulted for details.

\begin{definition}
\label{actdef}
Recall that an {\it action\/} of a group $G$ (with identity element
$1$) on a set $X$ is a map
$\mapdef{\gamma}{G\times X}{X}$ satisfying the following properties:
\begin{enumerate}
\item[(1)]
$\gamma(1, x) = x$, for all $x\in X$.
\item[(2)]
$\gamma(g_1,\gamma(g_2, x)) = \gamma(g_1g_2, x)$,
for all $g_1, g_2\in G$, and all $x\in X$.
\end {enumerate}
We usually abbreviate $\gamma(g, x)$ by $g\cdot x$.

\medskip
If $X$ is a topological space and $G$ is a topological group,
we say that the action is {\it
  continuous\/} iff the map $\gamma$ is continuous. In this case,
for every $g\in G$, the  map $x \mapsto
g\cdot x$ is a homeomorphism.
If $X$ is a smooth manifold and $G$ is a Lie group,
we say that the action is {\it
  smooth\/} iff the map $\gamma$ is smooth.
In this case, for every $g\in G$, the  map $x \mapsto
g\cdot x$ is a diffeomorphism.
\end{definition}

\remark
To be more precise, what we have defined in Definition \ref{actdef}
is a {\it left action\/} of the group $G$ on the set $X$.
There is also a notion of a {\it right action\/}, but we won't need it.

\medskip
The {\it quotient of $X$ by $G$\/}, denoted $X/G$, is the set of
orbits of $G$; that is, the set of equivalences classes of the
equivalence relation $\simeq$ defined such that, for any $x, y\in X$,
\[
x \simeq y \quad\hbox{iff}\quad  (\exists g\in G )(y = g\cdot x).
\] 
The {\it orbit\/} of $x\in X$ (the equivalence class of $x$) is the set
\[
O_x = \{g\cdot x \mid g\in G\}, 
\] 
also denoted by $G\cdot x$.
If $X$ is a topological space, we give $X/G$ the quotient topology.

\medskip
For any subset $V$ of $X$ and for any $g\in G$, we denote by $gV$ the set
\[
gV = \{g\cdot x \mid x\in V\}.
\]

\medskip
One problem is that  even if $X$ is Hausdorff, $X/G$ may not be.
Thus, we need to find conditions to ensure that $X/G$ is Hausdorff.

\medskip
By a {\it discrete group\/}, we mean a group equipped with the
discrete topology (every subset is open). In other words, we don't
care about the topology of $G$!
The following conditions prove to be useful.

\begin{definition}
\label{properfree}
Let $\mapdef{\cdot}{G\times X}{X}$ be the action of a group $G$ on a
set $X$. We say that $G$ acts {\it freely\/} (or that the action is
{\it free\/})  iff for all $x\in X$ and all $g\in G$, if $g\not= 1$, then
$g\cdot x \not= x$.

\medskip
If $X$ is a locally compact space and $G$ is a discrete group acting
continuously on $X$,
we say that $G$ acts {\it
  properly\/} (or that the action is {\it proper\/}) iff
\begin{enumerate}
\item[(i)]
For every $x\in X$, there is some open subset $V$ with $x\in V$ such
that $gV\cap V \not= \emptyset$ for only finitely many $g\in G$.  
\item[(ii)]
For all $x, y\in X$, if $y\notin G\cdot x$ ($y$ is not in the orbit of $x$),
then there exist some open sets $V, W$ with $x\in V$ and $y\in W$ such
that $gV \cap W = 0$ for all $g\in G$.
\end{enumerate}
\end{definition}

\medskip
The following proposition gives  necessary and sufficient conditions
for a discrete group to act freely and properly often found in the
literature
(for instance, O'Neill \cite{Oneill},  Berger and Gostiaux
\cite{BergerGos},
and do Carmo \cite{DoCarmo}, but beware that in this last reference
Hausdorff separation is not required!).

\begin{proposition}
\label{freep1}
If $X$ is a locally compact space and $G$ is a discrete group, then
a smooth action of $G$ on $M$ is free and proper iff the following
conditions hold:
\begin{enumerate}
\item[(i)]
For every $x\in X$, there is some open subset $V$ with $x\in V$ such
that $gV\cap V = \emptyset$ for all  $g\in G$ such that $g\not= 1$.  
\item[(ii)]
For all $x, y\in X$, if $y\notin G\cdot x$ ($y$ is not in the orbit of $x$),
then there exist some open sets $V, W$ with $x\in V$ and $y\in W$ such
that $gV \cap W = 0$ for all $g\in G$.
\end{enumerate}
\end{proposition}

\begin{proof}
Condition (i) of Proposition \ref{freep1} 
implies condition (i) of  Definition \ref{properfree}, and
condition (ii) is the same in  Proposition \ref{freep1} 
and  Definition \ref{properfree}.
If (i) holds, then the action must be free since if $g\cdot x = x$,
then $gV\cap V \not= \emptyset$, which implies that $g =1$.

\medskip
Conversely, we just have to prove that the conditions of  Definition
\ref{properfree}
imply condition (i) of Proposition \ref{freep1}.
By (i) of Definition \ref{properfree}, there is some open
subset $U$ containing $x$ and a finite number of elements of $G$,
say $g_1, \ldots, g_m$, with $g_i \not= 1$, such that
\[
g_iU\cap U \not= \emptyset, \quad i = 1, \ldots, m.
\]
Since our action is free and $g_i \not= 1$, we have $g_i \cdot x \not
= x$, so by Hausdorff separation, there exist some open subsets
$W_i, W_i'$, with $x\in W_i$ and  $g_i\cdot x\in W_i'$, such that $W_i\cap W_i' =
\emptyset$, $i = 1, \ldots, m$.
Then, if we let
\[
V = W \cap \bigg(\bigcap_{i = 1}^m (W_i \cap g_i^{-1} W_i')\bigg),
\]
we see that $V\cap g_iV = \emptyset$, and since $V \subseteq W$,
we also have $V\cap gV = \emptyset$ for all other $g\in G$. 
\end{proof}

\remark
The action of a discrete group satisfying the properties of Proposition
\ref{freep1}
is often called ``properly discontinuous.''  However, as pointed out
by Lee (\cite{Lee}, just before Proposition 9.18), this term is
self-contradictory
since such actions are smooth, and thus continuous!

\medskip
We also need covering maps.

\begin{definition}
\label{covermap}
Let $X$ and $Y$ be two topological spaces. A map $\mapdef{\pi}{X}{Y}$
is a {\it covering map\/} iff the following conditions hold:
\begin{enumerate}
\item[(1)]
The map $\pi$ is continuous and surjective.
\item[(2)]
For every $y\in Y$, there is some open subset $W\subseteq Y$ with
$y\in W$, such that
\[
\pi^{-1}(W) = \bigcup_{i\in I} U_i,
\]
where the $U_i \subseteq X$ are pairwise disjoint open subsets such
that the restriction of $\pi$ to $U_i$ is a homeomorphism for every
$i\in I$.
\end{enumerate}
If $X$ and $Y$ are smooth manifolds, we assume that $\pi$ is smooth
and that the restriction of $\pi$ to each $U_i$ is a diffeomorphism.
\end{definition}

\medskip
Then, we have the following useful result.

\begin{theorem}
\label{quotman}
Let $M$ be a smooth manifold and let $G$ be discrete  group
acting smoothly, freely and properly on $M$.
Then there is a unique structure of smooth manifold on $M/G$ such that
the projection map $\mapdef{\pi}{M}{M/G}$ is a covering map.
\end{theorem}

For a proof, see  Gallot, Hulin, Lafontaine \cite{Gallot}
(Theorem 1.88) or Lee \cite{Lee} (Theorem 9.19).

\medskip
Real projective spaces are illustrations of Theorem \ref{quotman}.
Indeed, if  $M$ is the unit $n$-sphere $S^n \subseteq \reals^{n+1}$
and $G = \{I, -I\}$, where $-I$ is the antipodal map, then the
conditions of Proposition \ref{freep1} are easily checked (since $S^n$
is compact), and consequently the quotient
\[
\mathbb{RP}^n = S^n/G
\]
is a smooth manifold and the projection map
$\mapdef{\pi}{S^n}{\mathbb{RP}^n}$
is a covering map. The fiber $\pi^{-1}([x])$ of every point 
$[x] \in \mathbb{RP}^n$ consists of two antipodal points: $x, -x\in S^n$.

\medskip
The next step is see how a Riemannian metric on $M$ 
induces a Riemannian metric on the quotient manifold $M/G$.

\begin{definition}
\label{localisom}
Given any two Riemmanian manifolds $(M, g)$ and $(N, h)$ 
a smooth map $\mapdef{f}{M}{N}$ is a {\it local isometry\/} iff
for all $p\in M$, the tangent map
$\mapdef{df_p}{T_p M}{T_{f(p)} N}$ is an orthogonal transformation
of the Euclidean spaces $(T_p M, g_p)$ and $(T_{f(p)} N, h_{f(p)}))$.
Furthermore, if $f$ is a diffeomorphism, we say that $f$ is an {\it isometry\/}. 
\end{definition}

\medskip
The Riemannian version of a covering map is the following:

\begin{definition}
\label{covermap2}
Let $(M, g)$ and $(N, h)$ be two Riemannian manifolds. A map $\mapdef{\pi}{M}{N}$
is a {\it Riemannian covering map\/} iff the following conditions hold:
\begin{enumerate}
\item[(1)]
The map $\pi$ is a smooth covering.
\item[(2)]
The map $\pi$ is a local isometry.
\end{enumerate}
\end{definition}

\medskip
The following theorem is the Riemannian version of Theorem \ref{quotman}.

\begin{theorem}
\label{quotman2}
Let $(M, h)$ be a Riemannian manifold and let $G$ be discrete  group
acting smoothly, freely and properly on $M$, and such that
the map $x \mapsto \sigma\cdot x$ is an isometry for  all $\sigma\in G$.
Then there is a unique structure of Riemannian manifold on $N = M/G$ such that
the projection map $\mapdef{\pi}{M}{M/G}$ is a Riemannian  covering map.
\end{theorem}

\begin{proof}[Proof sketch]
For a complete proof see  Gallot, Hulin, Lafontaine \cite{Gallot}
(Proposition 2.20). To define a Riemannian metric $g$ on $N = M/G$
we need to define an inner product $g_p$ on the tangent space $T_p N$
for every $p\in N$. Pick any $q_1\in \pi^{-1}(p)$ in the fibre of
$p$. Because $\pi$ is a Riemannian covering map, it is a local
diffeomorphism, and thus  $\mapdef{d\pi_{q_1}}{T_{q_1} M}{T_p M}$ is an
isometry. Then, given any two tangent vectors $u, v\in T_p N$,  we
define their inner product $g_p(u, v)$ by
\[
g_p(u, v) = h_{q_1}(d\pi_{q_1}^{-1}(u), d\pi_{q_1}^{-1}(v)). 
\]
Now, we need to show that $g_p$ does not depend on the choice of $q_1\in \pi^{-1}(p)$.
So, let $q_2\in \pi^{-1}(p)$ be any other point in the fibre of
$p$. By definition of $N = M/G$, we have $q_2  = g\cdot q_1$ for some $g\in
G$, and we know that the map $f\co q \mapsto g \cdot q$ is an isometry
of $M$. Now, since $\pi = \pi \circ f$ we have
\[
d\pi _{q_1}=   d \pi_{q_2} \circ df_{q_1},
\]
and since   $\mapdef{d\pi_{q_1}}{T_{q_1} M}{T_p M}$ and
$\mapdef{d\pi_{q_2}}{T_{q_2} M}{T_p M}$ 
are isometries, we get 
\[
d\pi _{q_2}^{-1}= df_{q_1}  \circ d \pi_{q_1}^{-1}.
\]
But $\mapdef{df_{q_1}}{T_{q_1}M}{T_{q_2} M}$ is also an isometry, 
so 
\[
h_{q_2}(d\pi_{q_2}^{-1}(u), d\pi_{q_2}^{-1}(v)) = 
h_{q_2}(df_{q_1}(d\pi_{q_1}^{-1}(u)), df_{q_1}(d\pi_{q_2}^{-1}(v))) = 
h_{q_1}(d\pi_{q_1}^{-1}(u), d\pi_{q_1}^{-1}(v)). 
\]
Therefore, the inner product $g_p  $ is well defined on $T_p N$.
\end{proof}

\medskip
Theorem \ref{quotman2} implies that every Riemannian metric $g$ on the
sphere $S^n$ induces a Riemannian metric $\widehat{g}$ on the projective
space $\mathbb{RP}^n$, in such a way that the projection 
$\mapdef{\pi}{S^n}{\mathbb{RP}^n}$ is a Riemannian covering.
In particular, if $U$ is an open hemisphere obtained by removing its
boundary $S^{n-1}$ from a closed hemisphere, then
$\pi$ is an isometry between $U$ and its image 
$\mathbb{RP}^n - \pi(S^{n-1}) \approx \mathbb{RP}^n -
\mathbb{RP}^{n-1}$.  

\medskip
We also observe that for any two points
$p= [x]$ and $q= [y]$ in $\mathbb{RP}^n$, where $x, y\in S^n$, 
if $x\cdot y = \cos\theta$, with  $0 \leq \theta \leq \pi$, 
then there are two possibilities:
\begin{enumerate}
\item
 $x\cdot y \geq 0$, which means that $0 \leq \theta \leq \pi/2$, or
\item
 $x\cdot y < 0$, which means that $\pi/2 < \theta \leq \pi$.
\end{enumerate}
In the second case, since $[-y] = [y]$ and $x\cdot (-y) = -x\cdot y$,
we can replace the representative $y$ of $q$ by $-y$, and we have
$x\cdot (-y) = \cos(\pi - \theta)$, with $0 \leq  \pi - \theta < \pi/2$.
Therefore, in all cases, for any two points $p, q\in \mathbb{RP}^n$,
we can find an open hemisphere $U$ such that $p = [x], q = [y]$, 
$x, y\in U$, and $x\cdot y \geq 0$; that is, the angle $\theta\geq 0$
between $x$ and $y$ is at most $\pi/2$.
This fact together with the following simple proposition will allow us
to figure out the distance (in the sense of Riemannian geometry)
between two points in $\mathbb{RP}^n$.

\begin{proposition}
\label{geolem1}
Let $\mapdef{\pi}{M}{N}$ be a  Riemannian covering map between two Riemannian manifolds  
 $(M, g)$ and $(N, h)$. Then, the geodesics of $(N,h)$ are the
 projections of geodesics in $(M,g)$ (i.e., curves $\pi\circ \gamma$ in
 $(N, h),$ where
 $\gamma$ is a geodesic in $(M, g)$),
and the geodesics of $(M,g)$ are the
 liftings of geodesics in $(N,h)$ (i.e., curves $\gamma$ of $(M, g)$,
 such that $\pi\circ \gamma$ is a geodesic in $(N, h)$).
\end{proposition}

The proof of Proposition \ref{geolem1} can be found in 
Gallot, Hulin, Lafontaine \cite{Gallot} (Proposition 2.81).

\medskip
Now, if $(M, g)$ is a connected Riemannian manifold, recall that we
define the distance $d(p, q)$ between two points $p, q\in M$ as 
\[
d(p, q) = \inf\{ L(\gamma) \mid \mapdef{\gamma}{[0, 1]}{M}\},
\]
where $\gamma$ is any piecewise $C^1$-curve from $p$ to $q$,
and 
\[
L(\gamma) = \int_0^1 \sqrt{g(\gamma'(t), \gamma'(t))}\, dt
\]
is the length of $\gamma$.
It is well known that $d$ is a metric on $M$.
The Hopf-Rinow Theorem (see Gallot, Hulin, Lafontaine \cite{Gallot},
Theorem 2.103) says among other things that $(M, g)$ is geodesically
complete
(which means that every geodesics $\gamma$ of $M$ can be extended to a
geodesic   $\widetilde{\gamma}$ defined on all of  $\reals$)
iff any two points of $M$ can be joined by a minimal geodesic iff $(M, d)$ is
a complete metric space. 
Therefore, in a complete (connected) manifold
\[
d(p, q) = \inf\{ L(\gamma) \mid \mapdef{\gamma}{[0, 1]}{M} \quad\text{is a
    geodesic}\}.
\]
In particular, compact manifolds are complete, so
the distance between two points is the infimum of the length of
minimal geodesics joining these points.

\medskip
Applying this to $\mathbb{RP}^n$ and the canonical Euclidean 
metric induced by $\reals^{n+1}$, since geodesics of $S^n$ are great
circles, by the discussion above, for any two points
$p = [x]$ and $q = [y]$ in $\mathbb{RP}^n$, with $x, y\in S^n$,
the distance between them is given by
\[
d(p, q) = d([x], [y]) =
\begin{cases}
\cos^{-1}(x\cdot y)& \text{if $x\cdot y \geq 0$} \\
\cos^{-1}(- x\cdot y) & \text{if $x\cdot y < 0$}.
\end{cases}
\]
Here $\cos^{-1}(z) = \arccos(z)$ is the unique angle $\theta\in
[0,\pi]$
such that $\cos(\theta) = z$. Equivalently,
\[
d([x], [y])  = \cos^{-1}(|x\cdot y)|,
\]
and
\[
d([x], [y]) =\min\{\cos^{-1}(x\cdot y), \pi - \cos^{-1}(x\cdot y)\}.
\]
If the representatives $x, y\in \reals^{n + 1}$ of $p = [x]$ and $q =
[q]$ are not unit vectors, then
\[
d([x], [y])  = \cos^{-1}\left(\frac{|x\cdot y|}{\norme{x}\norme{y}}\right).
\]
Note that $0 \leq d(p, q) \leq \pi/2$.

\medskip
Now, the Euclidean distance between $x$ and $y$ on $S^n$ is given by
\[
\norme{x - y}^2_2 = \norme{x}^2_2 + \norme{y}^2_2 - 2 x\cdot y
= 2 - 2 \cos\theta = 4\sin^2(\theta/2).
\]
Thus,
\[
\norme{x- y}_2 = 2 \sin(\theta/2), \quad 0 \leq \theta \leq \pi.
\]
It follows that for any $x\in S^n$, and for any subset $A \subseteq
S^n$, a point $a\in A$ minimizes the distance $d_{S^n}(x, a) =
\cos^{-1}(x\cdot a) = \theta$ on $S^n$ iff it minimizes the Euclidean distance
$\norme{x - a}_2 = 2\sin(\theta/2)$  (since $0 \leq
\theta\leq \pi$).  Then, on $\mathbb{RP}^n$, 
for any point $p = [x]\in \mathbb{RP}^n$ and any 
$A \subseteq \mathbb{RP}^n$, a point  $[a]\in A$ minimizes the distance
$d([x], [a])$ on $\mathbb{RP}^n$ iff it minimizes 
$\min\{\norme{x - a}_2, \norme{x + a}_2\}$.
So, we are looking for $[b]\in A$ such that
\begin{align*}
\min\{\norme{x - b}_2, \norme{x + b}_2\} & = 
\min_{[a] \in A} \min\{\norme{x - a}_2, \norme{x + a}_2\} \\
&  = \min\{ \min_{[a] \in A} \norme{x - a}_2, \min_{[a] \in A} \norme{x + a}_2\}. 
\end{align*}
If the subset $A\subseteq S^n$ is closed under the antipodal map (which
means that if $x\in A$, then $-x\in A$), then finding 
$\min_{a\in A} d([x], [a])$  on $\mathbb{RP}^n$ is equivalent to
finding $\min_{a\in A} \norme{x - a}_2$, the minimum of the Euclidean
distance. This is the case for the set $\s{X}$ in Section
\ref{ch3-sec2} and the set $\s{K}$ in
Section \ref{ch3-sec3}.

\medskip\noindent
{\bf Acknowlegments}: 
First, it must be said that the seminal and highly original work of
Jianbo Shi and Stella Yu on normalized cuts, 
was the source of inspiration for this document.
I also wish to thank Katerina Fragkiadaki for pointing out a number of
mistakes in an earlier version of this paper. Roberto Tron
also made several suggestions that contributed to improving this report.
Katerina, Ryan Kennedy, Andrew Yeager, and Weiyu Zhang
made many useful comments and suggestions.
Thanks to Dan Spielman for making available his lovely survey on
spectral graph theory, and to
Kostas for giving me the opportunity to
hold hostage a number of people for three Friday afternoons  in a row.

   %

\bibliographystyle{plain} 
\end{document}